\documentclass[sigconf, nonacm]{acmart}

\newcommand\vldbdoi{XX.XX/XXX.XX}
\newcommand\vldbpages{XXX-XXX}
\newcommand\vldbvolume{14}
\newcommand\vldbissue{1}
\newcommand\vldbyear{2020}
\newcommand\vldbauthors{\authors}
\newcommand\vldbtitle{\shorttitle} 
\newcommand\vldbavailabilityurl{URL_TO_YOUR_ARTIFACTS}
\newcommand\vldbpagestyle{plain} 

\usepackage{balance}
\usepackage{diagbox}
\usepackage{slashbox}
\usepackage{multirow}
\usepackage[caption=false]{subfig}
\usepackage{xcolor}
\usepackage{subfiles}
\usepackage{pifont}
\usepackage{adjustbox}
\usepackage{diagbox}
\usepackage{slashbox}
\usepackage{tcolorbox}
\usepackage{enumitem}

\newcommand{\rev}[1]{{\textcolor{black}{#1}}}
\newcommand{\vl}[1]{{\textcolor{black}{#1}}}

\begin{document}
\title{OEBench: Investigating Open Environment Challenges in Real-World Relational Data Streams}

\author{Yiqun Diao}
\affiliation{%
  \institution{National University of Singapore}
}
\email{yiqun@comp.nus.edu.sg}

\author{Yutong Yang}
\affiliation{%
  \institution{National University of Singapore}
}
\email{e0425605@u.nus.edu}

\author{Qinbin Li*}
\affiliation{%
  \institution{University of California, Berkeley}
}
\email{qinbin@berkeley.edu}

\author{Bingsheng He}
\affiliation{%
  \institution{National University of Singapore}
}
\email{hebs@comp.nus.edu.sg}

\author{Mian Lu}
\affiliation{%
  \institution{4Paradigm Inc.}
}
\email{lumian@4paradigm.com}

\begin{abstract}
How to get insights from relational data streams in a timely manner is a hot research topic. Data streams can present unique challenges, such as distribution drifts, outliers, emerging classes, and changing features, which have recently been described as \textit{open environment} challenges for machine learning. While existing studies have been done on incremental learning for data streams, their evaluations are mostly conducted with synthetic datasets. Thus, a natural question is how those open environment challenges look like and how existing incremental learning algorithms perform on real-world relational data streams. To fill this gap, we develop an Open Environment Benchmark named \textit{OEBench} to evaluate open environment challenges in real-world relational data streams. Specifically, we investigate 55 real-world relational data streams and establish that open environment scenarios are indeed widespread, which presents significant challenges for stream learning algorithms. Through benchmarks with existing incremental learning algorithms, we find that increased data quantity may not consistently enhance the model accuracy when applied in open environment scenarios, where machine learning models can be significantly compromised by missing values, distribution drifts, or anomalies in real-world data streams. The current techniques are insufficient in effectively mitigating these challenges brought by open environments. More researches are needed to address real-world open environment challenges. All datasets and code are open-sourced in \url{https://github.com/sjtudyq/OEBench}.
\end{abstract}

\maketitle

\pagestyle{\vldbpagestyle}
\begingroup\small\noindent\raggedright\textbf{PVLDB Reference Format:}\\
\vldbauthors. \vldbtitle. PVLDB, \vldbvolume(\vldbissue): \vldbpages, \vldbyear.\\
\href{https://doi.org/\vldbdoi}{doi:\vldbdoi}
\endgroup
\begingroup
\renewcommand\thefootnote{}\footnote{\noindent
*Corresponding author.

\noindent This work is licensed under the Creative Commons BY-NC-ND 4.0 International License. Visit \url{https://creativecommons.org/licenses/by-nc-nd/4.0/} to view a copy of this license. For any use beyond those covered by this license, obtain permission by emailing \href{mailto:info@vldb.org}{info@vldb.org}. Copyright is held by the owner/author(s). Publication rights licensed to the VLDB Endowment. \\
\raggedright Proceedings of the VLDB Endowment, Vol. \vldbvolume, No. \vldbissue\ %
ISSN 2150-8097. \\
\href{https://doi.org/\vldbdoi}{doi:\vldbdoi} \\
}\addtocounter{footnote}{-1}\endgroup

\ifdefempty{\vldbavailabilityurl}{}{
\vspace{.3cm}
\begingroup\small\noindent\raggedright\textbf{PVLDB Artifact Availability:}\\
The source code, data, and/or other artifacts have been made available at \url{https://github.com/sjtudyq/OEBench}.
\endgroup
}

\section{Introduction}


Recently, the concept of \textit{open environment} learning, where data or tasks can change over time, has been introduced by \citet{zhou2022open}. Four primary open environment challenges are identified. Firstly, the data distribution may shift owing to subtle environmental changes. Secondly, outliers or emerging new classes can occur, like new viruses or diseases. Thirdly, the feature set can evolve, with new attributes being added or existing ones being dropped due to factors such as the installation or breakdown of sensors. Lastly, the machine learning task itself may transform, e.g., from a focus on accuracy to computational efficiency. \rev{In this work, we focus on the change of data in real-world relational data streams, covering the first three challenges mentioned. The change of learning task is much more complicated and we regard it as future works.}

\rev{Relational data streams have a lot of applications such as stock market and sensor data. Thus, various stream learning algorithms have been developed to obtain the insights from the data streams timely. Previous stream learning studies generally focus on drift detection, efficiency, and real-time processing \citep{gama2009issues, gama2013evaluating, hayes2019memory, ksieniewicz2022stream, souza2020challenges}. Open environment learning acknowledges that the environment can change dynamically and unpredictably over time. It recognizes a series of open environment challenges, including (1) shifting data distributions over time, (2) the emergence of outliers or new classes, and (3) incremental/decremental dimensions in feature space. Open environment challenges focus more on the model's adaptability to these changes and robustness to unforeseen situations. We list some real-world examples of open environment challenges as follow. } 

\rev{\textbf{Air quality surveillance.} Consider an air quality surveillance system. (1) Factors such as industrial emissions, vehicular traffic, and meteorological variations can change unpredictably, which can cause the challenge of distribution drifts. (2) Unexpected events, like industrial spills or large-scale fires, can introduce new types of pollutants or unprecedented pollution levels not present in the training data, leading to the challenge of outliers or new classes. (3) Technological advancements lead to newer, more accurate air quality sensors replacing older ones, causing potential changes in the metrics or pollutants being monitored. This poses the challenge of incremental/decremental feature space.}

\rev{\textbf{Energy prediction.} In the realm of energy usage prediction, dynamic changes are common. (1) Societal behavior shifts or new industry practices can modify energy consumption or production patterns, leading to the challenge of distribution drifts. (2) Rapid technological adoption, like a surge in electric vehicle usage, or the launch of a new energy-intensive industry can introduce energy patterns not seen during model training, thereby presenting the problem of outliers or novel classes. (3) The evolution of technology might introduce new energy sources or retire older ones. This can bring about changes in the kind of data collected, representing the challenge of incremental/decremental feature space.}

However, there lacks empirical investigation of these scenarios in real-world relational data streams. Although various incremental learning algorithms have been developed~\citep{zenke2017continual,chaudhry2018riemannian,masana2022class}, their evaluations are mostly conducted with manually partitioned datasets. \textit{How do the open environment challenges such as distribution drifts, outliers, emerging classes, and changing features look like in real-world relational data streams? How do these open environment challenges affect the effectiveness and efficiency of stream learning algorithms?} \vl{It calls for a systematic methodology to identify the proposed three challenges and a comprehensive evaluation of existing incremental learning algorithms on real-world relational data streams.}

\vl{In response, this work makes the first attempt to systematically study and benchmark the open environment challenges in real-world relational data streams, which narrows down the gap between the vision paper \citep{zhou2022open} and real-world scenarios. Specifically, we investigate 55 real-world relational data streams collected from public repositories to show that open environment challenges are widespread. We develop a scalable, open-source benchmark to quantitatively study the open environment challenges in real-world relational data streams and evaluate their magnitude. As it is rather complex to analyze all data streams, and also for the simplicity of benchmark design, we select a small number of representative datasets, according to three different open environment aspects. Then we evaluate 10 existing stream learning and incremental learning algorithms in real-world data streams. Our collected datasets cover a much wider range in the three open environment challenges than previous datasets and benchmarks. Our methodology of processing and selecting real-world relational data streams can be easily extended to future new datasets. Our evaluation framework can be applied to new algorithms. }

    \rev{We verify that the open environment challenges widely exist in our collected 55 datasets: 90\% datasets have over 2\% detected outliers; 80\% datasets have over 10\% windows where distribution drifts are detected with its adjacent window. 40\% datasets have over 5\% data items with missing values. We also observe that the model accuracy sometimes degrades significantly at the occurrence of open environment challenges. Despite the ubiquity of open environment challenges in real-world relational data streams, it is still under explored how to address them.}
    
    We further categorize our findings in two categories: similar findings to previous studies and contrary findings to previous studies. Similar to previous studies, we verify that (1) more updates (smaller batch size or more epochs) can generally improve the model accuracy \citep{kandel2020effect}; (2) large models are prone to overfitting and lightweight models are recommended in relatively simple relational data streams \citep{caruana2000overfitting,shwartz2022tabular}; (3) distribution drifts and outliers can severely harm the accuracy of stream learning models \citep{hoens2012learning}. \vl{Note that, previous studies are mostly based on synthetic data streams, or have limited coverage on real-world datasets but overlook the complex open environment challenges in different real-world data streams.}
    
    \rev{Contrary to findings of prior studies: (1) Previous studies \citep{mikolajczyk2018data, stockwell2002effects} show that more training data usually improves model accuracy. In our study, we find that more training data does not necessarily improve model accuracy under some open environment settings. As we observe in case studies, at the point of obvious distribution drifts or outliers, the test loss surges since old data or unreliable data harm the model accuracy. (2) Previous studies \citep{perez2020improving, li2015outlier} conclude that removing outliers helps improve model accuracy. In our study, it does not always hold in some open environment scenarios, since it is unknown whether the detected outliers are really outliers in real-world data streams. (3) Previous studies \citep{zhou2022model, street2001streaming} claim that larger exemplar storage buffer size or larger ensemble size can help improve model accuracy. However, we find such conclusion is not guaranteed in some real-world data streams with open environment challenges. Under distribution drifts, sometimes old exemplars or old models may not well adapt to new environments and lead to biased supervision. Thus, open environment problems bring great challenges in learning real-world relational data streams. }


\section{Open Environment Challenges}

\rev{Consider a data stream $T$. We view it as a sequence of windows $\mathbf{T}=\{T_1, T_2, ..., T_n,...\}$. The window size is a tunable parameter. In each window $T_k$, we train a model $f_k$ using only the previous model $f_{k-1}$, data of the current window, and a limited number of samples from previous windows (if available). In an open environment learning context, the learning objective is that the trained model can generalize well on the upcoming window $T_{k+1}$.}

\rev{Consider the following example. Assume we want to predict air quality from observed statistics. Suppose the window size is one day. A model works for one day to perform inference on the observations. Then we update the model with observations in the past day. The goal is to maintain a good predictor that can work well on the near future data, under scenarios where the environments may change, e.g. climate change, extreme weather, sensor damage, and etc. This requires the model to well adapt to the following possible open environment challenges, including distribution drifts, outliers and incremental/decremental features.}

\subsection{Distribution Drifts}
\label{sec:drift}



In discussing distribution drifts, we adhere to the definitions in \citet{souza2020challenges}. Denote the feature and label of window $k$ as $X_k$ and $Y_k$. Distribution drifts can be categorized into three main types:
\begin{itemize}
\item Prior probability drift occurs when  $P(Y_i|X_i) = P(Y_j|X_j)$ and $P(Y_i) \neq P(Y_j)$. It happens exclusively in $\mathcal{Y} \rightarrow \mathcal{X}$ problems (the features are dependent on the labels). 
\item Covariate drift occurs when $P(Y_i|X_i) = P(Y_j|X_j)$ and $P(X_i) \neq P(X_j)$ . It only happens in $\mathcal{X} \rightarrow \mathcal{Y}$ problems (the labels are dependent on the features). 
\item Concept drift occurs when $P(Y_i|X_i) \neq P(Y_j|X_j)$. 
\end{itemize}



\rev{Distribution drifts are challenging since they can cause some historical data to be misleading for current windows. For example, an environment predictor trained on statistics during summer may not generalize well in winter. A possible solution is to apply drift detectors and re-train the model after drift alerts.}

\subsection{Outliers or New Classes}

\rev{Another challenge encountered in open environment learning is the emergence of outliers or new classes. In this context, an outlier refers to an observation that deviates significantly from the other observations, often due to measurement error or an exception in the data. Meanwhile, a new class refers to a novel category or label that is not present in the previous windows.}

\rev{The appearance of outliers or new classes can substantially harm the model accuracy. For example, an abnormal value of the target can lead to very high loss of the specific element, which could greatly affect the model parameters and lead to poor accuracy. To effectively manage this challenge, the learning model needs to be capable of identifying these outliers or new classes and adapt accordingly, either by learning the new class through additional training or by properly removing the outliers.}

\subsection{Incremental/Decremental Features}
\rev{Traditional machine learning methodologies are based on the assumption that all samples reside in the same feature space. Open environment learning challenges this assumption, considering instead a dynamic feature space that may incrementally expand or decrementally shrink. In other words, new data may come with additional features (incremental features) or lack some of the previously observed features (decremental features).}

\rev{Many existing studies have addressed the issue of missing values in machine learning, yet they often operate under the assumption that the entire feature space is known. This is not the case in open environment learning, where incremental or decremental features in the data stream are unexpected. This difference creates a challenge, as the model trained on previous data would have been based on a different feature space.}

\rev{Decremental features can be addressed by filling missing values. However, incremental features are difficult to address since the model does not account for the incoming feature. One simple approach to dealing with incremental features is to discard them, although this strategy leads to an under-utilization of potentially valuable information. An alternative solution is to retrain the model to incorporate the new features, at the risk of causing the model to forget previously acquired knowledge.}

\section{Related Works}
\label{chapter:background}



\subsection{Incremental Learning Algorithms}
\label{sec:incemental}
\rev{Most previous studies on incremental learning \citep{zenke2017continual,chaudhry2018riemannian,masana2022class} can be formulated as learning on a data stream $T$ divided into a sequence of windows $\mathbf{T}=\{T_1, T_2, ..., T_n,...\}$. All windows are non-overlapping with different classes. In each window $T_k$, the task is to train a model $f_k$ using only the previous model $f_{k-1}$, data of the current window, and a limited number of exemplars (if available). The goal of incremental learning is to learn a good model $f_k$ working well on seen classes in $T_1,T_2,...,T_{k}$.}

However, incremental learning methods may not be suited to the open environment context, as the goal of open environment learning is to work well for near future window $T_{k+1}$ where unpredictable changes can happen. Such changes could render old data unsuitable for a new environment. 

From the perspective of regularization, EWC \citep{kirkpatrick2017overcoming} penalizes the change in crucial parameters based on the Fisher Information Matrix. LwF \citep{li2017learning} integrates the prediction of the previous model, which reduces the over-confidence towards seen classes of the current window. From the perspective of storing exemplars, iCaRL \citep{rebuffi2017icarl} selects exemplars that are close to the mean representation of each class. From the perspective of parameter isolation, PackNet \citep{mallya2018packnet} prunes the neural network for new windows while keeping important parameters frozen. More detailed discussions are in Appendix \ref{sec:il_detail} of our full version \cite{diao2023oebench}.

\rev{A summary of incremental learning works is in Table \ref{tbl:ilwork}. As we can see, none of these incremental learning algorithms are designed to tackle changing feature spaces or outliers. Some algorithms are even inapplicable to real-world relational data streams due to the specific design for image, requiring auxiliary datasets or no scalability towards infinite streams. Therefore, prior incremental learning algorithms can hardly address open environment challenges in real-world relational data streams.}

\begin{table}[htpb]
\newcommand{\y}{\ding{51}}
\newcommand{\n}{\ding{55}}
\caption{Summary of incremental learning works and difficulties in adapting to real-world relational data streams. \y \ and \n \  indicate whether the methods consider each open environment challenge in real-world relational data streams.}
\label{tbl:ilwork}
\centering
\resizebox{1\columnwidth}{!}{
\begin{tabular}{|c|c|c|c|c|c|}
\hline
\multirow{3}{*}{Paper} & \multirow{3}{*}{Learn from} & Difficulties in  & Incremental/  & \multirow{3}{*}{Drifts}  & \multirow{3}{*}{Outliers}  \\
& &  real-world relational & decremental & &  \\ 
& & data streams & features & &  \\ 
\hline
\citep{kirkpatrick2017overcoming, zenke2017continual, aljundi2018memory, chaudhry2018riemannian} & Important parameters & N/A &\n & \y&\n  \\ \hline
\citep{li2017learning, hu2021distilling, toldo2022bring} & Prior model outputs & N/A &\n & \y&\n  \\ \hline
\citep{dhar2019learning, smith2021always, wu2021striking} & Prior model outputs & Specific for image &\n  & \n &\n \\ \hline
\citep{zhang2020class} & Prior model outputs & No auxiliary data &\n  & \n &\n \\ \hline
\citep{rebuffi2017icarl, chaudhry2018riemannian, wu2019large, hou2019learning, yan2021dynamically} & Stored exemplars & Not for regression &\n & \y&\n  \\ \hline
\multirow{2}{*}{\citep{mallya2018packnet, aljundi2017expert}} & \multirow{2}{*}{Fixed model part} & Not scalable for  & \multirow{2}{*}{\n} &\multirow{2}{*}{\n}&\multirow{2}{*}{\n}  \\
& & infinite streams  & & & \\
\hline
\end{tabular}
}
\end{table}

\subsection{Incremental Learning Benchmarks}
The evaluation of incremental learning techniques has led to various benchmarks. For example, \citet{masana2022class} compare 13 incremental learning algorithms. They divide the dataset into several windows by class, and evaluate the average accuracy across all previously seen classes in each window. Similar settings are widely adopted by incremental learning works \citep{zenke2017continual,chaudhry2018riemannian,masana2022class}. 

Besides splitting the datasets, other evaluation methods include permuting or rotating image datasets to construct drifts or introduce new classes \citep{farquhar2018towards, lomonaco2021avalanche, mai2022online}. Real-world video datasets are also proposed for incremental learning \citep{lomonaco2017core50, roady2020stream, she2020openloris}. However, these datasets merely simulate emerging classes and drifts in open environment challenges, and do not consider relational data streams.

\rev{For real-world relational data streams, there are few prior benchmarks. \citet{bifet2010moa} proposes a stream learning system MOA, but it only has five real-world stream datasets.} \citet{souza2020challenges} proposes to conduct experiments on insects and alter the temperature to simulate drifts. Both balanced and imbalanced insect classes are designed, and raw data are recorded to form new data streams. The metric is prequential accuracy, which means first testing and then training each window of data.

\rev{We summarize the data streams used in previous incremental learning works in Table \ref{tbl:benchwork}. As we can see, none of prior works consider all the three aspects of open environment challenges in real-world relational data streams. The first two rows are incremental learning benchmarks on computer vision datasets, which consider neither relational datasets nor incremental/decremental feature space challenges. USP DS Repository \citep{souza2020challenges} proposes real-world relational data streams, however it only explores the distribution drift challenge. Our OEBench considers three categories of open environment challenges in real-world relational data streams and covers a total of 55 datasets, which will be introduced in detail in the following section.}

\begin{table}[htpb]
\caption{Summary of the datasets explored in various incremental learning works.}
\label{tbl:benchwork}
\centering
\newcommand{\y}{\ding{51}}
\newcommand{\n}{\ding{55}}
\resizebox{\columnwidth}{!}{
\begin{tabular}{|c|c|c|c|c|c|c|}
\hline
\multirow{2}{*}{Paper} & \multirow{2}{*}{\#Datasets} & \multirow{2}{*}{Real-world} & Relational & Outliers or & Distribution & Incremental/decremental  \\
& & & data & new classes & drifts & feature space \\ 
\hline
\citep{masana2022class,zenke2017continual,farquhar2018towards, lomonaco2021avalanche,mai2022online} & 16 & \n & \n & \n & \y & \n \\  \hline
\citep{lomonaco2017core50, roady2020stream, she2020openloris} & 3 & \y & \n & \y & \y & \n \\  \hline
\citep{souza2020challenges} & 27 & \y & \y & \n & \y & \n \\  \hline
Ours & \textbf{55} & \y & \y & \y & \y & \y \\
\hline
\end{tabular}
}
\end{table}


\section{Design of OEBench}

\subsection{Design Goals}
\rev{Our OEBench is crafted according to the four benchmark design principles posited by \citet{gray1993benchmark}.}

\rev{\textbf{Relevance}. Our benchmark covers a broad range of open environment challenges including distribution drifts, outliers, and feature space evolution. We have selected 55 real-world relational data streams from diverse sources, including UCI Datasets, Kaggle Datasets, and the USP DS Repository \citep{souza2020challenges}. These datasets cover a variety of contexts on real-time stream learning, including environment surveillance, power consumption forecasting, sales prediction, and fraud detection.}

\rev{\textbf{Simplicity}. With an aim to identify typical open environment patterns and eliminate redundant testing, we choose five representative datasets for our benchmark. We first extract statistics on the three perspectives of the open environment challenges. Then we apply clustering and select the datasets nearest each cluster center to conduct further experiments.}

\rev{\textbf{Portability}. Our benchmark is applicable to new relational data streams. The pipeline for extracting open environment statistics and evaluating stream learning algorithms can be easily invoked or adapted in new systems.}

\rev{\textbf{Scalability}. The benchmark incorporates real-world relational data streams of diverse sizes (Table \ref{tbl:collect}) and varying open environment contexts (Figure \ref{fig:boxplot}). This allows for the simulation of multiple incoming data sizes and evolving patterns. Through this variation, we can test the performance of different algorithms under varying open environment conditions.}

\begin{figure}[ht]
        \centering
        \includegraphics[width=\columnwidth]{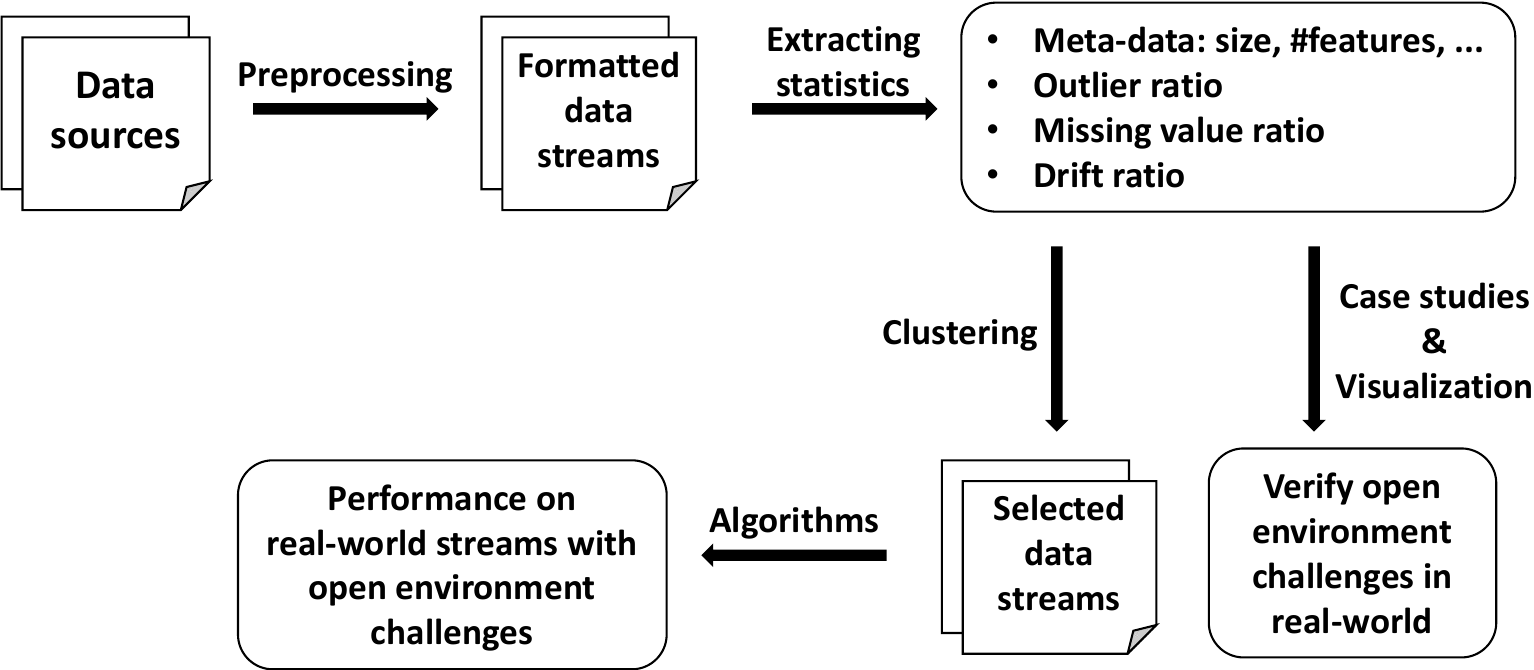}
        \caption{Open environment benchmark flowchart.} 
        \label{fig:flowchart}
\end{figure}

\subsection{Overview}
\rev{Based on the design goals, we build our benchmark as depicted in Figure \ref{fig:flowchart}. It composes of six parts: dataset collection, preprocessing, extracting open environment statistics, visualization, representative dataset selection and incremental learning algorithm evaluation. } 

\rev{\textbf{Dataset collection.} We first search the candidate datasets in public relational dataset resources including the UCI Machine Learning Repository, Kaggle, and the USP DS Repository \citep{souza2020challenges}. Then we keep the datasets with enough rows, columns and meaningful real-world scenarios. We finally keep a total of 55 datasets.}

\rev{\textbf{Preprocessing.} We record dataset metadata, discard textual features, sort the datasets by time, normalize each feature, and set the target and window size to assign a machine learning task for each dataset. This aims to transfer raw datasets into structured relational data streams for further processing.}

\rev{\textbf{Extracting open environment statistics.} We design criteria to evaluate the extent of three open environment challenges in each dataset. Specifically, statistics related to distribution drifts, outliers, and missing values are calculated. Since there are no ground truth label for drifts and outliers in the real-world datasets, we apply ensemble of drift detectors or outlier detectors to estimate their ratios.}

\rev{\textbf{Visualization.} To help data scientists to easily understand the open environment challenges in real-world relational data streams, we conduct visualization based on our extracted open environment statistics. Visualization also serves as a validation of detected drifts and outliers due to the lack of ground truth. Data distributions are plotted in scatter plots to illustrate distribution drifts and outliers. High-dimensional features are reduced to 2-D space by t-SNE \citep{van2008visualizing} for visualization.}

\rev{\textbf{Representative dataset selection.}  As 55 real-world datasets are too many for further experiments, we propose a clustering approach to select a limited number of representative datasets. Specifically, based on the extracted open environment statistics and dataset metadata statistics, we conduct K-means clustering on these information. Inside each cluster, the dataset closest to each center is selected as the representative dataset for the cluster.}

\rev{\textbf{Incremental learning algorithm evaluation.} Based on the selected representative datasets, we evaluate the effectiveness and efficiency of 10 incremental learning algorithms under different open environment scenarios. We also evaluate factors such as window size, batch size, epochs, model size on incremental learning under open environment challenges.}



\begin{table}[htpb]

\centering
\caption{Histogram information of the collected datasets.}
\newcommand{\y}{\ding{51}}
\newcommand{\n}{\ding{55}}
\label{tbl:collect}
\resizebox{\columnwidth}{!}{
\begin{tabular}{|c|c|c|c|c|}
\hline
Size & 5,000-20,000 & 20,001-50,000 & 50,001-200,000 & >200,000 \\ \hline
\#Datasets (USP DS) & 1 & 2 & 4 & 9 \\ \hline
\#Datasets (ours) & \textbf{13} & \textbf{17} & \textbf{13} & \textbf{12} \\ \hline
\hline
\#Features & 5-10 & 11-20 & 21-50 & >50 \\ \hline
\#Datasets (USP DS) & 3 & 0 & 12 & 1 \\ \hline
\#Datasets (ours) & \textbf{15} & \textbf{23} & \textbf{14} & \textbf{3} \\ \hline
\end{tabular}
}
\end{table}

\subsection{Dataset Collection}
\label{chapter:dataset}

In order to thoroughly investigate open environment learning scenarios on real-world relational data streams, we select the datasets with the following criteria.

\begin{itemize}
  \item The sample size must exceed 5,000. 

  \item The features should be relational data streams (excluding text data such as emails) with at least 5 features. The feature dimension should not exceed 1,000 after one-hot encoding. This prevents huge computational burdens.
  
  \item The stream learning scenario should be meaningful. For instance, we exclude the PokerHand dataset as the randomness of each hand undermines meaningful stream analysis.

\end{itemize}

Upon applying these selection criteria to potential datasets from sources including the UCI Machine Learning Repository, Kaggle, and the USP DS Repository \citep{souza2020challenges}, we get a collection of 55 publicly available real-world relational data streams that fulfill all the requirements. \rev{As shown in Table \ref{tbl:collect}, our collection covers a vast range in terms of data size and feature size. Notably, we have over three times the number of real-world relational data streams compared to the qualified datasets in USP DS Repository, which holds the record for the most comprehensive benchmark with its collection of real-world relational data streams. Detailed dataset descriptions are listed in Appendix \ref{sec:datades} of our full version \cite{diao2023oebench}.}

\subsection{Extracting Open Environment Statistics}
\label{sec:preprocess}
Real-world relational data streams display a variety of formats, feature spaces, and scales. To extract open environment statistical features from them, we implement the following preprocessing procedures:

\begin{enumerate}
\item Document dataset metadata, such as task, feature type, target, window size, and null indicators.
\item Order instances by time, then remove time-related attributes to maintain the temporal context without interfering with the dataset's statistical characteristics.
\item Employ one-hot encoding to convert categorical features into numerical format.
\item Utilize KNNImputer to fill in missing values, due to its generally reasonable and bounded predictions. The default value of k is set to 2.
\item Normalize each feature within the dataset.
\item Partition the dataset into non-overlapping windows to facilitate temporal analysis. The size of these windows is determined based on the time span and specific characteristics of each dataset.
\end{enumerate}

For datasets consisting of multiple tables, we focus solely on the main table as integrating multiple tables significantly complicates the analysis. For instance, in the fraud detection dataset \citep{ieee-fraud-detection}, many transaction records lack associated client identity information. Similarly, in loan risk prediction \citep{home-credit-default-risk}, some clients may have no prior loan applications, while others may have multiple application histories. These cases can lead to data items possessing highly varied feature sizes. We leave these complex cases as future works. 

Upon completion of preprocessing, we compute the statistics related to distribution drifts, outliers, and missing values for each dataset, in accordance with the open environment challenges proposed in \citet{zhou2022open}. Both global and per-window statistics are documented. For per-window statistics, we record both the maximum and average values across all windows as open environment statistics of the dataset. 


\paragraph{Distribution Drifts}

We detect data drifts using HDDDM \citep{ditzler2011hellinger}, KdqTree \citep{dasu2006information}, CDBD \citep{lindstrom2013drift}, PCACD \citep{qahtan2015pca} and the Kolmogorov-Smirnov (KS) test. \rev{We apply these methods since they are implemented in an open-source library Menelaus and easy to use. KS statistic is selected because it has statistical guarantee about the drift.} While HDDDM and KdqTree are applicable to multi-dimensional datasets, the remaining three methods must be applied to each dimension separately. We document the results of all methods as open environment statistics. By default, we utilize the first two principal components in PCACD and set $p=0.05$ for the KS test. Elaborations on drift detectors are in Appendix \ref{sec:drift_detection} of our full version \cite{diao2023oebench}. 

For each algorithm, we document the drift and warning percentages. The average and maximum percentages are stored as a global feature for each dataset. For one-dimensional drift detectors, we note the average and maximum percentages across all columns.

Regarding concept drifts, we employ DDM \citep{gama2004learning}, EDDM \citep{baena2006early}, ADWIN accuracy \citep{bifet2007learning}, and PERM \citep{harel2014concept}. \rev{Similarly, the first three algorithms can be called from Menelaus library. PERM \citep{harel2014concept} is additionally selected since it is applicable to regression tasks. Following the examples in Menelaus, we use Gaussian Naive Bayes for classification tasks and linear regression models for regression tasks.} In each window, the predictions are compared with the ground truth to detect concept drifts. Upon the detection of a drift, the model is retrained with the most recent data slices. Similarly, we store the drift and warning percentages as open environment statistics.

\paragraph{Outliers}
\rev{There are a lot of outlier detection algorithms in ADBench \citep{han2022adbench}. We follow the recommendation in ADBench \citep{han2022adbench} to choose ECOD \citep{li2022ecod} and IForest \citep{liu2008isolation} to detect outliers. ECOD \citep{li2022ecod} estimates the underlying distribution of input data to detect outliers in tail probabilities in each dimension. IForest \citep{liu2008isolation} randomly selects an attribute and makes a binary split. The binary tree partition process is conducted recursively to identify easily-isolated data points as outliers.} Within each window, we detect outliers by setting the threshold at three standard deviations above the mean score. We then calculate the average and maximum anomaly ratios within each window as open environment statistics of the dataset.

\paragraph{Missing Values}
We compute the following statistics about missing values.

\begin{enumerate}

\item \textbf{Ratio of data items with missing values}: This measures the completeness of the data items for each row.

\item \textbf{Ratio of missing columns}: This measures the completeness of the feature dimensions for each column.

\item \textbf{Ratio of empty cells}: This measures the completeness of cells within the dataset for the entire table.

\end{enumerate}

\paragraph{\vl{Validation}}
\vl{The statistics for missing values are straightforward. To validate the statistics for drifts and outliers, we use generated datasets with different levels of drifts and outliers. We generate a data stream of 50,000 samples and partition it into 100 windows. By default, each sample contains three dimensions $f_1,f_2,f_3$ sampled uniformly in (0,10).}

\vl{For concept drifts, we use the SEA concept generator \citep{bifet2009new}. We divide the dataset into 4, 8 or 16 blocks with different concepts $f_1+f_2\leq\theta$, with binary classification threshold $\theta\in\{9,8,7,9.5\}$. For 8 blocks, the threshold values are $\{9,8,7,9.5,9,8,7,9.5\}$, and similarly for 16 blocks, we repeat the threshold values for four times. Our average concept drift statistics for 4, 8, 16 different concepts are 0.000114, 0.000229, 0.000263 respectively, which complies with the frequency of concept drifts.}

\vl{For data drifts, similarly, we divide the dataset into 4, 8 or 16 blocks, each with features generated in $(0,\phi)$ where $\phi \in \{10,9,11,7\}$. Our average data drift statistics for 4, 8, 16 different blocks are 0.289, 0.322, 0.356 respectively, which is in accordance with the frequency of data drift. }

\vl{For outliers, we randomly change 400, 800 or 1600 data samples to outliers in the stream, by randomly adding or deducting 20 in their feature values. Our average anomaly statistics for 400, 800, 1600 outliers are 0.0088, 0.0161, 0.0320 respectively, which conforms to the expected order. The above experiments verify that our extracted statistics are reasonable. }

\subsection{Selection of Representative Datasets}
\label{sec:representative}

\rev{Since it is expensive to evaluate existing algorithms on all collected 55 datasets, we pick up representative datasets for further exploration. To select representative datasets, we first normalize all open environment statistics and dataset metadata statistics to the same scale. Then for each category of statistics (dataset basic information, missing values, data drifts, concept drifts, outliers), we transform them into 3-D space using Principal Component Analysis (PCA). This transformation ensures that each category is represented with the same number of statistics. We then apply K-means clustering to partition the datasets into five distinct clusters to group the datasets with similar characteristics. The datasets nearest to each cluster center are selected as representative datasets. }

\begin{figure}
        \centering
        \includegraphics[width=\columnwidth]{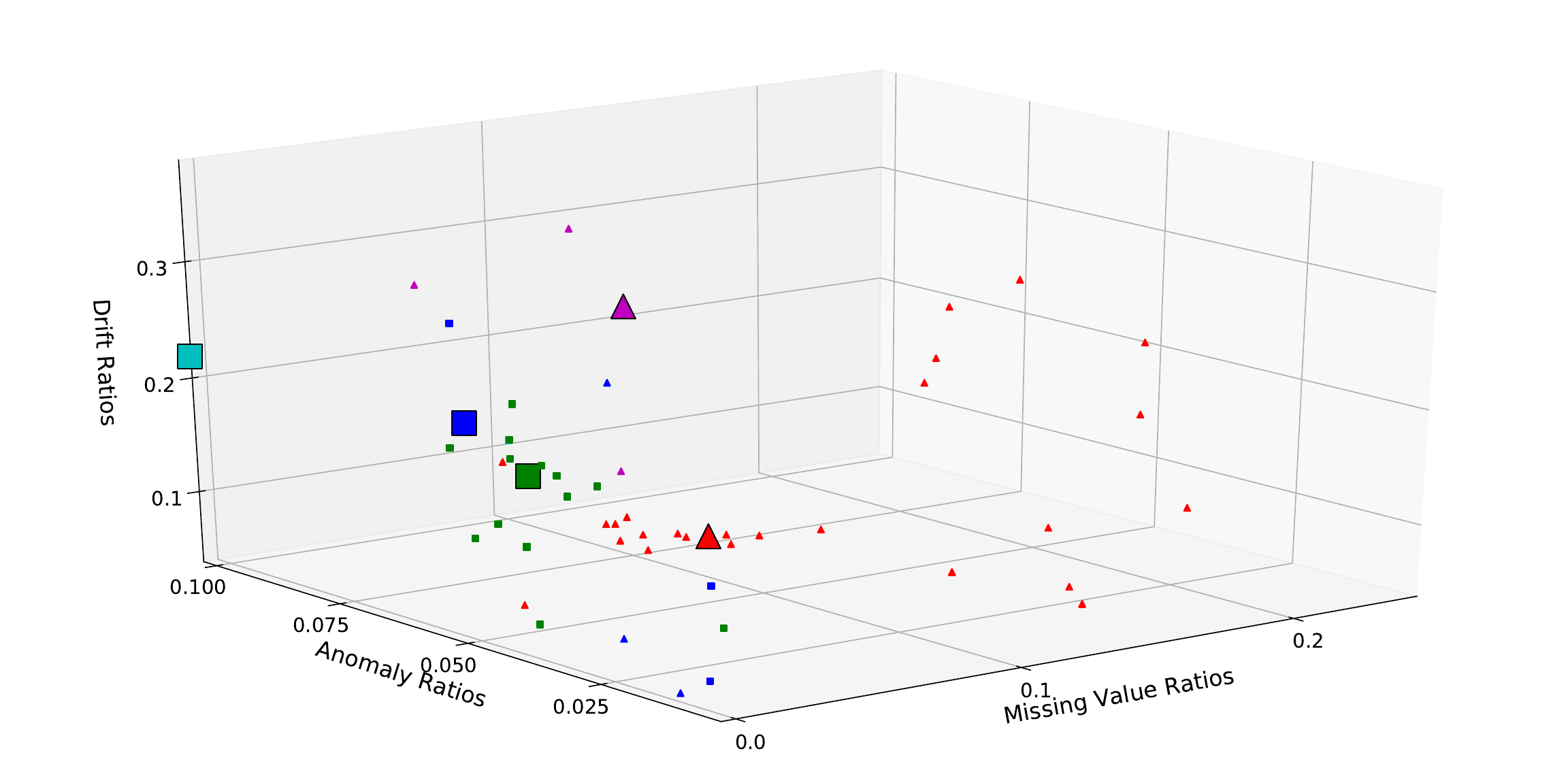}
        \caption{Visualization of clustering results on three open environment dimensions. Large points are selected datasets. Square represents classification task and triangle represents regression task. } 
        \label{fig:clustershow}
\end{figure}

\rev{Figure \ref{fig:clustershow} visualizes the clustering results on three open environment dimensions. Missing value ratio can be directly calculated by counting the ratio of missing items. Drift ratio is calculated by the ratio of both data drifts and concept drifts detected between consecutive two windows. Anomaly ratio is calculated by the ratio of detected anomalies among each window. Both drift and anomaly do not have ground truth and are calculated through ensemble of detectors.} 

\rev{As we can see, red, purple and cyan represent high missing value ratio, high drift ratio and high anomaly ratio respectively. Blue and green seems mixed, but blue represents datasets of commercial field while green represents datasets of nature science field, since we also take the task, field, dataset size into consideration when clustering. Table \ref{tbl:select} provides details of the five selected datasets, each showcasing unique characteristics. }

\vl{We use the results of K-means clustering due to its popularity and simplicity. Besides K-means, we also try other clustering algorithms including spectral clustering \citep{von2007tutorial}, agglomerative clustering \citep{mullner2011modern}, and Gaussian mixture model (GMM) \citep{reynolds2009gaussian}. The number of clusters remains five. We find out that Beijing Air Quality (Shunyi), Room Occupancy and Electricity Prices appear in the selected datasets by at least three out of four clustering algorithms. Besides, the results of all four algorithms contain one of the INSECTS dataset. The consensus of different clustering algorithms indicate that those datasets are representative. }

\begin{figure}[ht]
        \centering
        \includegraphics[width=\columnwidth]{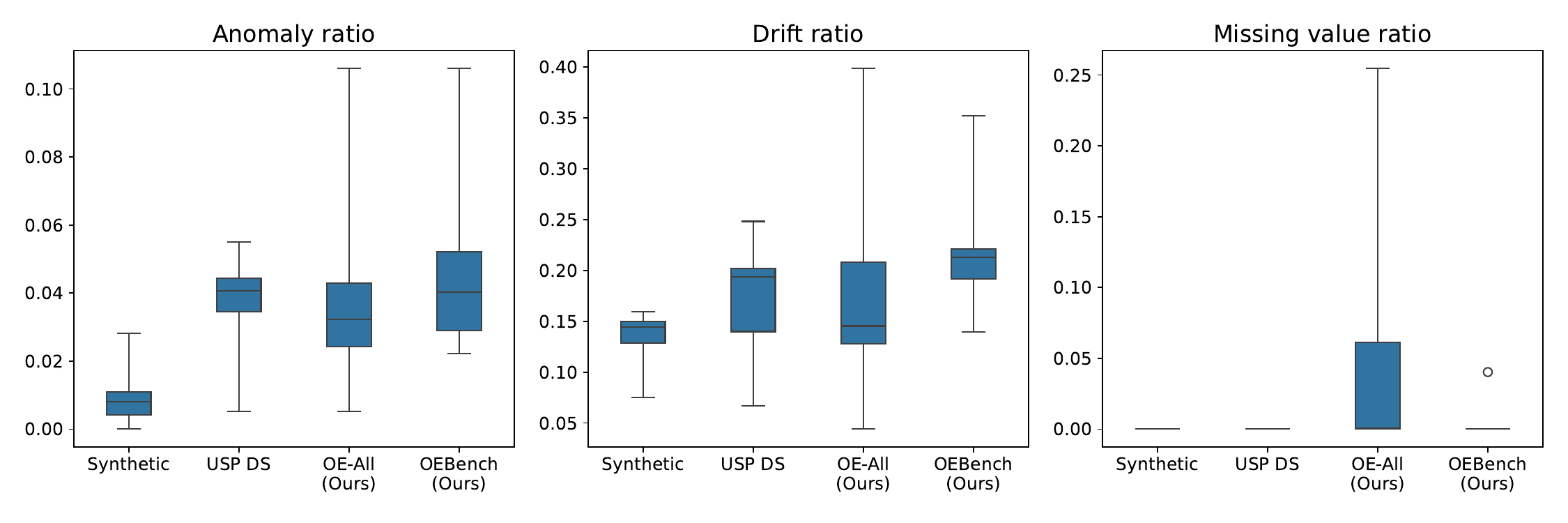}
        \caption{\vl{Statistical distribution of synthetic datasets, USP DS Repository and our datasets. We denote all 55 datasets as \textit{OE-All} and the selected five datasets as \textit{OEBench}. In missing value ratio, 1 out of the 5 datasets in OEBench contains missing values, which is the individual circle in the box plot.}} 
        \label{fig:boxplot}
\end{figure}

\vl{Figure \ref{fig:boxplot} presents a box plot of open environment statistics from four data sources: six popular synthetic datasets (SEA, STAGGER, Rotating Hyperplane, RBF, LED, and Waveform) in prior stream learning works \citep{bifet2009new}, the USP DS Repository \citep{souza2020challenges}, our collected datasets, and our selected representative datasets. As shown in Figure \ref{fig:boxplot}, our 55 collected datasets OE-All cover a much broader spectrum of open environment statistics than prior synthetic datasets and the real-world benchmark USP DS Repository. It is particularly evident in the aspect of missing values, where synthetic datasets and USP DS Repository do not explore. Further, OEBench representative datasets emulate the distribution of the 55 OE-All datasets, which also span a diverse range of open environment scenarios. }

\begin{table*}[htpb]

\centering
\caption{Five selected representative datasets. The levels of three open environment challenges are determined by the extracted statistics compared with the average statistics of all 55 datasets.}
\newcommand{\y}{\ding{51}}
\newcommand{\n}{\ding{55}}
\label{tbl:select}
\resizebox{2.1\columnwidth}{!}{
\begin{tabular}{|c|c|c|c|c|c|c|c|c|c|}
\hline
Dataset & Default window size & Instances & Features & Type & Task & Missing value ratio & Drift ratio & Anomaly ratio \\ 
\hline
Room Occupancy Estimation & 2 hours & 10,129 & 6 & Others & Classification & Low & Medium high & High \\ \hline
Electricity Prices & 2 weeks & 45,312 & 7 & Commerce & Classification & Low & Medium high & Medium high   \\ \hline
INSECTS-incremental-reoccurring (balanced) & 100 items & 79,986 & 33 & S\&T & Classification & Low & Medium low & Medium high  \\ \hline
Beijing Multi-Site Air-Quality Shunyi & 30 days & 35,064 & 11 & Ecology & Regression & High & Low & Medium low  \\ \hline
Power Consumption of Tetouan City & 15 days & 52,417 & 7 & Power & Regression & Low & High & Medium low \\

\hline
\end{tabular}
}
\end{table*}

\subsection{Selected Incremental Learning Algorithms}
\rev{We select representative incremental learning algorithms from the perspectives of regularization and experience replay, as introduced in Section \ref{sec:incemental}. Besides, we also explore tree-based and ensemble-based algorithms for real-world relational data stream learning.}

\textbf{Regularization.} We implement Elastic Weight Consolidation (EWC) \citep{kirkpatrick2017overcoming} and Learning without Forgetting (LwF) \citep{li2017learning}, considering their popularity and ease of implementation. 

\textbf{Experience replay.} We implement iCaRL \citep{rebuffi2017icarl} since it is popular and can be extended to infinite data streams. Various rehearsal-based or isolation-based methods \citep{hou2019learning, yan2021dynamically, wu2019large, mallya2018packnet, serra2018overcoming} are not suitable for real-world data streams, particularly for regression tasks. Real-world streams can be infinite, making it unrealistic to expand network architecture or isolate neurons.

\textbf{Tree model.} We apply the Adaptive Random Forest (ARF) \citep{gomes2017adaptive}. In this approach, each tree is trained with a subset of features and is subjected to drift detection. After drift is identified, a background tree is trained to replace the current tree.

\textbf{Ensemble.} We choose the Streaming Ensemble Algorithm (SEA) \citep{street2001streaming}. SEA maintains an ensemble and replaces older models with current models of better quality.

\section{Case Studies of Open Environment Scenarios}
\label{sec:visual}
In this section, we visualize and analyze some real-world relational data streams. Through these cases, we study how the open environment challenges look like in real-world relational data streams, and provide a deeper understanding of their implications.

\subsection{Distribution Drifts}

\begin{figure}[htbp]
\includegraphics[width=\columnwidth]{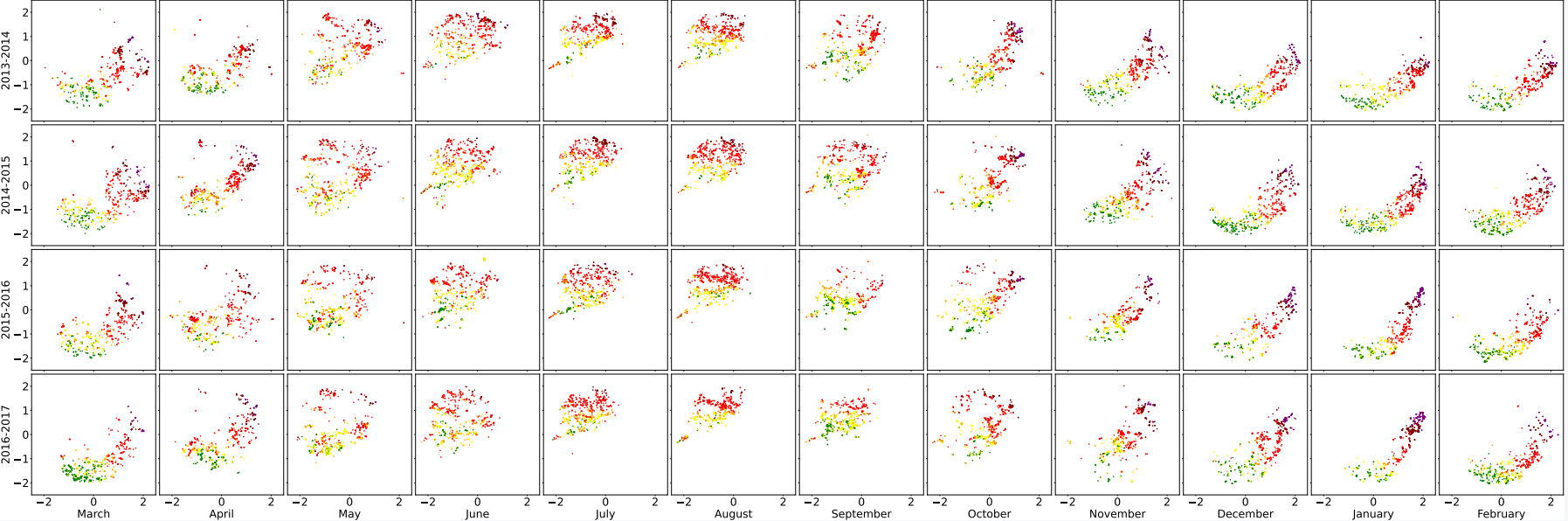}
\caption{T-SNE visualization of air quality dataset in Tiantan, Beijing. Each line spans a year and each sub-figure represents a month. Data items are labelled according to 6 categories of AQI based on the severity of health concern. }
\label{fig:tiantan}
\end{figure}

Distribution drifts can be classified into two categories: data drifts and concept drifts. Data drifts refer to changes in the distribution of the feature set. Given the dynamic nature of most real-world data, such drifts are common. Figure \ref{fig:tiantan} provides a clear illustration of cyclical data drift of air quality surveillance, occurring approximately on an annual basis. This cyclical pattern aligns with the inherent seasonality of the dataset. Similar patterns are observed across other air quality datasets as well.

On the other hand, concept drifts imply changes in the environment, resulting in an evolving relationship between input features and output labels. It can also be found in Figure \ref{fig:tiantan}. For example, the decision boundaries of July are different from that of March, probably due to different weathers in different seasons. 

\paragraph{Impact of distribution drifts}
In Figure \ref{fig:tiantan}, we observe that distribution drifts happen around window $7,11,19,23,32,35$. To investigate the impact of drifts on incremental learning on real-world data streams, we train a decision tree on (1) the first 11 windows, and (2) window 7 to 11 of the air quality dataset in Tiantan. Both models are tested on the next window 12. The test loss of training on all first 11 windows is 0.347, while only 0.299 for training on window 7 to 11. Considering the distribution drifts at around window 7, we can conclude that including historical data with different distributions harms the model accuracy towards new data distributions, since memorizing old data cannot well adapt to the new environment. 

\begin{figure}[htbp]
\includegraphics[width=0.7\columnwidth]{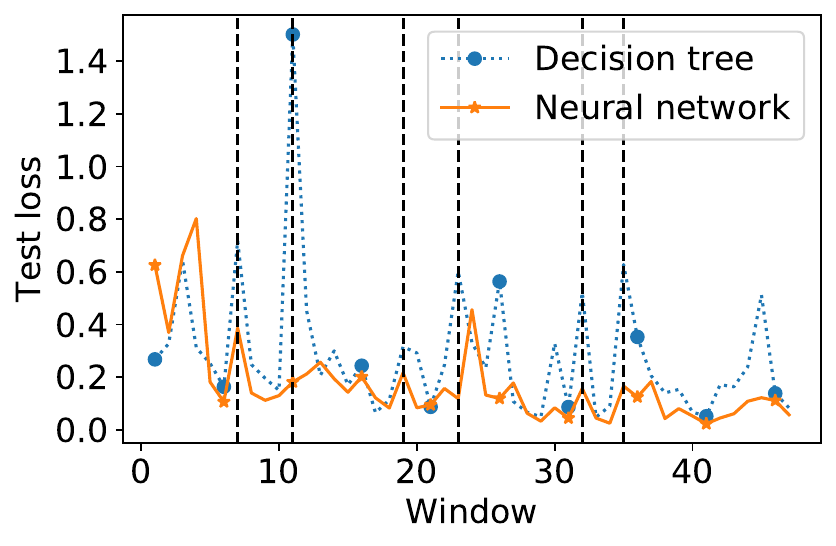}
\caption{Test loss of training a decision tree and a neural network on Beijing air quality dataset, Tiantan. Vertical lines denote windows around the happening of distribution drifts.}
\label{fig:drift_example}
\end{figure}

\rev{We also train a decision tree and a neural network on the air quality dataset in Tiantan. The test loss is shown in Figure \ref{fig:drift_example}. We mark the windows around drift occurrences by vertical lines, where we can clearly see the sudden increase of test loss. Therefore, distribution drifts bring great challenges to the model generalization in data streams. Under distribution drifts, old knowledge from the past data can have a negative impact for models to better adapt to the new environment. Such scenarios contradict with the goal of prior incremental learning works to perform well on all seen data. This implies that prior incremental learning algorithms may not work well on real-world open environment learning scenarios. }

\subsection{Outliers}
\label{sec:example_outliers}
We conduct case study on the Beijing PM2.5 dataset (from five Chinese cities PM2.5 dataset). Our analysis of outliers focuses on two extreme weather events in Beijing: the flood of July 21, 2012, and the period of intense haze from 2014 to 2015. By utilizing scatter plot visualization, we are able to identify anomalous data points.

We apply iForest \citep{liu2008isolation} to detect the outliers associated with the 2012 Beijing flood. This flood event, which led to the highest level of precipitation in the city since 1951, constitutes one of the most extreme weather in recent history.

As for the period of extreme haze that spanned from November 2014 to February 2015, we use ECOD \citep{li2022ecod} to visualize the outliers detected in the dataset. During this period, Beijing experienced record-breaking levels of PM2.5 pollution, peaking at 700 in certain regions.

\begin{figure}[h]
    \centering
    \subfloat[Beijing PM2.5 (2014 - 2015)]{\label{fig:hazeanomaly}\includegraphics[width=0.5\columnwidth]{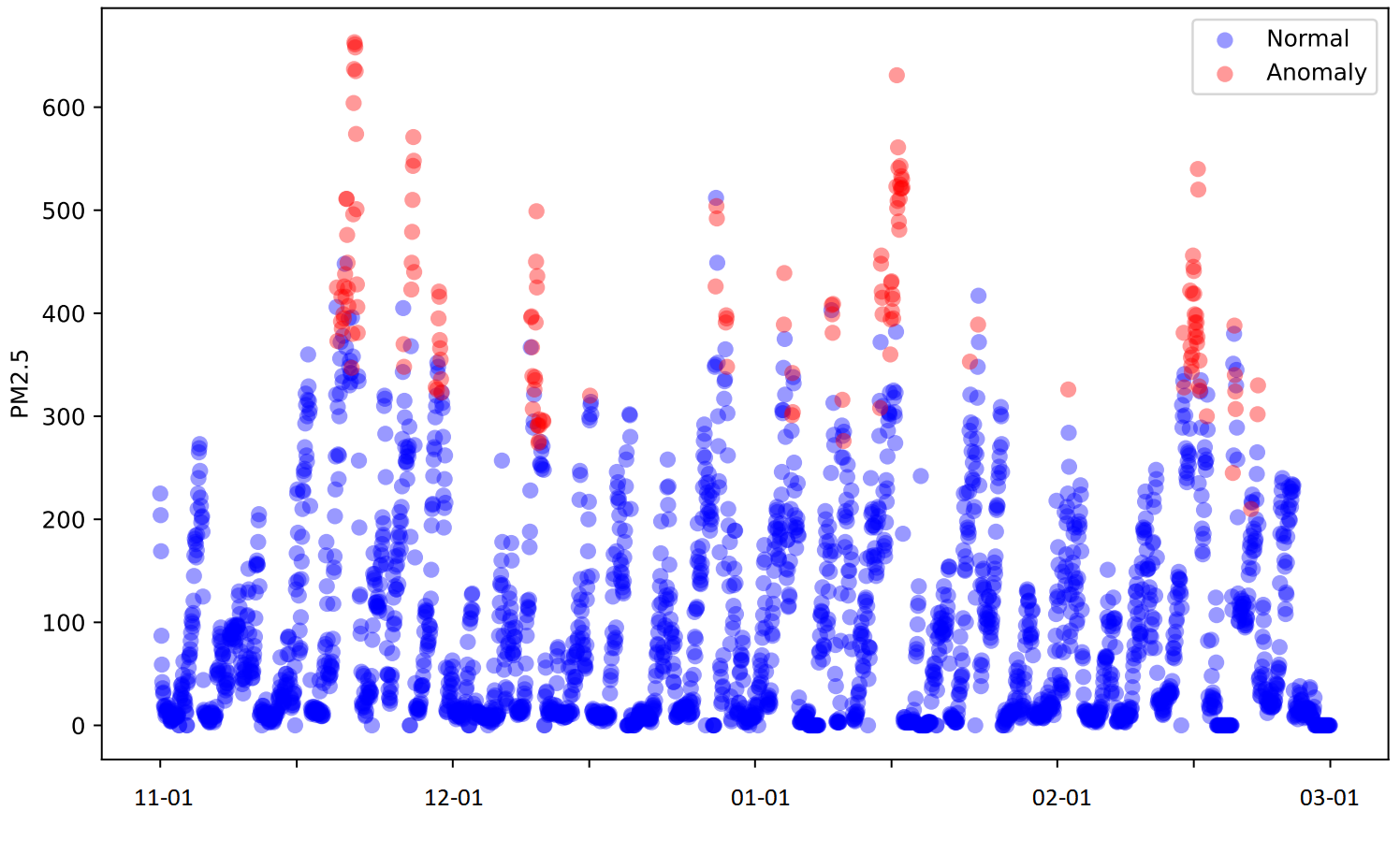}}
    \subfloat[Precipitation in Beijing 2012 flood]{\label{fig:rain}\includegraphics[width=0.5\columnwidth]{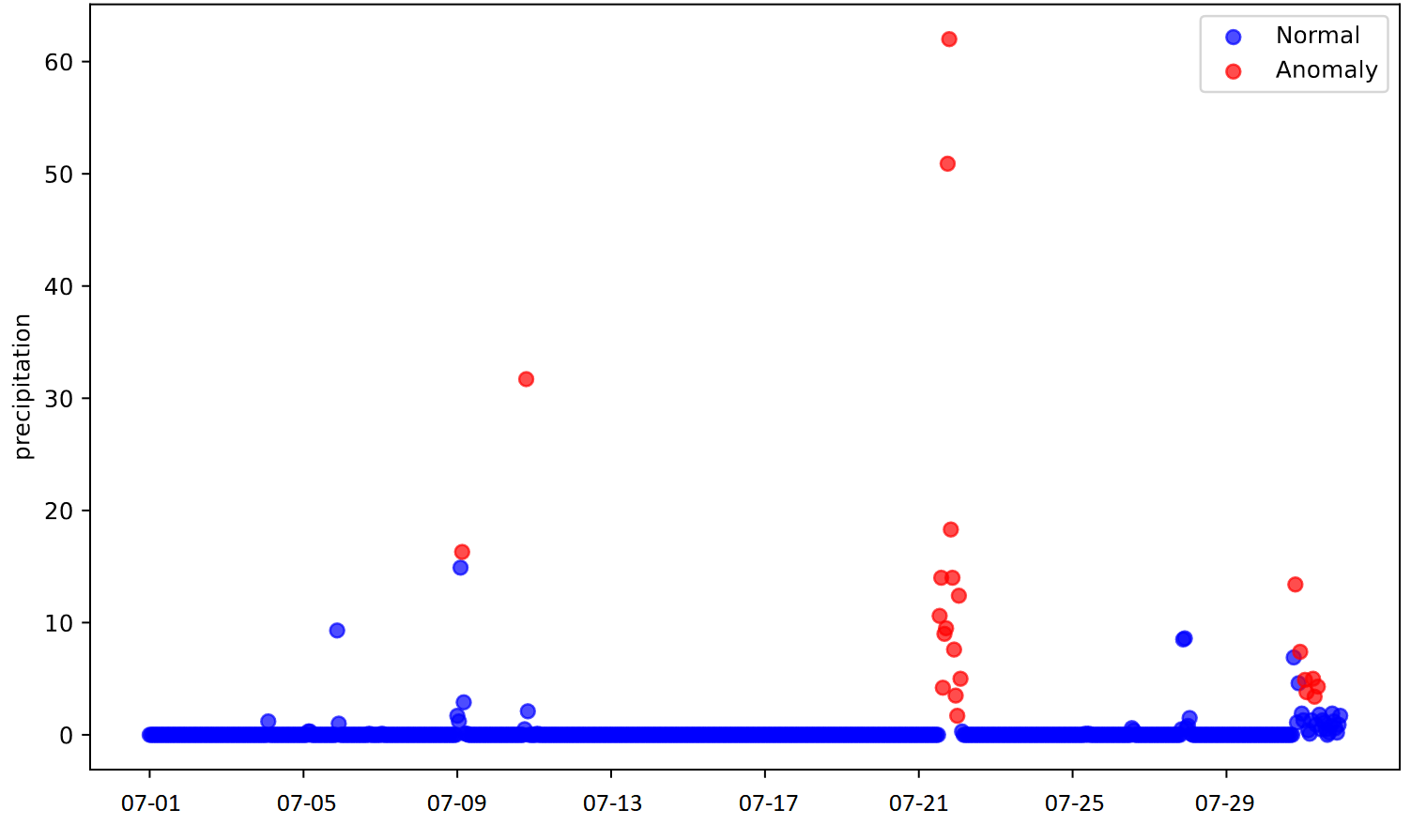}}
    \caption{Visualization of detected outliers.}
    
\end{figure}

Both iForest and ECOD are applied to these datasets. Interestingly, they yield similar outcomes, especially in identifying particularly unusual events such as the Beijing flood and the intense haze period. Thus, we present the visualization of one algorithm for each event.
Figures \ref{fig:hazeanomaly} (Beijing heavy haze from Nov 2014 to Feb 2015) and \ref{fig:rain} (Beijing 2012 July Flood) reveal a clear correlation between the detected outliers and these events. The outliers align with the temporal spans of the Beijing flood and the PM2.5 crisis, verifying the effectiveness of the anomaly detection methods in recognizing abnormal patterns.

\begin{figure}[h]
\centering
\includegraphics[width=0.7\columnwidth]{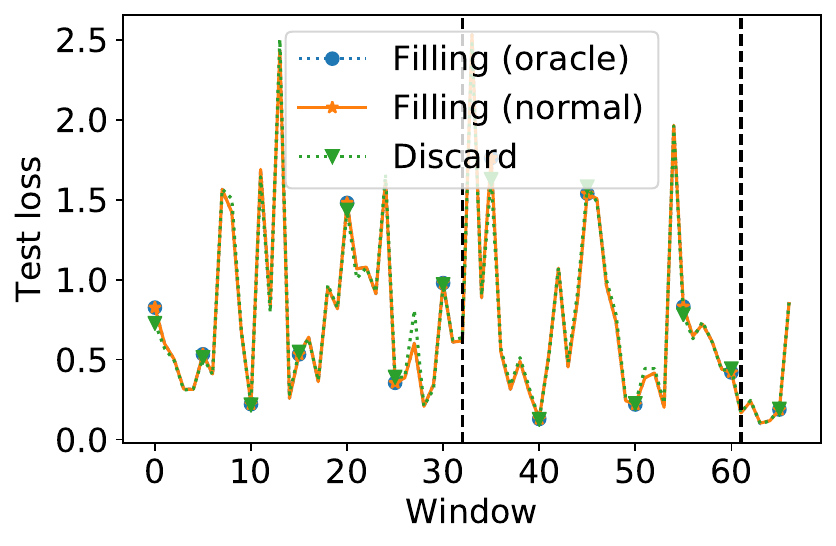}
\caption{Test loss of training a neural network on the Beijing PM2.5 Data of Five Chinese Cities dataset. Vertical lines are extreme weather events. ``Filling (oracle)'' means filling up missing values with the knowledge of the whole dataset. ``Filling (normal)'' means filling only with the knowledge of data in its current window. ``Discard'' means to discard the three frequently missing features.}
\label{fig:discard-acc}
\end{figure}

\rev{Outliers can bring great challenges to data stream learning. In Figure \ref{fig:discard-acc}, we locate the mentioned extreme weather situations with vertical lines. As we can see, after the abnormal situations, the model test loss increases. Especially in Beijing 2012 flood incident where the outliers happen in an abrupt way, we can witness a spike in the test loss (after the left vertical line). As machine learning models are data-driven, outliers may significantly harm the model since it can greatly affect the loss.}

\rev{The most extreme case in this dataset is the No. 51,278 data, where the precipitation becomes 999,990. The normal range of precipitation is 0-100. We are unable to plot this problem in Figure \ref{fig:discard-acc} since the test loss of the neural network suddenly becomes very large or even infinite in the following windows. It illustrates the vulnerability of neural networks when encountering outliers. On the decision tree, we also observe a 3x spike of the test loss after the No. 51,278 data, but it does not crash. Even a single unreliable data item can corrupt the model. Thus, in open environment scenarios, it is an important and challenging problem to detect outliers to reduce their harm to stream learning models. }



\subsection{Incremental/Decremental Features}
\label{sec:inc/dec}
Incremental or decremental features is widespread in real-world relational data streams, especially in ecology datasets where sensors are deployed to collect data. For instance, all the air quality datasets in our study have incremental or decremental features, due to the malfunction, repair, removal, or installation of sensors.

\begin{figure}[h]
\centering
\includegraphics[width=0.8\columnwidth]{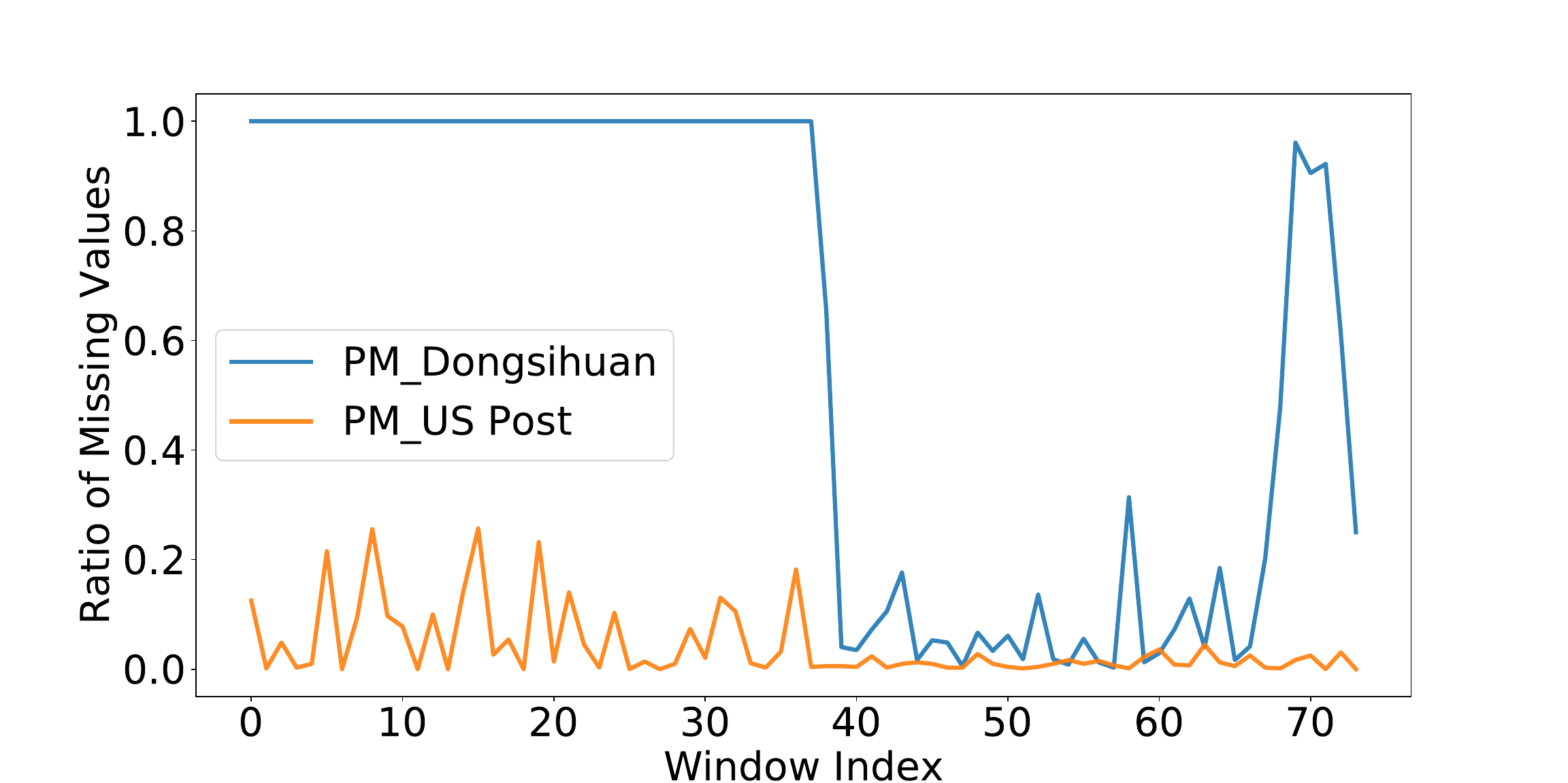}
\caption{Ratio of missing values per window in the air quality data in Beijing, PM2.5 Data of Five Chinese Cities dataset. }
\label{fig:beijing}
\end{figure}

\rev{Figure \ref{fig:beijing} shows the ratio of missing values of two features in the Beijing PM2.5 dataset (from five Chinese cities PM2.5 dataset). The appearance of filled missing values is associated with an incremental feature space, while missing values in an attribute correspond to a decremental feature space. }

\rev{Prior missing value filling studies \citep{emmanuel2021survey, pujianto2019k} mainly focus on cases like the orange line, where the model knows the existence of the feature and the ratio of missing values is small. However, open environment challenges include scenarios like the blue line. In the first 40 windows, the model is unaware of the existence of such feature. The sudden appearance of the feature leads to incremental feature space, which is rarely considered in prior works. Such scenarios can result from installing new sensors. This is more challenging due to the unpredictable changes. }

\rev{To further explore the impact of incremental and decremental features on data stream learning, we train a neural network on the dataset with different methods of processing the incremental and decremental features. The setups are elaborated in Section \ref{sec:exp_setup}. As shown in Figure \ref{fig:discard-acc}, we observe an interesting phenomenon that discarding these always-missing features have similar test loss compared with filling them up, which shows that more data does not necessarily lead to better model accuracy in some real-world data streams. }

\rev{In reality, if a sensor always fails to record values, its recorded values are more likely to be unreliable. Unreliable data can harm the model accuracy, which is also discussed in Section \ref{sec:example_outliers}. Due to these complex scenarios in real-world data streams, open environment machine learning tasks are quite challenging.}

\section{Experiments}


\subsection{Setups}
\label{sec:exp_setup}
\textbf{Models.} We adopt a multi-layer perceptron (MLP) as our default neural network (NN) architecture, which includes three hidden layers consisting of 32, 16, and 8 neurons respectively with ReLu activation. For tree-based models, we adopt decision tree or GBDT with an ensemble of five trees by default. \vl{For advanced models on relational datasets, we test TabNet \citep{arik2021tabnet} and ARM-Net \citep{cai2021arm}. }

\noindent\textbf{Hyper-parameters.} For each window, we train NN for 10 local epochs with batch size 64 and learning rate 0.01. The buffer size is set to 100. For EWC, we observe that regularization factors below $10^3$ yield results akin to naive NN, while factors exceeding $10^5$ can lead to loss explosions. Therefore, we adjust the regularization factor in $\{10^3,10^4,10^5\}$. For LwF, since the regularization loss is in the similar order of magnitude with the naive NN loss, we tune the regularization factor in $\{0.01,0.1,1\}$. Missing values are handled by employing the KNNImputer with $k=2$. We use an ensemble of five models by default for ARF and SEA. Decision tree-based methods, unlike neural networks, do not need multiple passes or batches of data for training.

\noindent\textbf{Algorithm implementations.} Given that real-world data streams may be infinite, it is unfeasible to keep models from all windows in EWC and LwF. As an alternative, we employ the model from the most recent window to perform regularization. Additionally, LwF and iCaRL are originally designed for classification tasks. To adapt them to regression tasks, we substitute the LwF regularizer with mean square error (MSE) loss and consider all data items as a single class for iCaRL. To normalize each feature, we use the mean and variance of the first window to rescale each dataset dimension. The purpose is to simulate the real-world scenarios where only the statistics of early samples are available to get started. 

\noindent\textbf{Metrics.} Our evaluation methodology employs a test-then-train paradigm within each window. Aside from the initial warm-up window, we test the model on the data of each window, followed by training on these data to update the model. We calculate the error rate for classification tasks and MSE loss for regression tasks. The final error rate or MSE loss is determined by averaging the results across all windows.

\noindent\textbf{Hardware setups.} All experiments run on a machine with 64 Intel(R) Xeon(R) Gold 6226R CPU @ 2.90GHz, 376 GB memory, and we use 2.7 TB hard disk as storage. The OS is Ubuntu 22.04.3 LTS.


\subsection{Evaluating Existing Algorithms on Real-World Relational Data Streams}
\label{sec:overall}

\begin{table*}[h]
\centering
\caption{Test loss / test error of stream learning algorithms on different characters of real-world datasets. Lower value indicates better result. We repeat all experiments for three times with different random seeds.}
\label{tbl:main}
\resizebox{2.1\columnwidth}{!}{
\begin{tabular}{|c|c|c|c|c|c|c|c|c|c||c|c|c|c|c|c|}
\hline
Task & Drift & Anomaly & Missing value & Dataset & Naive-NN & EWC & LwF & iCaRL & SEA-NN & Naive-DT & Naive-GBDT & SEA-DT & SEA-GBDT & ARF \\ \hline
Classification & Medium high & High & Low & ROOM & 0.214$\pm$0.004 & 0.207$\pm$0.003 & 0.207$\pm$0.014 & \textbf{0.136$\pm$0.021} & 0.207$\pm$0.037 & 0.198$\pm$0.006 & 0.181$\pm$0.004 &0.191$\pm$0.004 & \textbf{0.151$\pm$0.002} & 0.250$\pm$0.004 \\ \hline
Classification & Medium high & Medium high & Low & ELECTRICITY & 0.311$\pm$0.012 & 0.311$\pm$0.012 & 0.311$\pm$0.012 & \textbf{0.286$\pm$0.013} & 0.332$\pm$0.022 & 0.272$\pm$0.001 & 0.256$\pm$0.000 & 0.263$\pm$0.009 & 0.264$\pm$0.001 & \textbf{0.250$\pm$0.002} \\ \hline
Classification & Medium low & Medium high & Low & INSECTS & \textbf{0.269$\pm$0.006} & \textbf{0.269$\pm$0.006} & \textbf{0.269$\pm$0.006} & 0.306$\pm$0.005 & 0.321$\pm$0.007 & 0.329$\pm$0.001 & 0.306$\pm$0.000 & \textbf{0.291$\pm$0.004} & \textbf{0.291$\pm$0.002} & 0.294$\pm$0.001 \\ \hline

Regression & Low & Medium low & High & AIR & \textbf{0.166$\pm$0.002} & \textbf{0.166$\pm$0.002} & \textbf{0.166$\pm$0.002} & 0.182$\pm$0.008 & 0.213$\pm$0.023 & 0.263$\pm$0.013 & 0.498$\pm$0.002 & \textbf{0.199$\pm$0.010} & 0.519$\pm$0.003 & N/A \\ \hline
Regression & High & Medium low & Low & POWER & 0.793$\pm$0.005 & 0.794$\pm$0.004 & \textbf{0.779$\pm$0.004} & 0.818$\pm$0.014 & 0.783$\pm$0.015 & 1.278$\pm$0.003 & \textbf{0.800$\pm$0.000} & 0.845$\pm$0.007 & 0.835$\pm$0.002 & N/A \\ \hline

\end{tabular}
 }
\end{table*}


\noindent
\begin{tcolorbox}[width=\linewidth,colback=white,boxrule=1pt,arc=0pt,outer arc=0pt,left=0pt,right=0pt,top=0pt,bottom=0pt,boxsep=1pt,halign=left]
\rev{\textbf{Finding (1):} There is no silver bullet that performs well in all cases of open environment challenges. Therefore, open environment challenges are difficult to deal with and previous algorithms are not specifically designed to address them. }
\end{tcolorbox}

\begin{figure*}[ht]
    \centering
    \subfloat[NN vs tree]{\includegraphics[width=0.7\columnwidth]{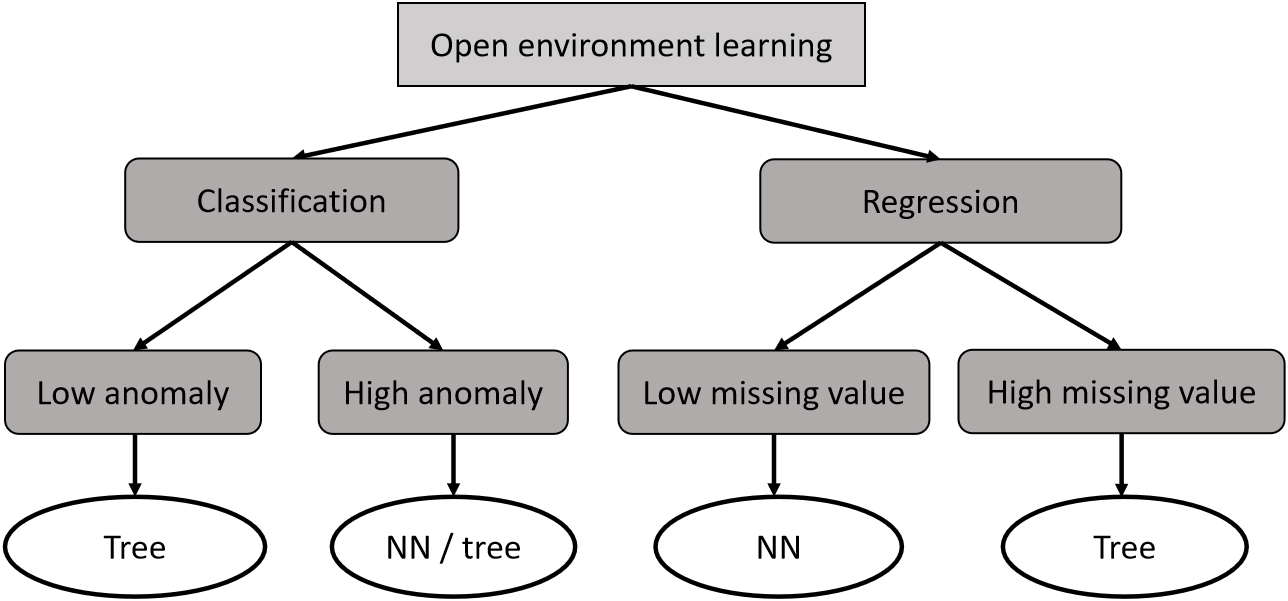}}
    \subfloat[NN models]{\includegraphics[width=0.7\columnwidth]{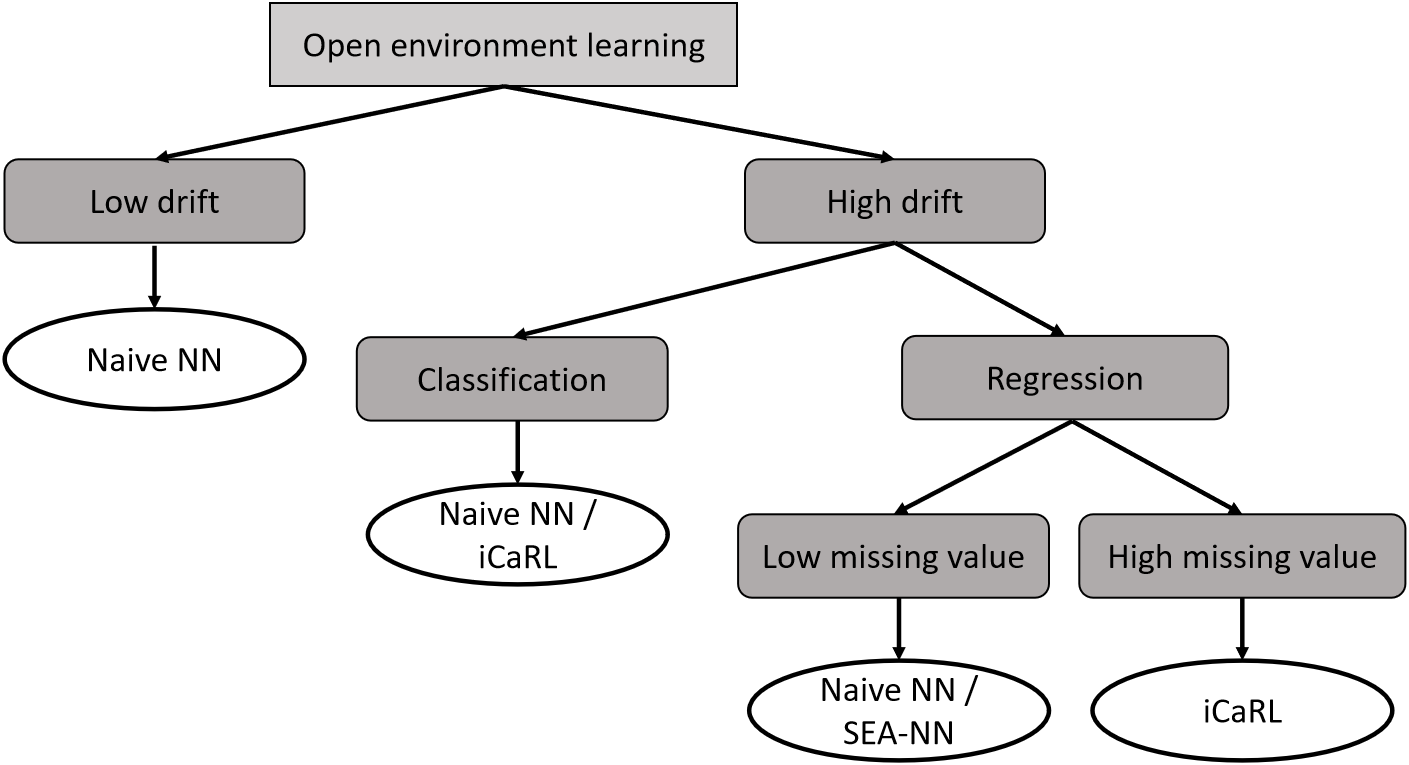}}
    \subfloat[Tree models]{\includegraphics[width=0.7\columnwidth]{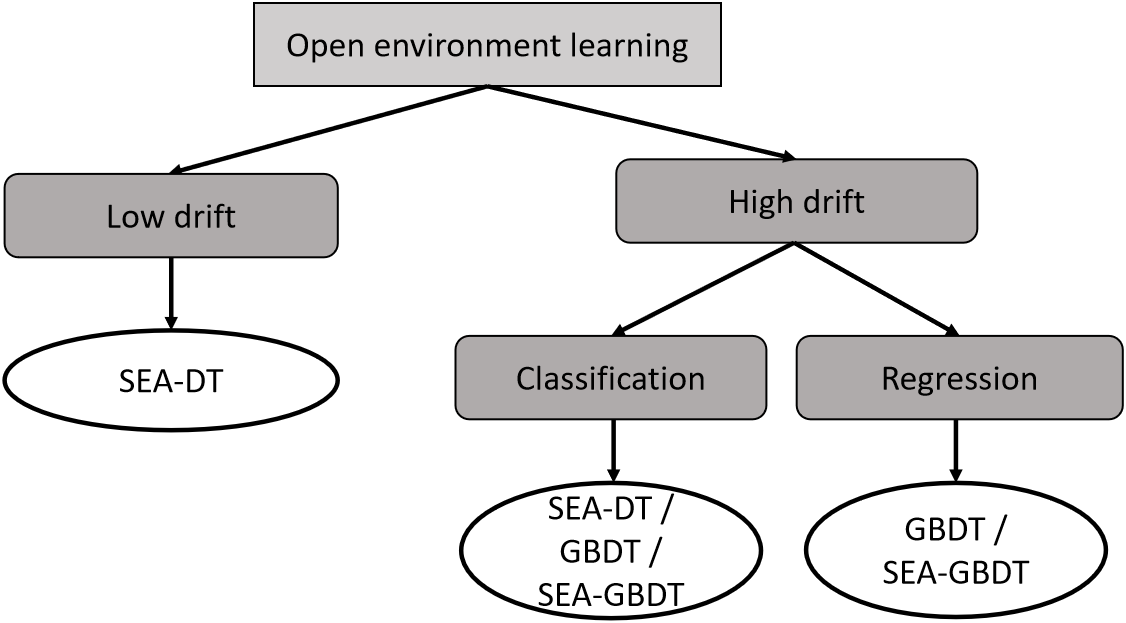}}    
    \caption{Recommendations of the (almost) best algorithm in different cases of open environment learning.}
    \label{fig:recom}
\end{figure*}

\rev{We compare the model accuracy of existing algorithms on real-world relational data streams as shown in Table \ref{tbl:main}. Besides the five representative datasets, we also experiment on the other 50 datasets of our benchmark to enhance our assessment of the strengths of these algorithms. The full results are shown in Appendix \ref{sec:all55} of our full version \cite{diao2023oebench}. From the results, it is evident that no single algorithm consistently outperforms the others across all cases of open environment challenges. Based on the results of all 55 datasets, we synthesize our recommendations for different contexts into a decision tree, which can be viewed in Figure \ref{fig:recom}.}

\rev{In terms of effectiveness, tree models are generally recommended in classification tasks with low anomalies and regression tasks with high missing values. NN models are recommended in regression tasks with low missing values. Another work \citep{shwartz2022tabular} also verifies that tree models work better than NN models in classification tasks on tabular datasets. NN models are generally over-parameterized than tree models, which results in NN models prone to overfitting in simple tasks like tabular dataset classification. Regression tasks generally require the model to learn more complex functions compared with learning decision boundaries in classification tasks, which is probably the reason why over-parameterized NN models have better accuracy in regression tasks with low missing values. However, NN models are sensitive to the variations of input data \cite{goodfellow2014explaining, kurakin2016adversarial}, therefore tree models are recommended if there are many missing values. When time or memory constraints are tight, tree models are better suited due to their efficiency, as discussed in the following section. }

\rev{Among NN-based methods, naive NN and iCaRL typically outperform other NN-based techniques in classification tasks with high drifts. ICaRL is designed for classification tasks to store exemplars of each class, and it has the advantage of alleviating forgetting when the drift is relatively high. It also works well on regression tasks with high missing values. For data streams with relatively low drifts, we recommend naive NN. For regression tasks with low missing values, naive NN and SEA-NN tend to outperform other methods.}

\rev{Among tree-based methods, GBDT and SEA-GBDT yield the best accuracy under high drifts. When the drifts are relatively low, SEA-DT is recommended. SEA-DT also works well in classification tasks with high drifts. }

\subsection{Time and Memory Consumption}
\rev{On efficiency, we compare the throughput in Table \ref{tbl:throughput} and memory consumption in Table \ref{tbl:memory}. Decision trees have much higher throughput and lower memory costs than NN-based methods. ARF is very bad since it detects drifts in the background, which takes too much computation and memory. Therefore, we recommend to use DT or GBDT when time or memory constraints are tight. }


\rev{When considering both effectiveness and efficiency, EWC, LwF and ARF can be excluded as suitable choices for these explored real-world data streams. EWC and LwF have marginal improvement on training a naive NN while doubling the computational costs, which illustrates that simply applying incremental learning algorithms does not necessarily work well in open environment learning tasks. ARF incurs significantly longer computation time, ranging from 30 to 1,000 times longer than other tree-based algorithms, without delivering a significant boost in accuracy. Therefore, these methods are not well-suited to real-world relational data streams.}

\begin{table*}[h]
\centering
\caption{Throughput of explored stream learning algorithms on selected real-world datasets. All experiments run on 4 CPU threads. For NN-based methods, we set the default number of epochs as 10. }
\label{tbl:throughput}
\resizebox{1.8\columnwidth}{!}{
\begin{tabular}{|c|c|c|c|c|c||c|c|c|c|c|c|}
\hline
Dataset & Naive-NN & EWC & LwF & iCaRL & SEA-NN & Naive-DT & Naive-GBDT & ARF & SEA-DT & SEA-GBDT \\ \hline
ROOM & 8,168 & 3,376 & 4,646 & 4,185 & 7,083 & \textbf{202,580} & 44,039 & 234 & 27,375 & 27,375 \\ \hline
ELECTRICITY & 8,375 & 3,437 & 4,800 & 4,572 & 7,501 & \textbf{266,541} & 71,923 & 320 & 50,912 & 28,814 \\ \hline
INSECTS & 6,124 & 2,630 & 3,692 & 3,372 & 5,052 & \textbf{55,545} & 3,146 & 44 & 7,180 & 1,713 \\ \hline
AIR & 7,861 & 3,770 & 5,118 & 4,466 & 7,381 & \textbf{48,032} & 32,466 & N/A & 20,871 & 15,311 \\ \hline
POWER & 9,260 & 4,973 & 6,315 & 5,747 & 9,006 & \textbf{169,087} & 134,402 & N/A & 119,129 & 91,959 \\ 

\hline

\end{tabular}
 }
\end{table*}

\begin{table*}[h]
\centering
\caption{Memory consumption (KB) of explored stream learning algorithms on selected real-world datasets. }
\label{tbl:memory}
\resizebox{1.8\columnwidth}{!}{
\begin{tabular}{|c|c|c|c|c|c||c|c|c|c|c|c|}
\hline
Dataset & Naive-NN & EWC & LwF & iCaRL & SEA-NN & Naive-DT & Naive-GBDT & ARF & SEA-DT & SEA-GBDT \\ \hline
ROOM & 22.7 & 50.8 & 45.3 & 23.4 & 106.6 & 1.9 & 6.6 & 228.1 & 4.2 & 20.1 \\ \hline
ELECTRICITY & 22.7 & 50.9 & 45.4 & 23.1 & 106.6 & 1.9 & 6.4 & 961.4 & 4.2 & 19.7 \\ \hline
INSECTS & 22.7 & 50.9 & 45.4 & 23.9 & 106.6 & 2.0 & 6.8 & 2,223.2 & 4.3 & 21.2 \\ \hline
AIR & 22.7 & 50.9 & 45.4 & 22.8 & 106.6 & 1.7 & 6.1 & N/A & 3.3 & 18.6 \\ \hline
POWER & 22.7 & 50.9 & 45.4 & 22.8 & 106.6 & 1.7 & 6.0 & N/A & 3.3 & 18.6 \\ 

\hline

\end{tabular}
 }
\end{table*}

\subsection{Challenges of Distribution Drifts}

\noindent
\begin{tcolorbox}[width=\linewidth,colback=white,boxrule=1pt,arc=0pt,outer arc=0pt,left=0pt,right=0pt,top=0pt,bottom=0pt,boxsep=1pt,halign=left]
\rev{\textbf{Finding (2):} Distribution drifts significantly degrade the effectiveness of stream learning algorithms. NN-based algorithms can better adapt to drifts than tree-based algorithms.}
\end{tcolorbox}

\begin{figure}[ht]
    \centering
    \subfloat[ROOM]{\includegraphics[width=0.23\textwidth]{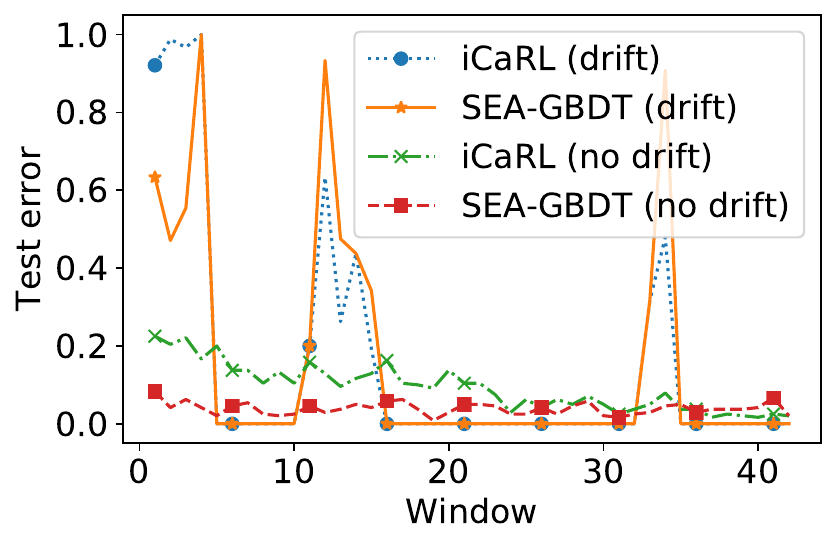}}
    \subfloat[AIR]{\includegraphics[width=0.23\textwidth]{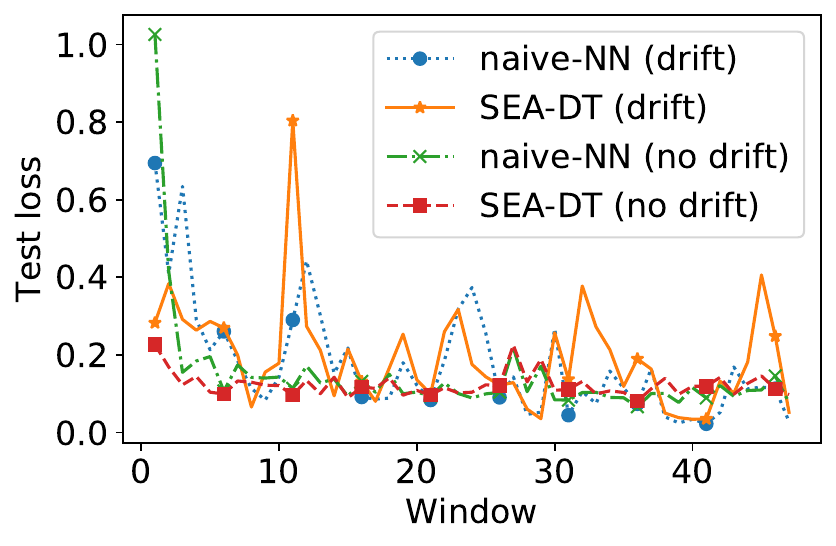}}
    \caption{Test loss / error curves on the ROOM and AIR dataset. Drift means the original dataset, while no drift means randomly shuffling the dataset to eliminate drifts.}
    \label{fig:compare_no_drift}
\end{figure}

\rev{In this section, we explore the challenges of distribution drifts by comparing the accuracy of stream learning algorithms on drifted datasets and non-drifted datasets. The non-drifted datasets are constructed by randomly shuffling the original datasets. We test the recommended algorithms on the ROOM and AIR dataset, and results are shown in Figure \ref{fig:compare_no_drift}. Drifted datasets lead to spikes in the test loss, while we can witness steady loss decrease in non-drifted datasets. This illustrates the great challenges of the distribution drifts in real-world data streams.}

\rev{A difficulty to address distribution drifts in real-world data streams is that there is no ground truth of the drift occurrences. Therefore, it is challenging to compare among drift detectors on real-world data streams. In our study, we directly compare the test loss of all windows. A stream learning algorithm handling drifts well can quickly adapt to new environment to achieve a lower loss. In Figure \ref{fig:compare_no_drift}, we find out that the selected NN-based algorithms can better adapt to drifts than tree-based algorithms, generally having lower loss at the drift occurrence points. A possible explanation is that neural networks are over-parameterized and can quickly converge to a good solution in the new environment. }

\subsection{Challenges of Outliers}

\noindent
\begin{tcolorbox}[width=\linewidth,colback=white,boxrule=1pt,arc=0pt,outer arc=0pt,left=0pt,right=0pt,top=0pt,bottom=0pt,boxsep=1pt,halign=left]
\rev{\textbf{Finding (3):} \vl{Among streaming outlier detectors, RShash and HSTree are recommended in data streams with more anomalies, while xStream is recommended in data streams with fewer anomalies.} Removing detected outliers does not necessarily improve accuracy in complex real-world open environment scenarios. }
\end{tcolorbox}

\vl{As discussed in Section \ref{sec:example_outliers}, even a single outlier can destroy the stream learning model. Therefore, it is crucial to detect and remove outliers in data streams. However, in real-world data streams, the lack of ground truth makes it challenging to compare streaming outlier detectors. To address this challenge, we mark the outliers by the ensemble of anomaly detectors on the whole dataset, and test streaming outlier detectors with the marked ground truth. } 

\vl{To mark the outliers in the whole dataset, we apply the ensemble of ECOD \citep{li2022ecod} and IForest \citep{liu2008isolation}, as they are recommended by a popular benchmark \citep{han2022adbench}. We evaluate the five multivariate streaming outlier detectors in StreamAD library: xStream \citep{manzoor2018xstream}, RShash \citep{sathe2016subspace}, HSTree \citep{tan2011fast}, LODA \citep{pevny2016loda}, and RRCF \citep{guha2016robust}. Results are shown in Table \ref{tbl:stream_ad}. We find that RShash and HSTree work better in data streams with relatively more anomalies, while xStream is recommended when the data streams have relatively fewer anomalies. When designing multivariate streaming outlier detectors, it may be good to consider applying dimensionality reduction or ensemble. Dimensionality reduction helps filter out noisy features and detect the abnormal features. Ensemble can boost accuracy from models of different perspectives. }

\begin{table}[h]
\centering
\caption{\vl{AUROC of streaming outlier detectors.} }
\label{tbl:stream_ad}
\resizebox{\columnwidth}{!}{
\begin{tabular}{|c|c|c|c|c|c|c|c|}
\hline
Anomaly statistic & Dataset & xStream & RShash & HSTree & LODA & RRCF \\ \hline
High & ROOM & 0.941 & \textbf{0.967} & 0.948 & 0.214 & 0.592 \\ \hline
Medium high & ELECTRICITY & 0.675 & \textbf{0.862} & 0.851 & 0.632 & 0.667 \\ \hline
Medium high & INSECTS & 0.842 & 0.927 & \textbf{0.965} & 0.807 & 0.678 \\ \hline
Medium low & AIR & \textbf{0.884} & 0.707 & 0.649 & 0.562 & 0.599 \\ \hline
Medium low & POWER & \textbf{0.879} & 0.784 & 0.829 & 0.499 & 0.570 \\ \hline

\end{tabular}
 }
\end{table}

\begin{figure}[ht]
    \centering
    \subfloat[ROOM]{\includegraphics[width=0.23\textwidth]{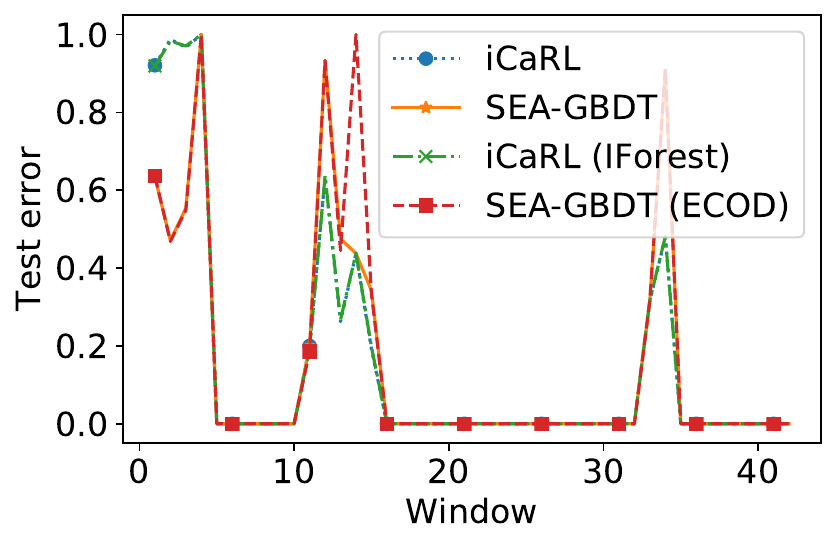}}
    \subfloat[AIR]{\includegraphics[width=0.23\textwidth]{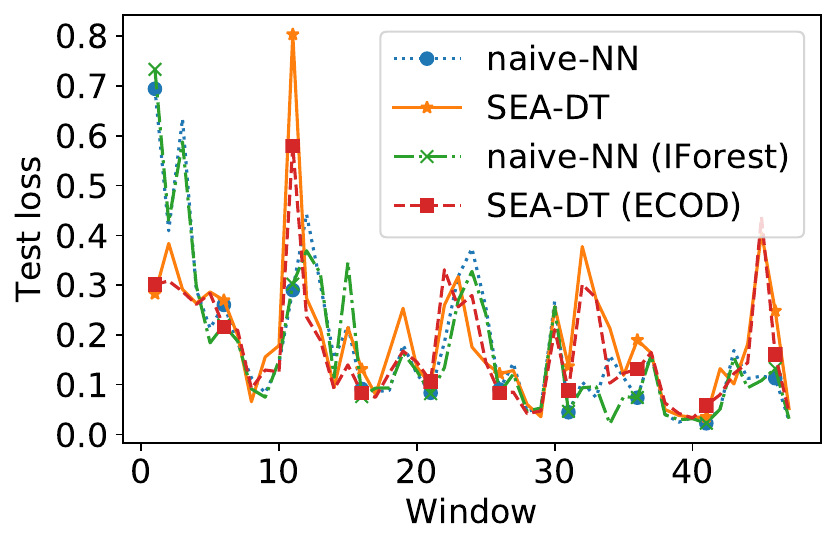}}
    \caption{Test loss / error on the ROOM and AIR dataset. ECOD and IForest mean removing outliers with the detector.}
    \label{fig:compare_no_outlier}
\end{figure}

\rev{We also compare the test loss on whether to remove detected outliers or not before training. We experiment on the ROOM and AIR dataset using recommended algorithms. Results are shown in Figure \ref{fig:compare_no_outlier}. As we can see, removing outliers improves accuracy in the AIR dataset, but it is ineffective in the ROOM dataset. This illustrates that addressing outliers in real-world data streams is a challenging topic and requires further researches. }

\subsection{Challenges of Missing Values}

\noindent
\begin{tcolorbox}[width=\linewidth,colback=white,boxrule=1pt,arc=0pt,outer arc=0pt,left=0pt,right=0pt,top=0pt,bottom=0pt,boxsep=1pt,halign=left]
\rev{\textbf{Finding (4):} KNN imputer is generally more effective than regression imputer, filling with average and filling with zero. This indicates that missing values can usually be effectively estimated from nearby samples in real-world data streams. }
\end{tcolorbox}

\begin{figure}[h]
    \centering
    \subfloat[NN models]{\includegraphics[width=0.23\textwidth]{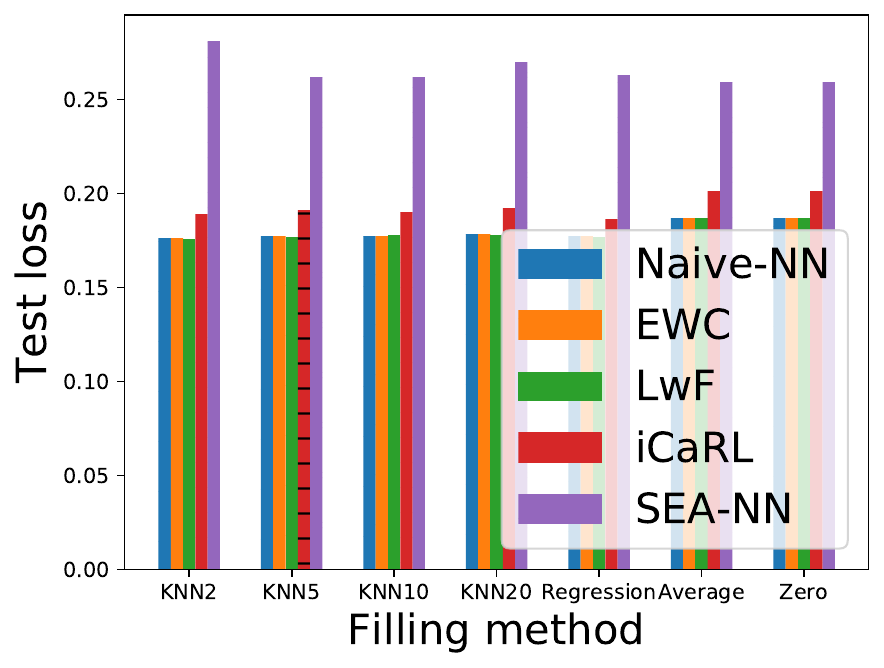}}
    \subfloat[Tree models]{\includegraphics[width=0.23\textwidth]{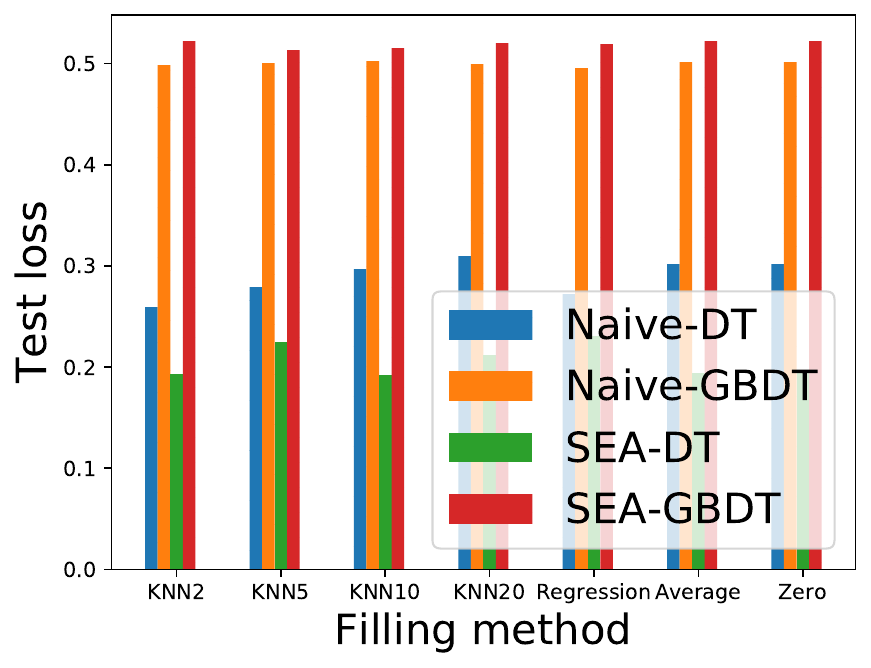}}
    \caption{Test error / test loss with respect to different missing value filling methods on the AIR dataset.}
    \vspace{-10pt}
    \label{fig:miss}
\end{figure}

In this section, we explore various missing value filling methods on the AIR dataset. We experiment with KNN imputer ($k=2,5,10,20$), regression imputer, filling with average and filling with zero. Figure \ref{fig:miss} reveals that KNN imputer generally outperforms other methods in terms of accuracy. Moreover, different values of $k$ for the KNN imputer do not significantly influence the accuracy. Considering that a smaller $k$ can save computation, we recommend using the KNN imputer with $k=2$ as a default setting.

\subsection{\vl{Experiments on Advanced Models}}

\noindent
\begin{tcolorbox}[width=\linewidth,colback=white,boxrule=1pt,arc=0pt,outer arc=0pt,left=0pt,right=0pt,top=0pt,bottom=0pt,boxsep=1pt,halign=left]
\vl{\textbf{Finding (5):} Advanced and complex models may easily overfit in open-environment regression tasks, creating a demand for models capable of learning generalizable knowledge for streaming data. }
\end{tcolorbox}

\vl{To explore the effectiveness and efficiency of advanced models for relational dataset, we test TabNet \citep{arik2021tabnet} and ARM-Net \citep{cai2021arm}. TabNet \citep{arik2021tabnet} combines neural network with decision tree to achieve end-to-end learning with interpretability. ARM-Net \citep{cai2021arm} learns adaptive feature interaction in the exponential space to capture cross feature information, which can help filter noisy or irrelevant features. }

\begin{figure}[h]
\centering
\includegraphics[width=0.7\columnwidth]{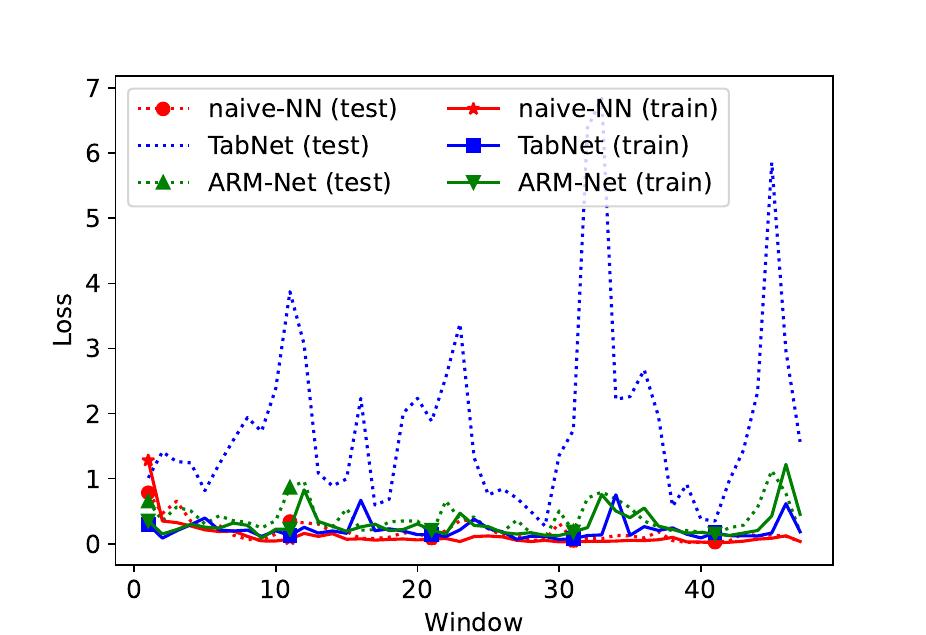}
\caption{\vl{Train and test loss on the AIR dataset. }}
\label{fig:overfit}
\end{figure}

\vl{We use their public codes with default settings. To be consistent, the number of epochs is set as 10. Results are shown in Table \ref{tbl:adv_model}. As we can see, ARM-Net has clear accuracy improvement on classification tasks (the first three rows). However, both TabNet and ARM-Net cannot work well in open environment regression tasks. As shown in Figure \ref{fig:overfit}, we observe that the test loss is much higher than train loss in TabNet and ARM-Net, due to the overfitting of complex models. Considering that the running time of TabNet and ARM-Net is over 20x than three-layer naive NN, both models may not be suitable to time-sensitive open environment learning in relational data streams. More results are shown in Appendix \ref{sec:large} of our full version \cite{diao2023oebench}. }

\begin{table}[h]
\centering
\caption{\vl{Test loss of advanced models on relational datasets.} }
\label{tbl:adv_model}
\resizebox{0.8\columnwidth}{!}{
\begin{tabular}{|c|c|c|c|}
\hline
Dataset & Naive-NN & TabNet & ARM-Net \\ \hline
ROOM & 0.214$\pm$0.004 & 0.213$\pm$0.011 & \textbf{0.204$\pm$0.010} \\ \hline
ELECTRICITY & 0.311$\pm$0.012 & 0.340$\pm$0.006 & \textbf{0.293$\pm$0.009} \\ \hline
INSECTS & 0.269$\pm$0.006 & 0.507$\pm$0.007 & \textbf{0.249$\pm$0.000} \\ \hline
AIR & \textbf{0.166$\pm$0.002} & 1.942$\pm$0.074 & 0.408$\pm$0.014 \\ \hline
POWER & \textbf{0.793$\pm$0.005} & 1.578$\pm$0.102 & 0.958$\pm$0.007 \\ \hline

\end{tabular}
 }
\end{table}

\subsection{\vl{Investigating Synthetic Datasets}}

\noindent
\begin{tcolorbox}[width=\linewidth,colback=white,boxrule=1pt,arc=0pt,outer arc=0pt,left=0pt,right=0pt,top=0pt,bottom=0pt,boxsep=1pt,halign=left]
\vl{\textbf{Finding (6):} Synthetic datasets have limited coverage on open environment challenges. Conclusions drawn from experiments on synthetic datasets are much biased from experiments on our OEBench. }
\end{tcolorbox}

\vl{We investigate six popular synthetic data streams summarized in a highly cited work \citep{bifet2009new}: SEA (used in \citep{wang2015concept,abbasi2021elstream,mahdi2020diversity,wang2022noise,gama2011learning,brzezinski2011accuracy}), STAGGER (used in \citep{anderson2019recurring,beringer2007efficient,yang2006mining,gama2011learning}), Rotating Hyperplane (used in \citep{wang2015concept,krawczyk2018combining,abbasi2021elstream,beringer2007efficient,yang2006mining,gama2011learning,brzezinski2011accuracy}), RBF (used in \citep{krawczyk2018combining,abbasi2021elstream,anderson2019recurring}), LED (used in \citep{pesaranghader2018mcdiarmid,abbasi2021elstream,wang2022noise,anderson2019recurring,gama2003accurate,gama2011learning,brzezinski2011accuracy}), and Waveform (used in \citep{gama2003accurate,gama2011learning,brzezinski2011accuracy}). Among the six datasets, SEA and STAGGER create abrupt concept drifts by suddenly changing the decision rules at certain points. Rotating Hyperplane and RBF create incremental drifts by continuously altering the parameters of the decision boundaries. LED generates drifts by swapping features, and it also includes many irrelevant features. Waveform focuses on simulating noisy and irrelevant attributes. }

\vl{These synthetic datasets do not cover regression task and have no missing values. Besides, as shown in Figure \ref{fig:boxplot}, the anomaly ratio of synthetic datasets is significantly lower than that of our collected real-world datasets. Among the three open environment challenges, these synthetic datasets only simulate the drifts, while ignoring the other two challenges of outliers and missing values. Moreover, synthetic datasets have very narrow range of drift statistics, and their drift ratios are also relatively low compared to our collected real-world data streams. }

\vl{We also evaluate stream learning algorithms on these synthetic datasets. According to previous recommendations, we test naive NN, iCaRL, naive GBDT, SEA-DT, SEA-GBDT and ARM-Net. Results on SEA and LED are shown in Figure \ref{fig:synthetic}. More results are shown in Appendix \ref{sec:all55} of our full version \cite{diao2023oebench}. As we can see, GBDT or SEA-GBDT can work well on these synthetic datsets. ARM-Net can help tackle irrelevant features and achieve better results in the LED dataset. This is in contrast with our findings in Section \ref{sec:overall}. Thus, as these synthetic datasets have limited coverage on open environment challenges, the conclusions drawn from experiments on synthetic datasets are much biased from experiments on our OEBench. Algorithms that can handle synthetic datasets well may not work well on real-world data streams. }

\begin{figure}[ht]
    \centering
    \subfloat[SEA]{\includegraphics[width=0.23\textwidth]{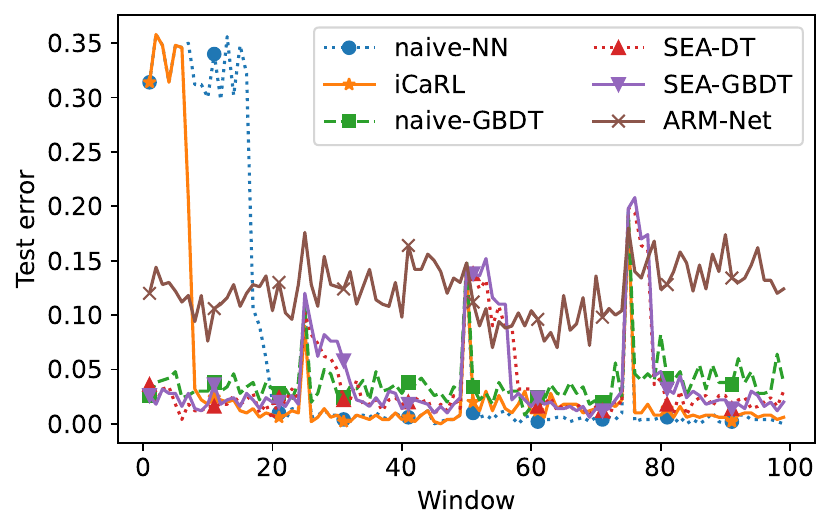}}
    \subfloat[LED]{\includegraphics[width=0.23\textwidth]{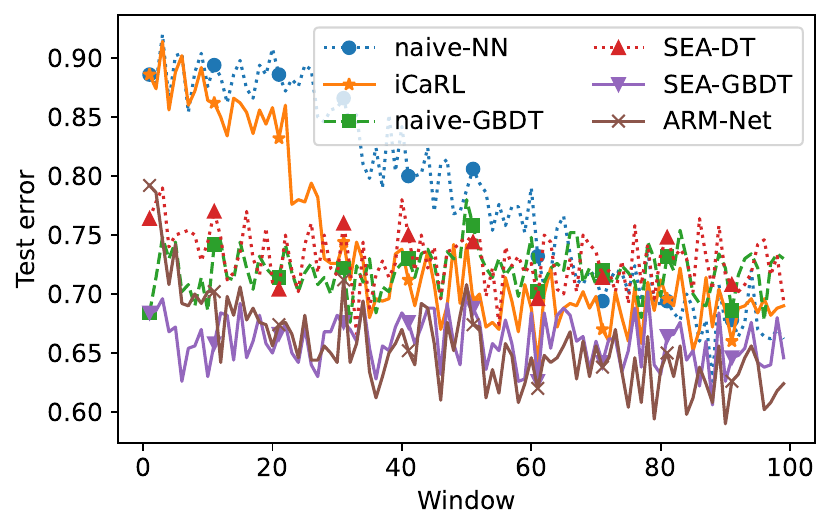}}
    \caption{\vl{Test error on synthetic datsets.}}
    \label{fig:synthetic}
\end{figure}

\section{Conclusion}
\rev{In this paper, we explore real-world relational data streams considering the recently proposed open environment challenges \citep{zhou2022open}. We propose a benchmark \textit{OEBench} consisting of six stages: dataset collection, preprocessing, extracting open environment statistics, visualization, representative dataset selection and incremental learning algorithm evaluation. Specifically, we study 55 diverse real-world relational data streams, select five representative datasets based on the open environment statistics, and evaluate the performance of 10 existing incremental learning algorithms. We also conduct case studies with visualizations to explore how open environment challenges look like and their impacts on real-world relational data stream learning. The results highlight the inadequacy of current methods in effectively addressing these challenges in real-world relational data streams. Further researches are encouraged towards more effective and efficient approaches to data stream learning under open environment challenges.}




\bibliographystyle{ACM-Reference-Format}
\bibliography{main}

\clearpage
\appendix
\section{More Backgrounds and Related Works}

\subsection{More Real-World Examples on Open Environment Challenges}

    \rev{\textbf{Autonomous vehicles.} A self-driving car operates in an open environment where it can encounter unseen scenarios during training. (1) Many factors (e.g. traffic situation, weather) can change from time to time. The autonomous driving model must continuously adapt to these situations, which leads to the challenge of distribution drifts. (2) It may come across unique traffic situations, new traffic signs, or unusual pedestrian behaviors that are not present in its training data. These belong to the challenge of outliers or new classes. (3) Old cameras can malfunction while new cameras can be installed, which constitutes the challenge of incremental/decremental feature space. }
    
    \rev{\textbf{Real-time health monitoring.} Consider a wearable device like a smartwatch that monitors a user's health conditions for disease prediction. (1) The wearer's health condition might change, reflecting alterations in the underlying data distributions and causing the challenge of distribution drifts. (2) Moreover, the user might develop a medical condition that is absent during the initial training of the model, introducing the problem of outliers or novel classes. (3) Additionally, software updates in the device can modify the methods of data collection, leading to potential changes in the input feature space (incremental/decremental feature space). }
    
    \rev{\textbf{Financial fraud detection.} In the financial industry, detecting fraudulent transactions is critical. (1) Over time, typical consumer behaviors may change, causing a shift in transaction data distributions and thus, the challenge of distribution drifts. (2) Fraudsters may devise new fraudulent strategies that have not been captured in the existing data, introducing the issue of outliers or novel classes. (3) With the continuous evolution of new financial technology (e.g., mobile payment methods, cryptocurrency transactions), the types of data collected could change, leading to changes in the input feature space (incremental/decremental feature space).}

\subsection{Detailed Discussions of Incremental Learning Algorithms}
\label{sec:il_detail}

Incremental learning techniques can be divided into three categories: regularization, use of exemplars, and parameter isolation. Regularization-based algorithms preserve the knowledge on older windows by penalizing changes in important parameters associated with those windows. Exemplar-based algorithms store selected samples from old windows to alleviate catastrophic forgetting in learning new samples. Parameter-isolation aims to add new modules or freeze part of the neurons to preserve the acquired knowledge. 

From the perspective of regularization, various approaches have been proposed. Elastic Weight Consolidation (EWC) \citep{kirkpatrick2017overcoming} penalizes the change in crucial parameters from previous windows based on the Fisher Information Matrix. PathInt \citep{zenke2017continual} estimates the importance of parameters by tracing the values of parameters at each window. If a parameter significantly influences the loss of previous windows, it is assigned greater penalty in the current window. Memory Aware Synapses (MAS) \citep{aljundi2018memory} calculates the importance factor by online accumulation of gradient magnitude. Rwalk \citep{chaudhry2018riemannian} revises EWC with KL divergence regularization for memory efficiency and combines it with PathInt to compute each parameter's importance. LwF \citep{li2017learning} integrates the prediction of the previous model into the current loss function, which reduces the over-confidence towards seen classes of the current window. LwM \citep{dhar2019learning} further introduces attention distillation loss to force the attention heat map of the current model similar to that of the previous model in vision tasks. DMC \citep{zhang2020class} proposes to perform distillation on previous models and the current model to balance the supervision between previous and current windows. \citet{smith2021always} improve data-free knowledge distillation to generate examples from previous models and train the current model with more balanced datasets. DDE \citep{hu2021distilling} refers to casual inference and proposes to regularize by representation distance of previous models. SPB \citep{wu2021striking} regularizes the training by penalizing representation change with contrastive loss. \citet{toldo2022bring} explore the feature drifts and revive features of old classes during training. 

From the perspective of storing exemplars, iCaRL \citep{rebuffi2017icarl} iteratively selects exemplars that are close to the mean representation of the corresponding class. Rwalk \citep{chaudhry2018riemannian} chooses exemplars which are hard to classify. BiC \citep{wu2019large} applies a similar exemplar selection strategy but uses the exemplars as a validation dataset to train a bias-correcting linear model for the last layer. LUCIR \citep{hou2019learning} utilizes exemplars to push the embedding of current classes away from previous similar classes. It also normalizes the last classification layer to reduce the bias towards the current window. Der \citep{yan2021dynamically} decouples the representation learning with classifier learning. For a new window, it trains a new encoder with encoders of previous windows fixed, and stores exemplars as a balanced dataset to train the final classification layer. RMM \citep{liu2021rmm} utilizes reinforcement learning to allocate the memory for exemplars of different classes.

From the perspective of parameter isolation, PackNet \citep{mallya2018packnet} prunes the neural network for new windows while keeping important parameters frozen. Expert Gate \citep{aljundi2017expert} maintains the model of each window and trains a gating function to select the appropriate expert model. However, real-world data streams can be infinite, making it infeasible to continually expand network architecture or isolate neurons.

\subsection{Drift Detection}
\label{sec:drift_detection}
Drift detection plays an important role in open environment learning. By identifying the occurrence, type, and magnitude of distribution drifts, algorithms can be modified and updated accordingly. 

\rev{Drift detection methods fall into two broad categories: data drift detection and concept drift detection. Denote features as $X$ and labels as $Y$. Data drift detectors estimate the change of $(X, Y)$, while concept drift detectors estimate the change of the learned function $f:\mathcal{X} \rightarrow \mathcal{Y}$. Therefore, the input of data drift detectors is usually the dataset itself, while concept drift detectors mostly require the error rate of a model trained on the data stream.  }

In terms of data drift detection, \citet{lu2018learning} propose to compute the similarity between new and old data with certain distance metrics, and then conduct a hypothesis test to decide whether a drift occurs. Frequently employed statistical tests include the Kolmogorov-Smirnov test \citep{berger2014kolmogorov}, Wilcoxon test, and Kullback-Leibler divergence \citep{joyce2011kullback}. \citet{kifer2004detecting} propose to apply the Kolmogorov-Smirnov test between adjacent windows. Distance metrics, such as the Hellinger distance, are often used to determine the similarity between different distributions. However, without an appropriate window size, potential drifts might go unnoticed. Adaptive window-based solutions, like ADaptive WINdowing (ADWIN) \citep{bifet2007learning}, are proposed to detect distribution drifts using an adaptive window size. 

In terms of concept drift detection, a typical technique is to closely observe the model accuracy. A substantial decrease in the model accuracy may signal a change in the mapping function from inputs to targets. Algorithms like the Drift Detection Method (DDM) \citep{gama2004learning} and Early Drift Detection Method (EDDM) \citep{baena2006early} signal drifts when the error rate falls out of control. However, the task of drift detection becomes complex when the ground truth is not readily available.


Evaluating drift detection methods can be challenging due to the uncertainty of drift occurrences in real-world datasets \citep{lu2018learning}. Thus, many works \citep{wang2015concept,losing2016knn,lu2016concept,bu2016pdf} experiment on manually designed datasets such as rotating hyperplane \citep{hulten2001mining} and Usenet \citep{katakis2008ensemble}. Real-world datasets, like Electricity Prices \citep{harries1999splice} and Covertype \citep{asuncion2007uci}, are also used in a prequential setting, where data is first tested then trained in each window.

We provide a summary of 16 popular drift detection algorithms in Table \ref{tbl:driftdetection}. The comparison includes detector type, required input data, data type (categorical or numerical), applicable task, and compatibility with stream or batch data processing. \rev{We have made the following findings:}

\begin{itemize}
    \item \rev{Some approaches are designed to detect drifts on 1-D data, e.g. KS statistic, ADWIN \citep{bifet2007learning} and HDDM \citep{frias2014online}. To adapt them to multi-dimensional data, we have to conduct drift detection in each dimension separately.}
    \item \rev{Many concept drift detectors are designed only for classification or even binary classification tasks. Only PERM \citep{harel2014concept} and EIA \citep{baier2020handling} are applicable to regression tasks. However, EIA \citep{baier2020handling} compares the current model with a simple model of predicting with the latest target, which is not quite reasonable. Thus, to detect concept drifts in regression tasks, PERM \citep{harel2014concept} is the most suitable choice. One can also consider adapting the error rate to the regression loss in DDM \citep{gama2004learning} or other methods.}
    \item \rev{Some approaches like KS statistic are designed to compare batches. To adapt to data streams, we can divide the streams into windows and run the methods on adjacent windows.}
\end{itemize}




\begin{table*}[htpb]
\caption{Summary of drift detection methods.}
\label{tbl:driftdetection}
\centering
\newcommand{\y}{\ding{51}}
\newcommand{\n}{\ding{55}}
\begin{tabular}{|c|c|c|c|c|c|c|}
\hline
Method & Detector type & Input & Data type & Applicable task & Stream & Batch\\
\hline
DDM \citep{gama2004learning} & Concept drift & Error rate & Categorical / Numerical & Classification  & \y & \n\\ \hline
EDDM \citep{baena2006early} & Concept drift & Error rate & Categorical / Numerical & Classification  & \y & \n\\ \hline
ADWIN accuracy \citep{bifet2007learning} & Concept drift & Error rate & Categorical / Numerical & Classification & \y & \n\\ \hline
FW-DDM \citep{liu2017fuzzy} & Concept drift & Error rate & Categorical / Numerical & Classification  & \y & \n\\ \hline
ECDD \citep{ross2012exponentially} & Concept drift & Error rate & Categorical / Numerical & Classification  & \y & \n\\ \hline
LFR \citep{wang2015concept} & Concept drift & Error rate & Categorical / Numerical & Binary classification & \y & \n\\ \hline
MD3 \citep{sethi2015don} & Concept drift & Margin data rate & Categorical / Numerical & Binary classification & \y & \n\\ \hline
PERM \citep{harel2014concept} & Concept drift & Test loss & Categorical / Numerical & Classification / Regression & \n & \y\\ \hline
EIA \citep{baier2020handling} & Concept drift & Error intersection & Categorical / Numerical & Classification / Regression & \n & \y\\ \hline
KS statistic & Data drift & 1-D data & Categorical / Numerical & Classification / Regression & \n & \y\\ \hline
ADWIN \citep{bifet2007learning} & Data drift & 1-D data & Numerical & Classification / Regression & \y & \n\\ \hline
HDDM \citep{frias2014online} & Data drift & 1-D data & Numerical & Classification / Regression & \y & \n\\ \hline
CDBD \citep{lindstrom2013drift} & Data drift & Confidence score & Categorical / Numerical & Classification / Regression  & \n & \y\\ \hline
HDDDM \citep{ditzler2011hellinger} & Data drift & Multi-dimension data & Categorical / Numerical & Classification / Regression & \n & \y\\ \hline
kdq-Tree \citep{dasu2006information} & Data drift & Multi-dimension data & Categorical / Numerical & Classification / Regression & \n & \y\\ \hline
PCA-CD \citep{qahtan2015pca} & Data drift & Multi-dimension data & Numerical & Classification / Regression & \n & \y\\
\hline
\end{tabular}

\end{table*}

\section{Additional Experiments}

\subsection{Experiments on More Datasets}
\label{sec:all55}
\rev{We show the experimental results of all 55 datasets in Table \ref{tbl:all}. As we can see, no algorithm consistently outperforms others. The recommendations on different open environment cases are shown in the main paper.}

\begin{figure*}[ht]
    \centering
    \subfloat[Rotation Hyperplane]{\includegraphics[width=0.25\textwidth]{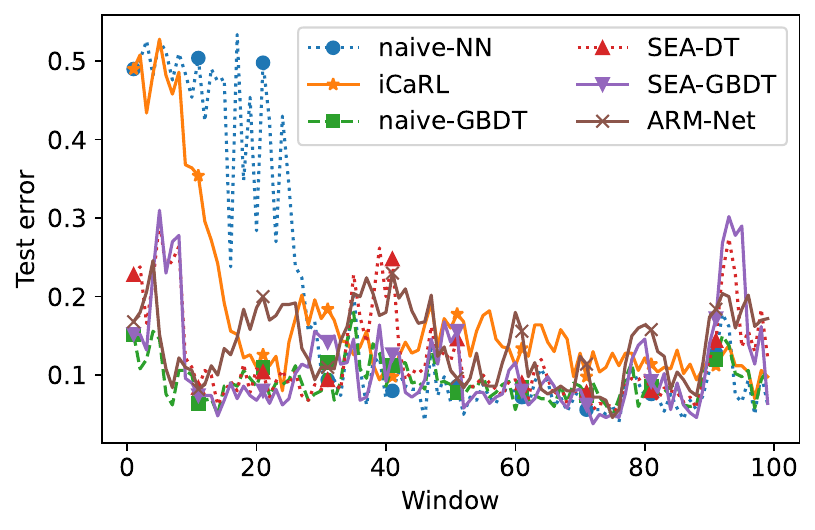}}
    \subfloat[STAGGER]{\includegraphics[width=0.25\textwidth]{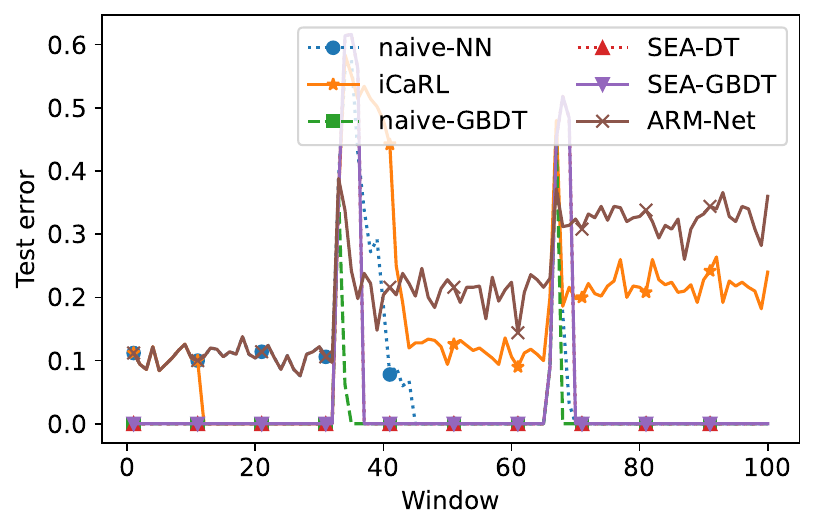}}
    \subfloat[RBF]{\includegraphics[width=0.25\textwidth]{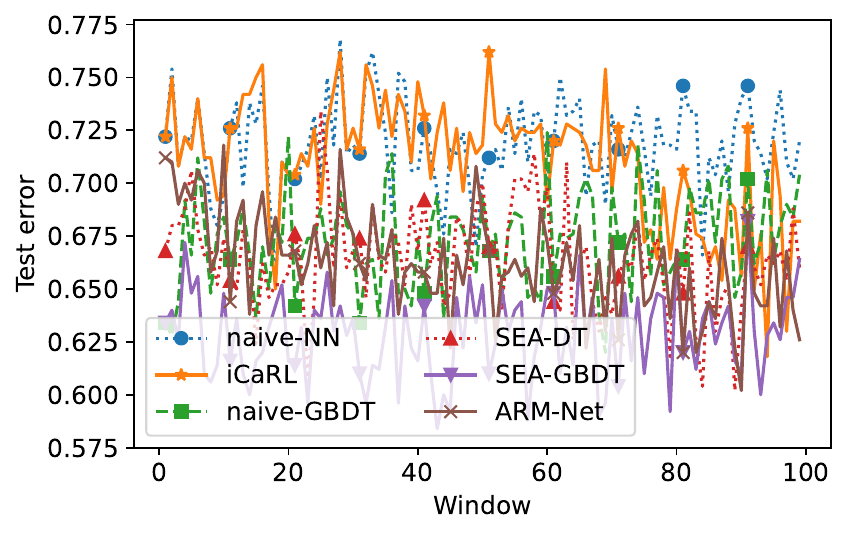}}
    \subfloat[Waveform]{\includegraphics[width=0.25\textwidth]{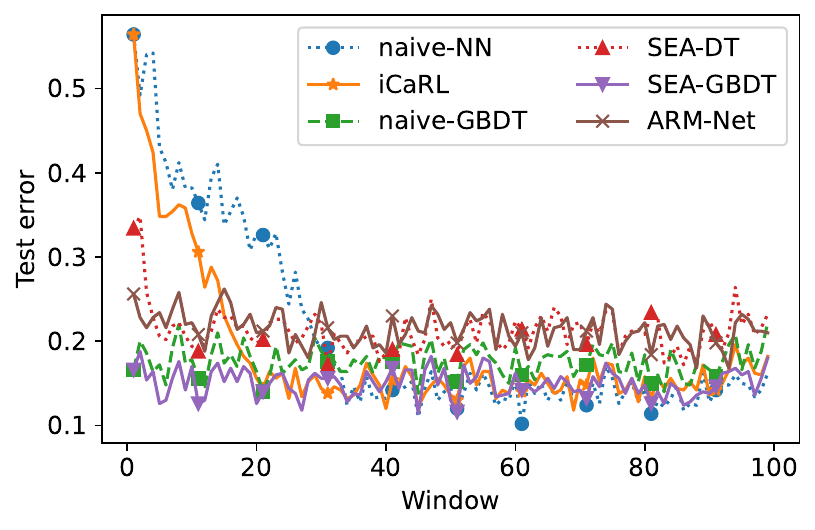}}
    \caption{Test error on four synthetic datasets.}
    \label{fig:synthetic_more}
\end{figure*}

\vl{We also show the results of four additional synthetic datasets in Figure \ref{fig:synthetic_more}. As we can see, just using GBDT or SEA-GBDT is good for synthetic datasets.} 

\begin{table*}[h]
\centering
\caption{Test loss / test error of stream learning algorithms on real-world relational data streams with different characteristics. Lower value indicates better result. We repeat all experiments for three times with different random seeds.}
\label{tbl:all}
\resizebox{2.1\columnwidth}{!}{
\begin{tabular}{|c|c|c|c|c|c|c|c|c|c||c|c|c|c|c|}
\hline
Task & Drift & Anomaly & Missing value & Dataset & Naive-NN & EWC & LwF & iCaRL & SEA-NN & Naive-DT & Naive-GBDT & SEA-DT & SEA-GBDT \\ \hline
Classification & Medium high & High & Low & Room occupancy & 0.214$\pm$0.004 & 0.207$\pm$0.003 & 0.207$\pm$0.014 & \textbf{0.136$\pm$0.021} & 0.207$\pm$0.037 & 0.198$\pm$0.006 & 0.181$\pm$0.004 &0.191$\pm$0.004 & \textbf{0.151$\pm$0.002} \\ \hline
Classification & Medium high & Medium high & Low & Electricity price & 0.311$\pm$0.012 & 0.311$\pm$0.012 & 0.311$\pm$0.012 & \textbf{0.286$\pm$0.013} & 0.332$\pm$0.022 & 0.272$\pm$0.001 & \textbf{0.256$\pm$0.000} & 0.263$\pm$0.009 & 0.264$\pm$0.001 \\ \hline
Classification & Medium high & Medium high & Low & Covertype & \textbf{0.320$\pm$0.006} & \textbf{0.320$\pm$0.004} & \textbf{0.320$\pm$0.003} & 0.349$\pm$0.006 & 0.455$\pm$0.008 & 0.382$\pm$0.001 & \textbf{0.380$\pm$0.000} & 0.432$\pm$0.006 & 0.422$\pm$0.000   \\ \hline
Classification & Medium high & Medium low & Low & NOAA & 0.717$\pm$0.020 & 0.716$\pm$0.021&0.718$\pm$0.020&\textbf{0.695$\pm$0.011}&0.757$\pm$0.008&\textbf{0.556$\pm$0.003}&0.568$\pm$0.001&0.594$\pm$0.004&0.625$\pm$0.029 \\ \hline
Classification & Medium high &  Medium high & Low & RSSI & 0.379$\pm$0.003 & 0.379$\pm$0.003 & 0.379$\pm$0.003 & \textbf{0.331$\pm$0.003} & 0.431$\pm$0.032  & 0.372$\pm$0.001  & 0.277$\pm$0.000 & 0.309$\pm$0.004 & \textbf{0.256$\pm$0.000} \\ \hline
Classification & Medium low & Medium high & Low & INSECTS (incremental, reoccurring, balanced) & \textbf{0.269$\pm$0.006} & \textbf{0.269$\pm$0.006} & \textbf{0.269$\pm$0.006} & 0.306$\pm$0.005 & 0.321$\pm$0.007 & 0.329$\pm$0.001 & 0.306$\pm$0.000 & \textbf{0.291$\pm$0.004} & \textbf{0.291$\pm$0.002} \\ \hline
Classification & Medium low & Medium high & Low & INSECTS (abrupt, imbalanced) &\textbf{ 0.202$\pm$0.003}&\textbf{0.202$\pm$0.003}&\textbf{0.202$\pm$0.003}&0.234$\pm$0.001&\textbf{0.202$\pm$0.002}&0.366$\pm$0.000&0.328$\pm$0.000&\textbf{0.252$\pm$0.001}&0.256$\pm$0.001 \\ \hline
Classification & Low & Low & Low & Safe driver & \textbf{0.036$\pm$0.000} & \textbf{0.036$\pm$0.000} & \textbf{0.036$\pm$0.000} & 0.039$\pm$0.000 & \textbf{0.036$\pm$0.000} & 0.081$\pm$0.000 & \textbf{0.036$\pm$0.000} & 0.040$\pm$0.000 & \textbf{0.036$\pm$0.000}  \\ \hline
Classification & Medium low & Medium high & Low & INSECTS (abrupt, balanced) & 0.337$\pm$0.021 & 0.338$\pm$0.022 & 0.337$\pm$0.021 & \textbf{0.328$\pm$0.012} & 0.395$\pm$0.023 & 0.383$\pm$0.000 & 0.356$\pm$0.000 & 0.328$\pm$0.002 & \textbf{0.309$\pm$0.003}   \\ \hline
Classification & Medium low & Low & Low & KDDCUP99 & 0.227$\pm$0.184&0.381$\pm$0.235&0.792$\pm$0.009&0.117$\pm$0.104&\textbf{0.013$\pm$0.004}&0.015$\pm$0.000&0.015$\pm$0.000&0.008$\pm$0.000&\textbf{0.007$\pm$0.000} \\ \hline
Classification & Low & Medium high & Low & INSECTS (out of control) & 0.312$\pm$0.001&0.312$\pm$0.001&0.440$\pm$0.221&0.359$\pm$0.001&\textbf{0.296$\pm$0.001}&0.537$\pm$0.000&0.563$\pm$0.000&\textbf{0.399$\pm$0.002}&0.423$\pm$0.001 \\ \hline
Classification & Medium low & Medium high & Low & INSECTS (incremental, imbalanced) & \textbf{0.215$\pm$0.002}&\textbf{0.215$\pm$0.002}&0.216$\pm$0.001&0.252$\pm$0.002&0.238$\pm$0.002&0.316$\pm$0.000&0.263$\pm$0.000&\textbf{0.255$\pm$0.002}&0.259$\pm$0.001 \\ \hline
Classification & Medium high & Medium low & Low & INSECTS (incremental, balanced) & 0.460$\pm$0.019&0.460$\pm$0.019&\textbf{0.459$\pm$0.019}&0.484$\pm$0.009&0.628$\pm$0.109&0.463$\pm$0.001&\textbf{0.376$\pm$0.001}&0.438$\pm$0.006&0.405$\pm$0.000 \\ \hline
Classification & Medium high & High & Low & INSECTS (incremental, abrupt, balanced) & \textbf{0.288$\pm$0.002}&\textbf{0.288$\pm$0.002}&\textbf{0.288$\pm$0.003}&0.326$\pm$0.003&0.344$\pm$0.008&0.346$\pm$0.001&0.319$\pm$0.001&\textbf{0.312$\pm$0.001}&0.319$\pm$0.005 \\ \hline
Classification & Medium high & Medium high & Low & INSECTS (gradual, imbalanced) & \textbf{0.220$\pm$0.001}&\textbf{0.220$\pm$0.001}&\textbf{0.220$\pm$0.001}&0.262$\pm$0.003&0.221$\pm$0.000&0.367$\pm$0.002&0.327$\pm$0.000&\textbf{0.256$\pm$0.001}&0.261$\pm$0.001 \\ \hline
Classification & Medium high & Medium high & Low & INSECTS (incremental, abrupt, imbalanced) & 0.220$\pm$0.001&0.220$\pm$0.000&\textbf{0.219$\pm$0.001}&0.266$\pm$0.003&0.223$\pm$0.001&0.385$\pm$0.000&0.349$\pm$0.000&\textbf{0.279$\pm$0.000}&0.285$\pm$0.001 \\ \hline
Classification & Medium high & Medium high & Low & INSECTS (incremental, reoccurring, imbalanced) & \textbf{0.216$\pm$0.000}&\textbf{0.216$\pm$0.000}&\textbf{0.216$\pm$0.001}&0.262$\pm$0.002&0.218$\pm$0.001&0.386$\pm$0.000&0.349$\pm$0.000&\textbf{0.275$\pm$0.000}&0.284$\pm$0.000 \\ \hline
Classification & Medium high & Medium high & Low & INSECTS (gradual, balanced) & 0.374$\pm$0.015&0.374$\pm$0.015&0.374$\pm$0.015&\textbf{0.349$\pm$0.015}&0.421$\pm$0.011&0.342$\pm$0.002&0.320$\pm$0.000&0.294$\pm$0.007&\textbf{0.282$\pm$0.004} \\ \hline
Classification & High & High & Low & Bitcoin & \textbf{0.014$\pm$0.000}&\textbf{0.014$\pm$0.000}&\textbf{0.014$\pm$0.000}&0.055$\pm$0.011&\textbf{0.014$\pm$0.000}&0.020$\pm$0.000&0.015$\pm$0.000&\textbf{0.014$\pm$0.000}&\textbf{0.014$\pm$0.000} \\ \hline
Classification & Medium low & Low & Low & Airlines & 0.381$\pm$0.002&0.381$\pm$0.002&0.381$\pm$0.002&\textbf{0.380$\pm$0.001}&0.405$\pm$0.008&0.398$\pm$0.001&0.422$\pm$0.000&\textbf{0.374$\pm$0.001}&0.382$\pm$0.000 \\ \hline

Regression & Low & Medium low & High & Air quality (Shunyi) & \textbf{0.166$\pm$0.002} & \textbf{0.166$\pm$0.002} & \textbf{0.166$\pm$0.002} & 0.182$\pm$0.008 & 0.213$\pm$0.023 & 0.263$\pm$0.013 & 0.498$\pm$0.002 & \textbf{0.199$\pm$0.010} & 0.519$\pm$0.003 \\ \hline
Regression & High & Medium low & Low & Tetouan power consumption & 0.793$\pm$0.005 & 0.794$\pm$0.004 & \textbf{0.779$\pm$0.004} & 0.818$\pm$0.014 & 0.783$\pm$0.015 & 1.278$\pm$0.003 & \textbf{0.800$\pm$0.000} & 0.845$\pm$0.007 & 0.835$\pm$0.002  \\ \hline
Regression & Medium high & Medium low & Low & Bike sharing & \textbf{0.039$\pm$0.002} & \textbf{0.039$\pm$0.002} & 0.040$\pm$0.002 & 0.044$\pm$0.004 & 0.130$\pm$0.008 & \textbf{0.016$\pm$0.001} & 0.379$\pm$0.000 & 0.039$\pm$0.002 & 0.467$\pm$0.000 \\ \hline
Regression & Medium low & Medium low & High & Indian weather (Bangalore) & \textbf{0.213$\pm$0.012 }&0.214$\pm$0.012 &0.214$\pm$0.012&0.222$\pm$0.013&0.292$\pm$0.055&0.258$\pm$0.002&0.445$\pm$0.000&\textbf{0.163$\pm$0.001}&0.442$\pm$0.000
 \\ \hline
Regression & High & Medium high & High & Italian air quality & 0.280$\pm$0.006&0.280$\pm$0.006&0.280$\pm$0.006&\textbf{0.273$\pm$0.003}&0.496$\pm$0.043&\textbf{0.355$\pm$0.004}&0.538$\pm$0.000&0.412$\pm$0.004&0.718$\pm$0.000
 \\ \hline
Regression & Medium high &High & High & 5 cities PM2.5 (Chengdu) & 1.071$\pm$0.007&1.068$\pm$0.010&1.072$\pm$0.010&\textbf{1.016$\pm$0.006}&1.186$\pm$0.041&1.727$\pm$0.006&\textbf{1.139$\pm$0.000}&1.370$\pm$0.012&1.215$\pm$0.000

 \\ \hline
Regression & High & High &Low& Energy prediction & 0.965$\pm$0.009&0.966$\pm$0.008&0.966$\pm$0.010&0.969$\pm$0.007&\textbf{0.940$\pm$0.001}&5.623$\pm$0.070&1.241$\pm$0.007&2.824$\pm$0.133&\textbf{1.028$\pm$0.025}
 \\ \hline
Regression & High & Medium high & Low & Household & 0.492$\pm$0.002&0.494$\pm$0.001&
0.492$\pm$0.002&0.536$\pm$0.007&\textbf{0.439$\pm$0.004}&0.529$\pm$0.000&0.464$\pm$0.000&0.465$\pm$0.001&\textbf{0.443$\pm$0.000} \\ \hline

Regression & Low & Low & Low & Allstate claim & 0.492$\pm$0.002 &0.492$\pm$0.002&0.489$\pm$0.001&0.495$\pm$0.001&\textbf{0.474$\pm$0.002}&1.05$\pm$0.003&0.782$\pm$0.000&\textbf{0.593$\pm$0.001}&0.773$\pm$0.000  \\ \hline
Regression & Medium low & Low & Low & Air quality (Wanliu) & 0.186$\pm$0.009&0.186$\pm$0.009&\textbf{0.185$\pm$0.008}&0.208$\pm$0.015&0.226$\pm$0.040&0.308$\pm$0.005&0.527$\pm$0.002&\textbf{0.239$\pm$0.008}&0.556$\pm$0.001  \\ \hline
Regression & Medium low & Medium low & Low & Air quality (Wanshouxingong) & \textbf{0.170$\pm$0.004}&\textbf{0.170$\pm$0.004}&\textbf{0.170$\pm$0.004}&0.184$\pm$0.007&0.267$\pm$0.061&0.289$\pm$0.006&0.523$\pm$0.000&\textbf{0.214$\pm$0.011}&0.596$\pm$0.005  \\ \hline
Regression & Medium low & Medium low & Low & Air quality (Gucheng) & \textbf{0.184$\pm$0.009}&\textbf{0.184$\pm$0.009}&\textbf{0.184$\pm$0.009}&0.202$\pm$0.016&0.284$\pm$0.029&\textbf{0.372$\pm$0.013}&0.570$\pm$0.000&0.386$\pm$0.017&0.696$\pm$0.001  \\ \hline
Regression & Medium low & Medium low & Low & Air quality (Huairou) & \textbf{0.174$\pm$0.008}&\textbf{0.174$\pm$0.008}&\textbf{0.174$\pm$0.008}&0.197$\pm$0.01&0.236$\pm$0.031&0.307$\pm$0.009&0.519$\pm$0.003&\textbf{0.224$\pm$0.013}&0.490$\pm$0.004  \\ \hline
Regression & Medium low & Medium low & Low & Air quality (Nongzhanguan) & \textbf{0.154$\pm$0.012}&\textbf{0.154$\pm$0.011}&\textbf{0.154$\pm$0.012}&0.163$\pm$0.011&0.257$\pm$0.024&0.272$\pm$0.014&0.504$\pm$0.002&\textbf{0.200$\pm$0.028}&0.572$\pm$0.027  \\ \hline
Regression & Medium low & Medium low & Low & Air quality (Changping) & \textbf{0.171$\pm$0.009}&0.172$\pm$0.009&\textbf{0.171$\pm$0.008}&0.18$\pm$0.014&0.191$\pm$0.027&0.337$\pm$0.015&0.549$\pm$0.000&\textbf{0.203$\pm$0.013}&0.489$\pm$0.002  \\ \hline
Regression & Medium low & Medium low & Low & Air quality (Dingling) & \textbf{0.164$\pm$0.011}&\textbf{0.164$\pm$0.011}&\textbf{0.164$\pm$0.011}&0.180$\pm$0.010&0.186$\pm$0.031&0.317$\pm$0.005&0.551$\pm$0.000&\textbf{0.203$\pm$0.020}&0.542$\pm$0.001  \\ \hline
Regression & Medium low & Medium low & Low & Air quality (Aotizhongxin) & 0.181$\pm$0.008&\textbf{0.180$\pm$0.008}&0.181$\pm$0.008&0.197$\pm$0.016&0.222$\pm$0.017&0.349$\pm$0.008&0.515$\pm$0.002&\textbf{0.280$\pm$0.057}&0.561$\pm$0.002  \\ \hline
Regression & Medium low & Medium high & Low & Air quality (Dongsi) & \textbf{0.173$\pm$0.008}&\textbf{0.173$\pm$0.008}&\textbf{0.173$\pm$0.008}&0.186$\pm$0.014&0.245$\pm$0.018&0.279$\pm$0.003&0.517$\pm$0.002&\textbf{0.267$\pm$0.011}&0.544$\pm$0.001  \\ \hline
Regression & Medium low & Medium low & Low & Air quality (Guanyuan) & 0.168$\pm$0.011&\textbf{0.167$\pm$0.011}&0.168$\pm$0.011&0.18$\pm$0.019&0.239$\pm$0.032&0.265$\pm$0.004&0.518$\pm$0.000&\textbf{0.214$\pm$0.015}&0.571$\pm$0.001  \\ \hline
Regression & Medium low & Medium high & Low & Air quality (Tiantan) & \textbf{0.164$\pm$0.01}&\textbf{0.164$\pm$0.01}&\textbf{0.164$\pm$0.01}&0.182$\pm$0.013&0.191$\pm$0.023&0.266$\pm$0.013&0.515$\pm$0.000&\textbf{0.180$\pm$0.007}&0.559$\pm$0.007  \\ \hline
Regression & Medium low & Low & High & Indian weather (Lucknow) & \textbf{0.206$\pm$0.009}&\textbf{0.206$\pm$0.009}&\textbf{0.206$\pm$0.009}&0.215$\pm$0.012&0.268$\pm$0.026&0.215$\pm$0.002&0.450$\pm$0.000&\textbf{0.156$\pm$0.004}&0.441$\pm$0.000  \\ \hline
Regression & Low & Low & High & Indian weather (Mumbai) & \textbf{0.427$\pm$0.013}&\textbf{0.427$\pm$0.013}&0.428$\pm$0.013&0.456$\pm$0.022&0.506$\pm$0.036&0.629$\pm$0.004&0.588$\pm$0.000&\textbf{0.393$\pm$0.003}&0.579$\pm$0.000  \\ \hline
Regression & Low & Medium low & High & Indian weather (Rajasthan) & \textbf{0.213$\pm$0.012}&\textbf{0.213$\pm$0.012}&0.214$\pm$0.012&0.222$\pm$0.013&0.292$\pm$0.055&0.258$\pm$0.002&0.445$\pm$0.000&\textbf{0.163$\pm$0.001}&0.442$\pm$0.000  \\ \hline
Regression & Low & Low & High & Indian weather (Bhubhneshwar) & \textbf{0.427$\pm$0.013}&\textbf{0.427$\pm$0.013}&0.428$\pm$0.013&0.456$\pm$0.022&0.506$\pm$0.036&0.629$\pm$0.004&0.588$\pm$0.000&\textbf{0.393$\pm$0.003}&0.579$\pm$0.000  \\ \hline
Regression & Low & Low & High & Indian weather (Delhi) & \textbf{0.129$\pm$0.009}&\textbf{0.129$\pm$0.009}&\textbf{0.129$\pm$0.009}&0.143$\pm$0.011&0.180$\pm$0.023&0.102$\pm$0.001&0.401$\pm$0.000&\textbf{0.070$\pm$0.001}&0.398$\pm$0.000  \\ \hline
Regression & Low & Low & High & Indian weather (Chennai) & \textbf{0.220$\pm$0.009}&\textbf{0.220$\pm$0.009}&0.221$\pm$0.009&0.234$\pm$0.011&0.292$\pm$0.028&0.273$\pm$0.001&0.444$\pm$0.000&\textbf{0.183$\pm$0.001}&0.444$\pm$0.000  \\ \hline
Regression & Medium high & Medium low & Low & Taxi duration & 1.000$\pm$0.001&1.001$\pm$0.002&1.001$\pm$0.002&0.999$\pm$0.002&\textbf{0.997$\pm$0.000}&1.955$\pm$0.008&1.096$\pm$0.000&1.156$\pm$0.010&\textbf{1.018$\pm$0.000}  \\ \hline
Regression & Low & Medium low & Low & Traffic volume & N/A & N/A & N/A & N/A & N/A & 1.838$\pm$0.004 & \textbf{0.973$\pm$0.000} & 1.142$\pm$0.056 & 0.976$\pm$0.000  \\ \hline
Regression & Medium low & Medium low & Low & News popularity & \textbf{0.465$\pm$0.004}&\textbf{0.465$\pm$0.004}&\textbf{0.465$\pm$0.004}&N/A&0.473$\pm$0.011&0.705$\pm$0.000&0.572$\pm$0.000&0.523$\pm$0.000&\textbf{0.482$\pm$0.000}  \\ \hline
Regression & Medium high & High & Low & Beijing PM2.5 & \textbf{0.830$\pm$0.008}&0.836$\pm$0.012&\textbf{0.830$\pm$0.006}&0.857$\pm$0.012&0.988$\pm$0.164&1.464$\pm$0.002&\textbf{0.930$\pm$0.000}&1.265$\pm$0.006&0.960$\pm$0.000  \\ \hline
Regression & Medium high & Medium high & Low & Election & 0.506$\pm$0.173&N/A&0.513$\pm$0.154&0.566$\pm$0.312&\textbf{0.329$\pm$0.143}&0.023$\pm$0.001&0.353$\pm$0.000&\textbf{0.002$\pm$0.000}&0.313$\pm$0.004  \\ \hline
Regression & Medium high & High & High & 5 cities PM2.5 (Shenyang) & \textbf{0.173$\pm$0.016}&\textbf{0.173$\pm$0.015}&0.174$\pm$0.016&0.205$\pm$0.020&0.476$\pm$0.068&0.363$\pm$0.006&0.545$\pm$0.000&\textbf{0.277$\pm$0.021}&0.619$\pm$0.001  \\ \hline
Regression & High & Medium low & High & 5 cities PM2.5 (Guangzhou) & N/A &\textbf{0.419$\pm$0.107}&0.585$\pm$0.299&0.617$\pm$0.574&0.471$\pm$0.190&0.027$\pm$0.002&0.432$\pm$0.000&\textbf{0.012$\pm$0.001}&0.407$\pm$0.000  \\ \hline
Regression & Medium high & Medium high & High & 5 cities PM2.5 (Beijing) & N/A & N/A & N/A & N/A & N/A & 1.480$\pm$0.005&\textbf{0.896$\pm$0.000}&1.107$\pm$0.025&0.926$\pm$0.001  \\ \hline
Regression & Medium high & Medium low & High & 5 cities PM2.5 (Shanghai) & 0.869$\pm$0.003&0.870$\pm$0.004&0.868$\pm$0.004&\textbf{0.835$\pm$0.008}&0.926$\pm$0.021&1.629$\pm$0.005&\textbf{0.936$\pm$0.001}&1.116$\pm$0.040&0.974$\pm$0.000  \\ \hline


\end{tabular}
 }
\end{table*}

\subsection{Factors to Improve Learning on Data Streams}

\noindent
\begin{tcolorbox}[width=\linewidth,colback=white,boxrule=1pt,arc=0pt,outer arc=0pt,left=0pt,right=0pt,top=0pt,bottom=0pt,boxsep=1pt,halign=left]
\rev{\textbf{Finding (7):} Smaller window size, smaller batch size or more epochs can usually reduce loss while increasing computation costs. This indicates that more updates in training can generally improve model accuracy. In practice, we recommend to train more steps given enough computation resources. }
\end{tcolorbox}

\subsubsection{Number of Epochs} 
\begin{figure*}[ht]
    \centering
    \subfloat[ROOM]{\includegraphics[width=0.2\textwidth]{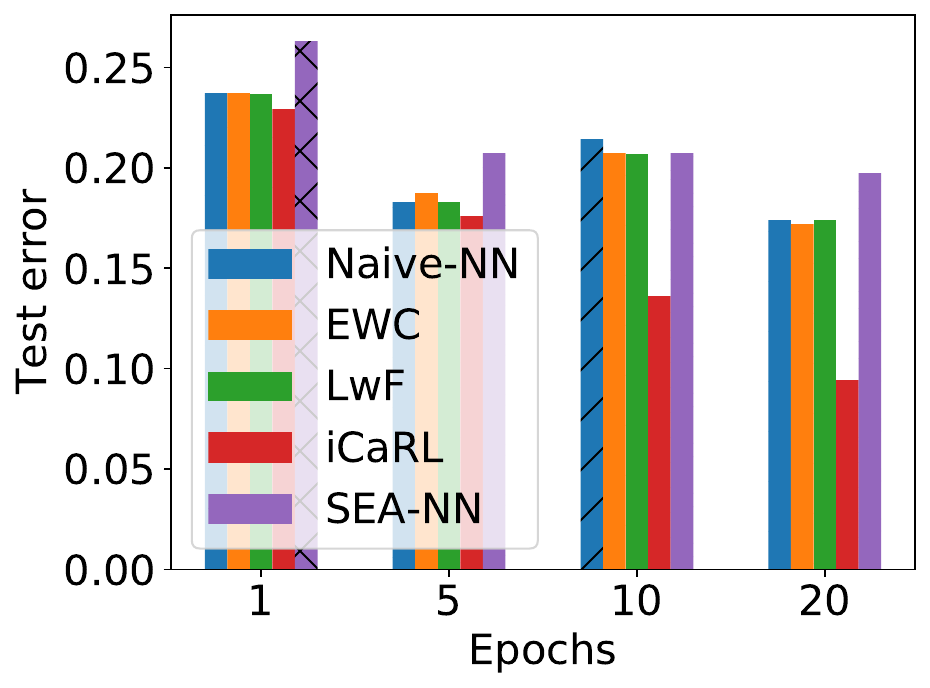}}
    \subfloat[ELECTRICITY]{\includegraphics[width=0.2\textwidth]{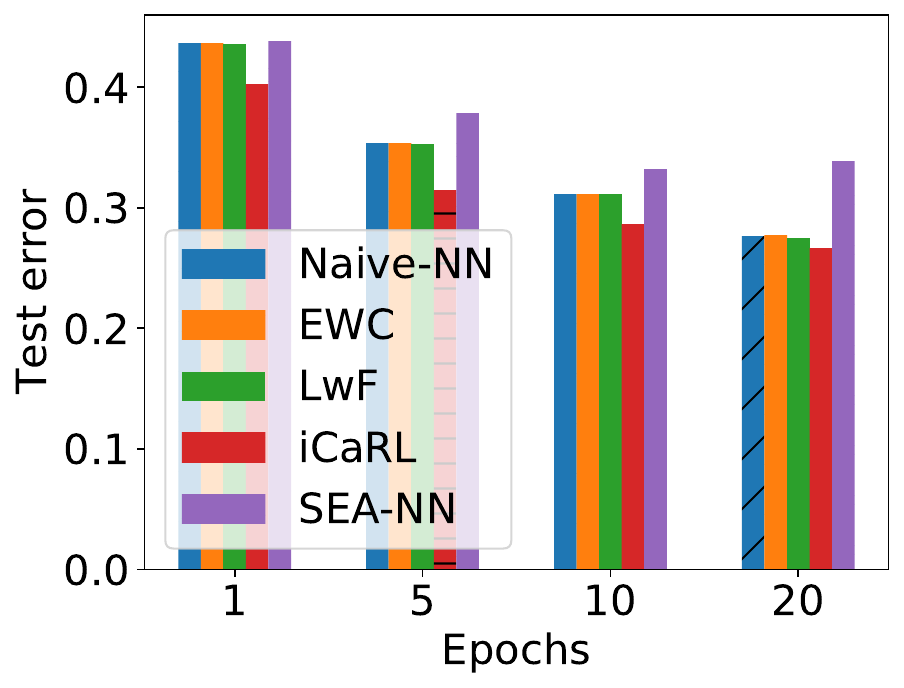}}
    \subfloat[INSECTS]{\includegraphics[width=0.2\textwidth]{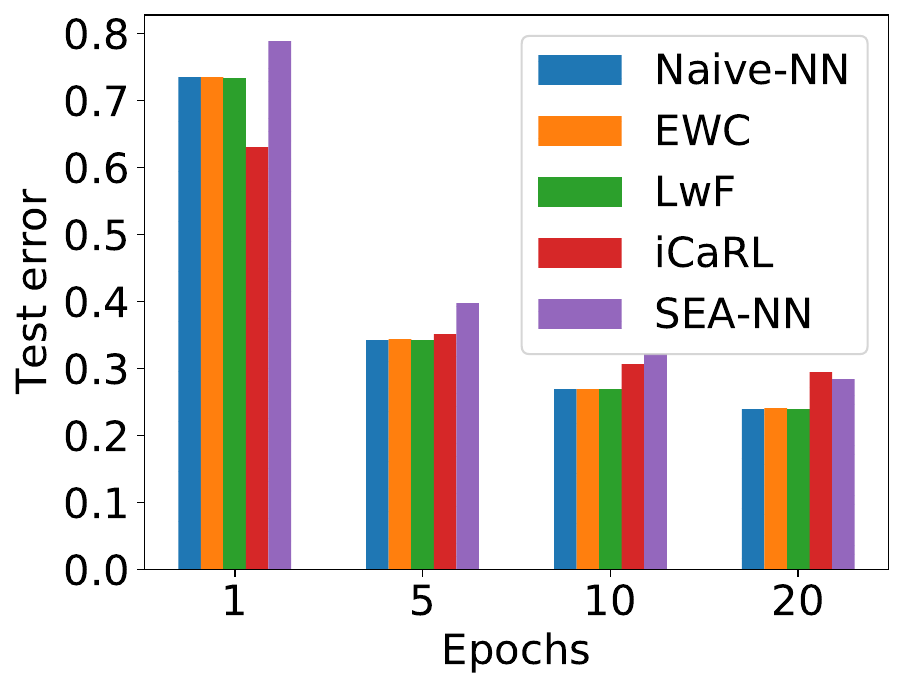}}
    \subfloat[AIR]{\includegraphics[width=0.2\textwidth]{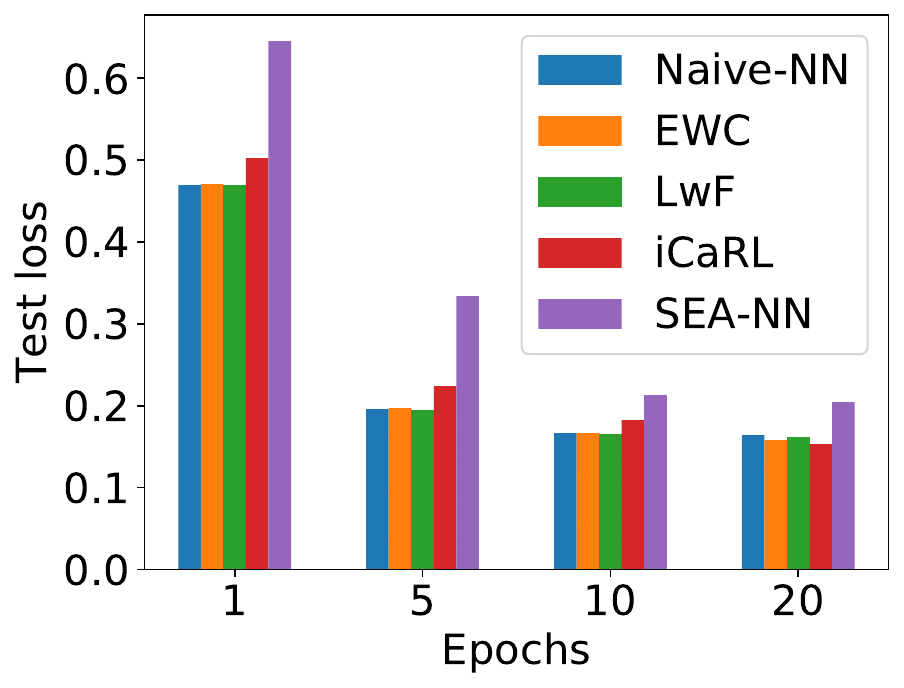}}
    \subfloat[POWER]{\includegraphics[width=0.2\textwidth]{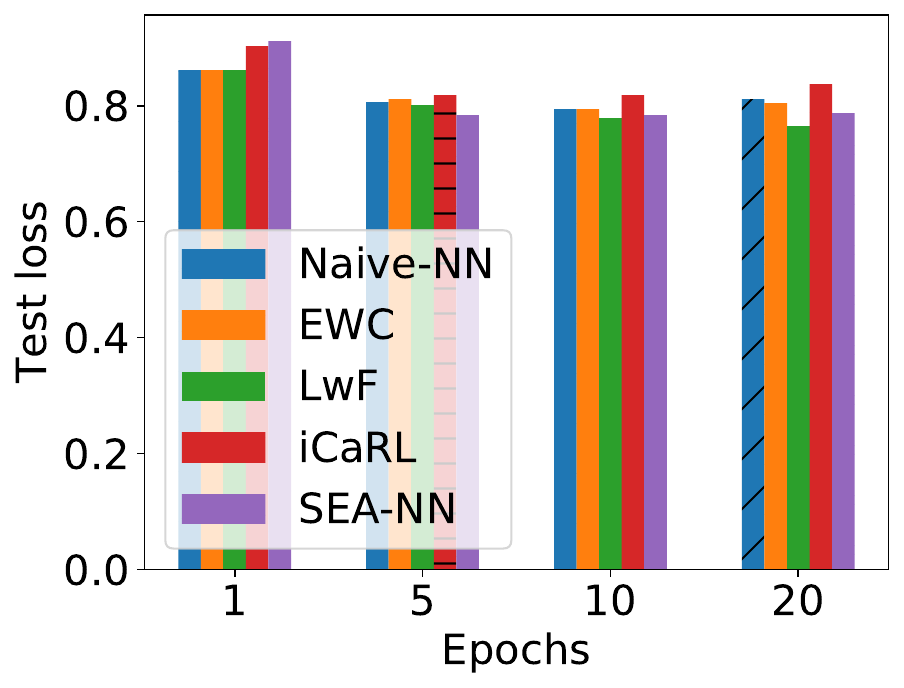}}
    \caption{Test error / test loss with different number of epochs per window.}
    \label{fig:epoch}
\end{figure*}

For NN-based techniques, our findings suggest that increasing the number of local epochs from 1 to 20 typically enhances model accuracy, as depicted in Figure \ref{fig:epoch}. However, in the case of the POWER dataset, training 10 epochs results in lower loss than 20 epochs for most algorithms. Given that the computational cost increases linearly with respect to the number of local epochs, we choose to train 10 epochs in our default setting.

\subsubsection{Window Size}

\begin{figure*}[htbp]
    \centering
    {\includegraphics[width=0.193\textwidth]{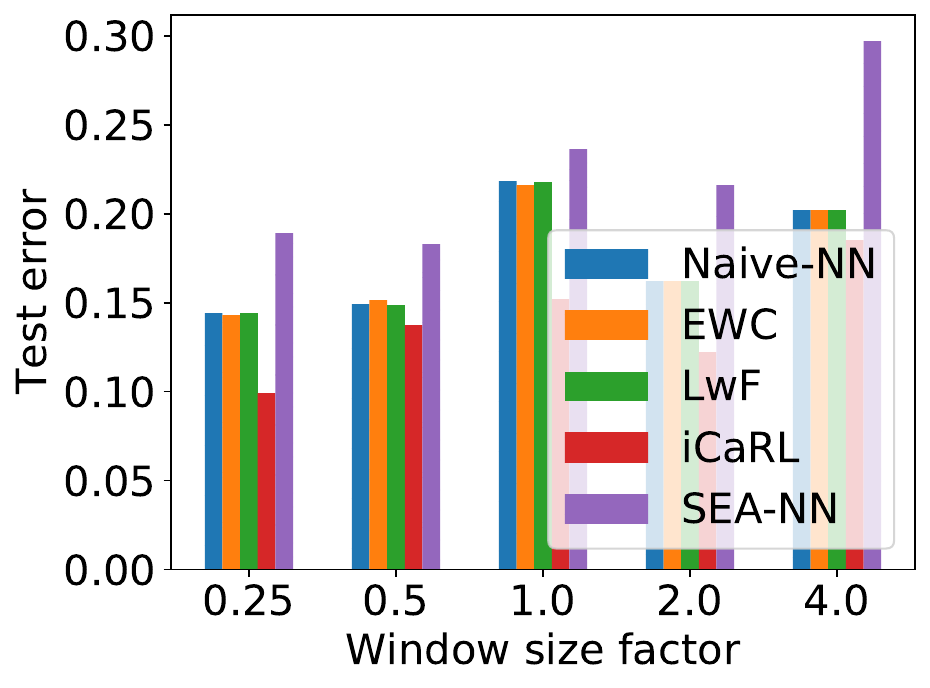}}
    {\includegraphics[width=0.193\textwidth]{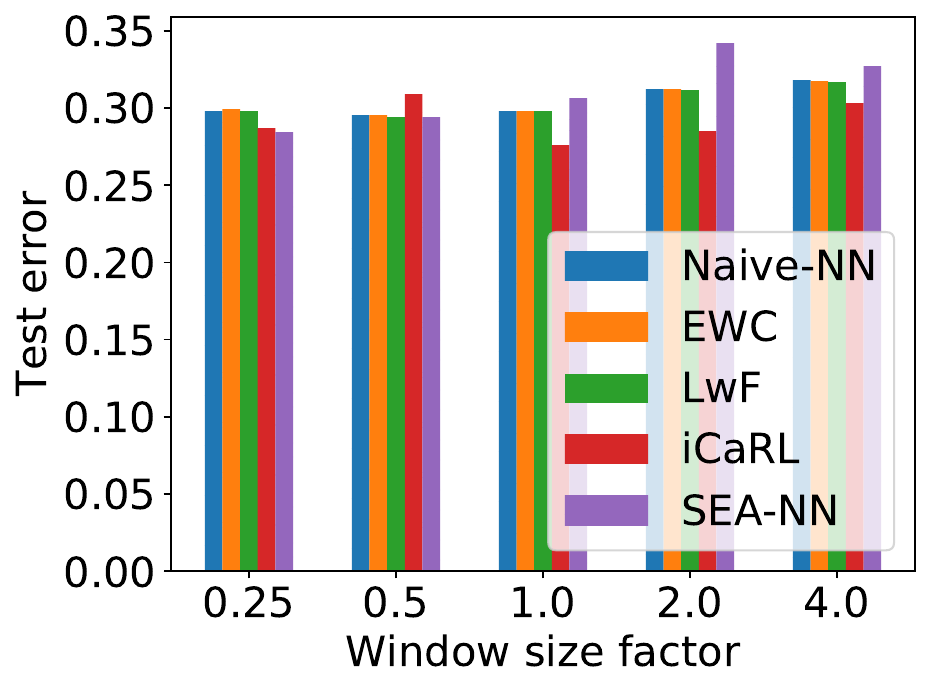}}
    {\includegraphics[width=0.193\textwidth]{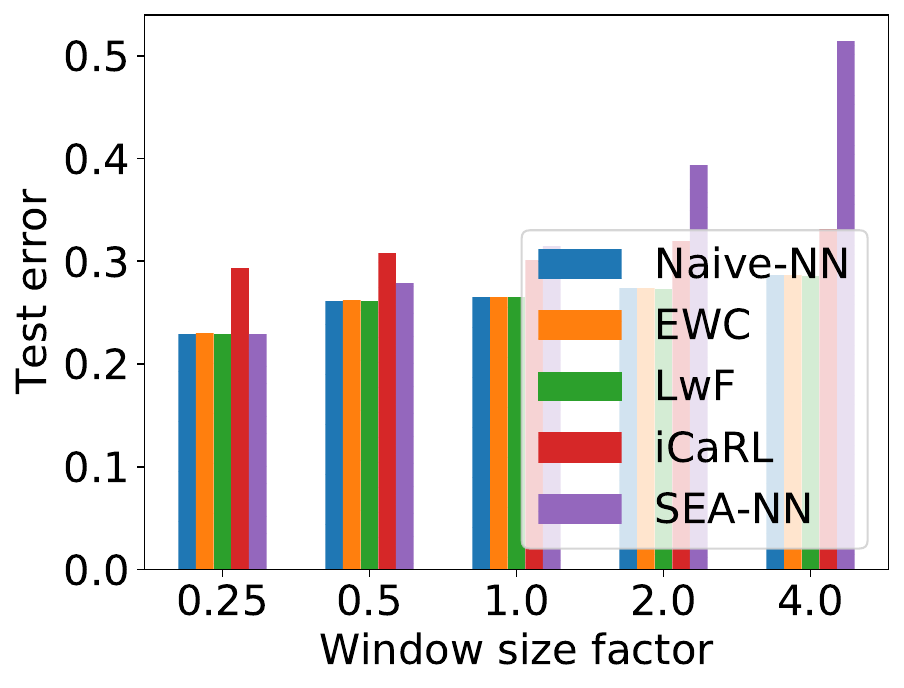}}
    {\includegraphics[width=0.193\textwidth]{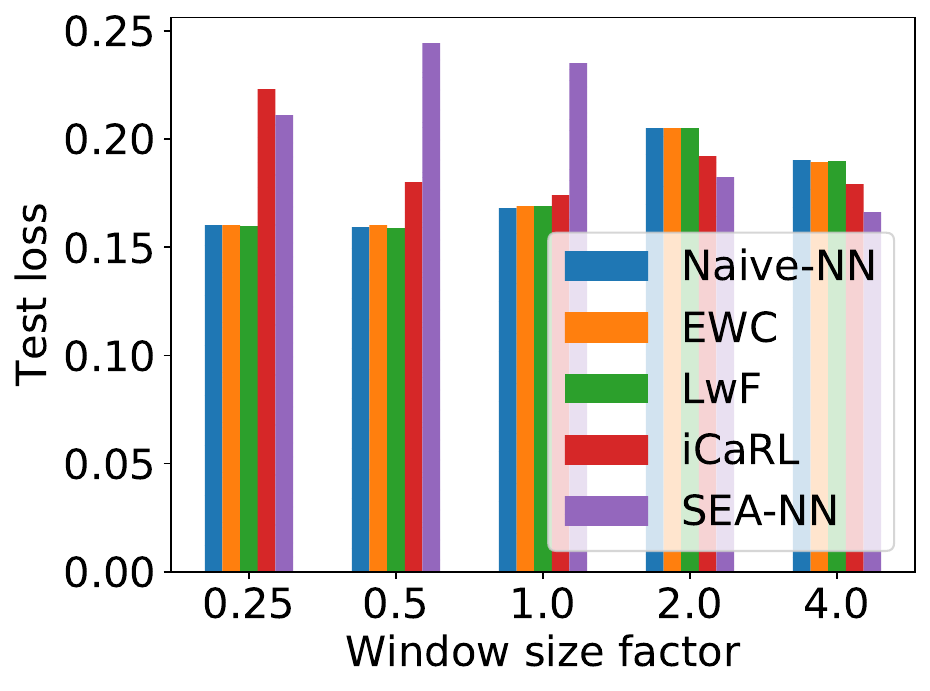}}
    {\includegraphics[width=0.193\textwidth]{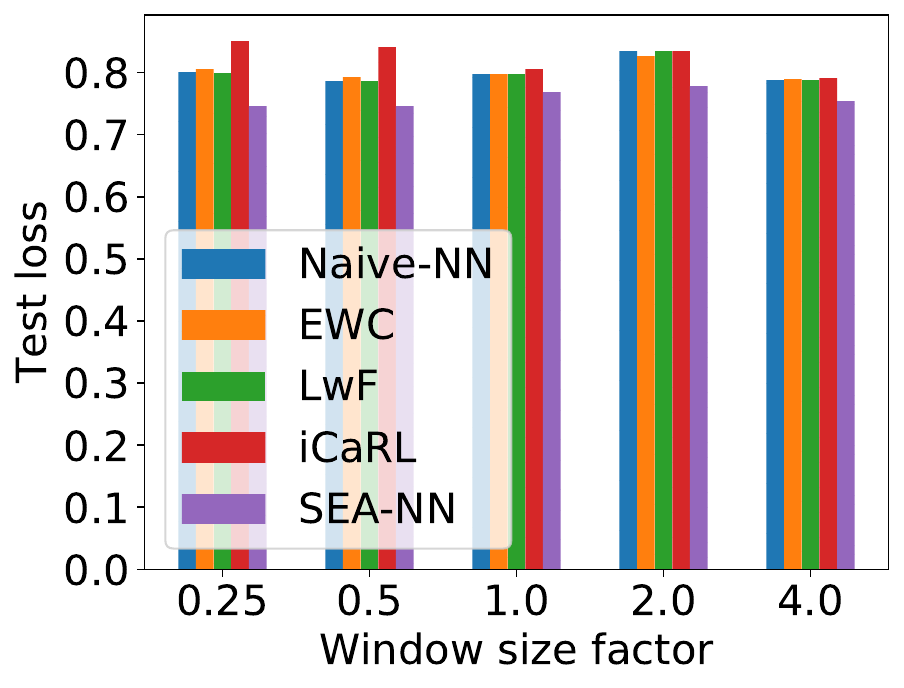}}
    \\
    \centering
    \subfloat[ROOM]{\includegraphics[width=0.2\textwidth]{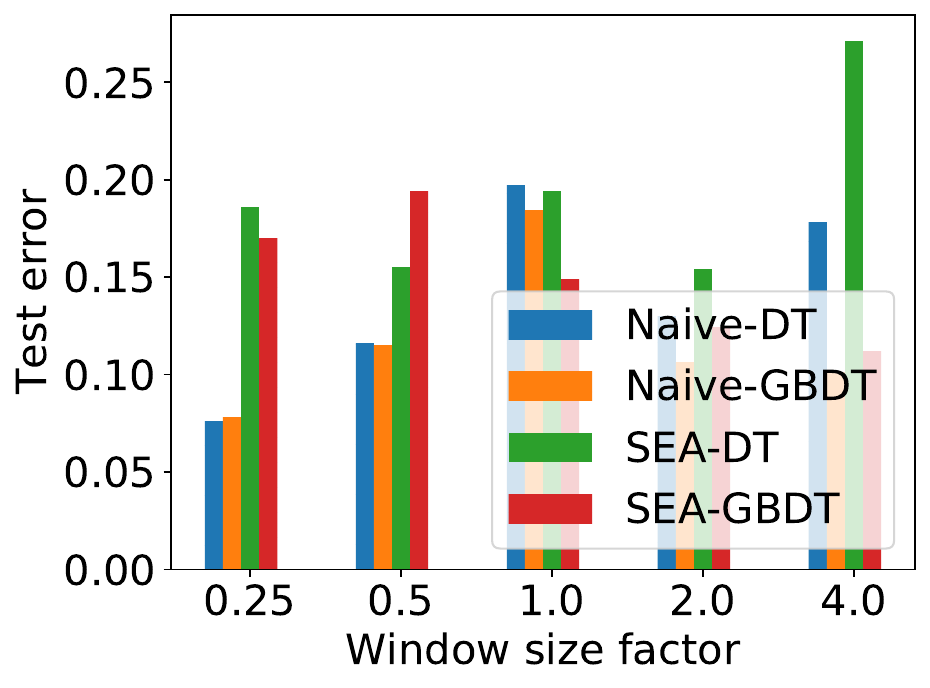}}
    \subfloat[ELECTRICITY]{\includegraphics[width=0.2\textwidth]{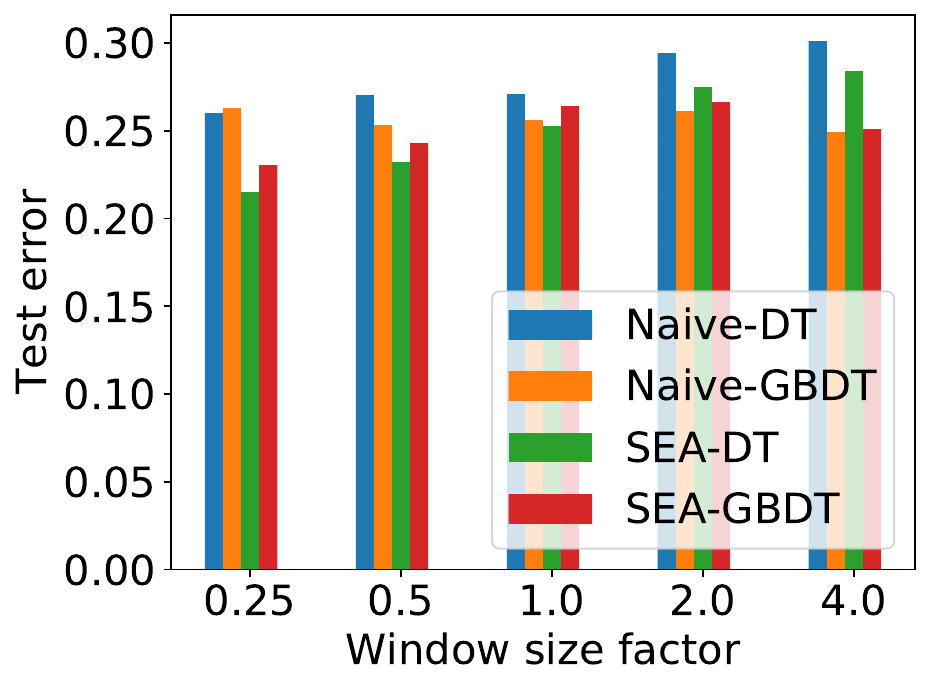}}
    \subfloat[INSECTS]{\includegraphics[width=0.2\textwidth]{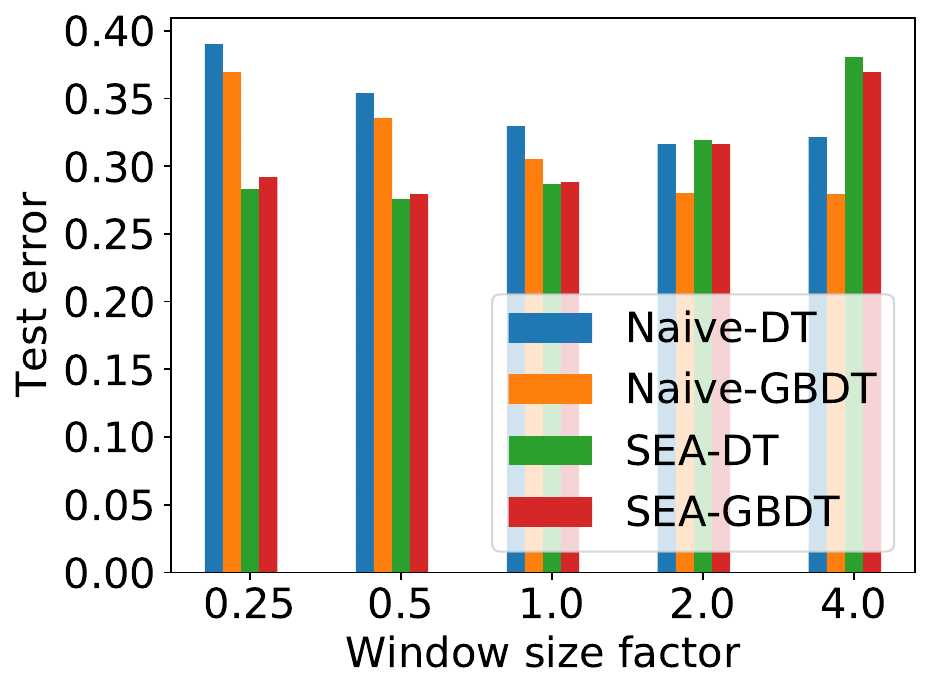}}
    \subfloat[AIR]{\includegraphics[width=0.2\textwidth]{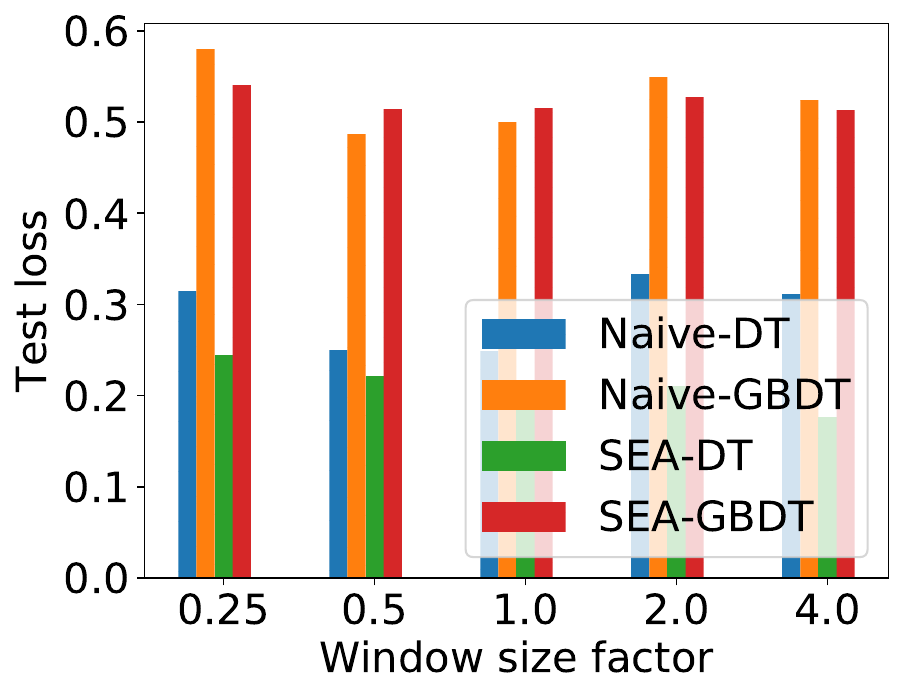}}
    \subfloat[POWER]{\includegraphics[width=0.2\textwidth]{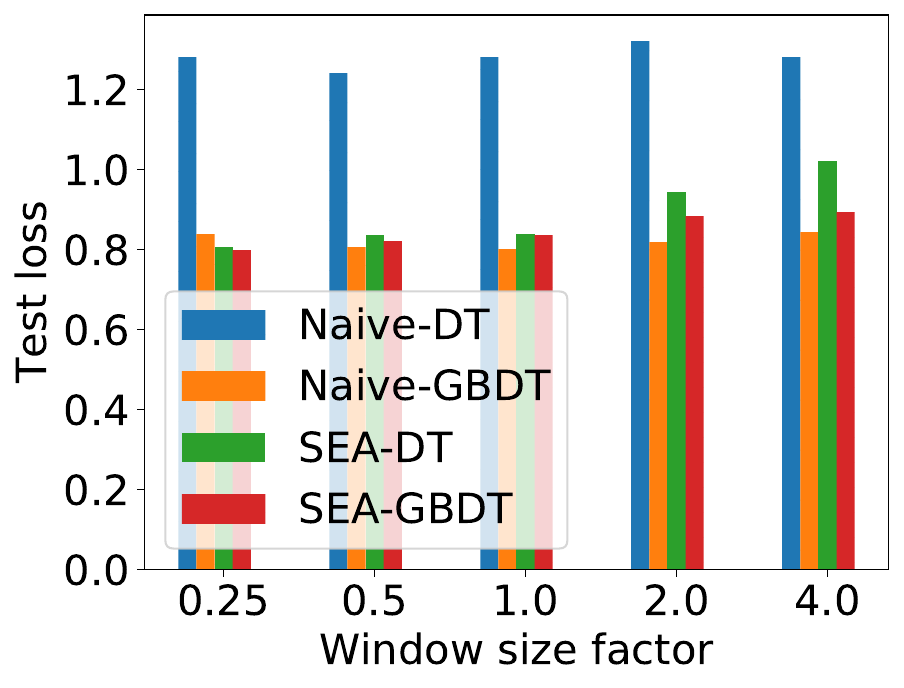}}
    \caption{Test error / test loss with different window size. The first row contains NN-based methods and the second row contains tree-based methods.}
    \label{fig:window}
\end{figure*}

In this part, we explore the impact of window size on model accuracy. The window size factor, which is adjusted within the range $\{0.25,0.5,1,2,4\}$, is multiplied by each dataset's default window size for our experiments. Figure \ref{fig:window} shows that a smaller window size generally reduces test loss in most cases. However, excessively small window sizes can harm model accuracy, as observed in the INSECTS dataset, potentially due to overfitting to the current window.

\subsubsection{Batch Size}

\begin{figure*}[ht]
    \centering
    \subfloat[ROOM]{\includegraphics[width=0.2\textwidth]{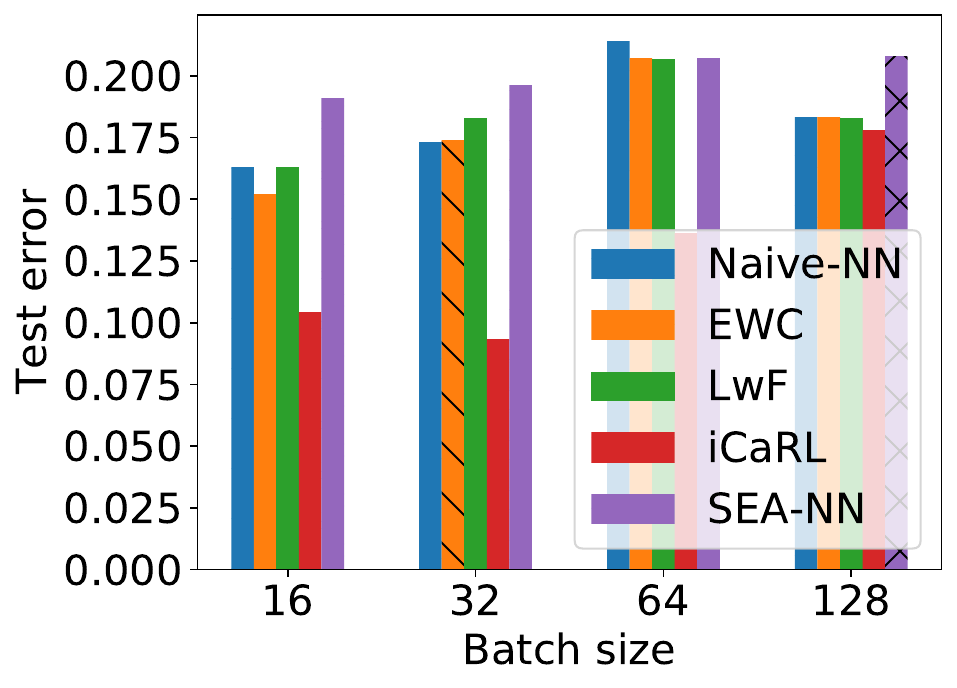}}
    \subfloat[ELECTRICITY]{\includegraphics[width=0.2\textwidth]{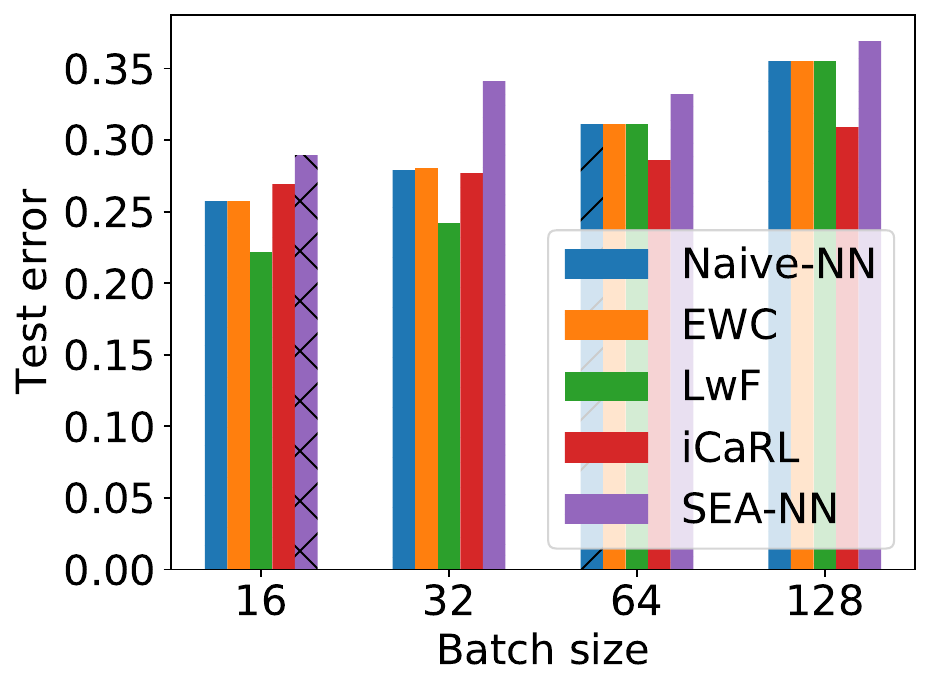}}
    \subfloat[INSECTS]{\includegraphics[width=0.2\textwidth]{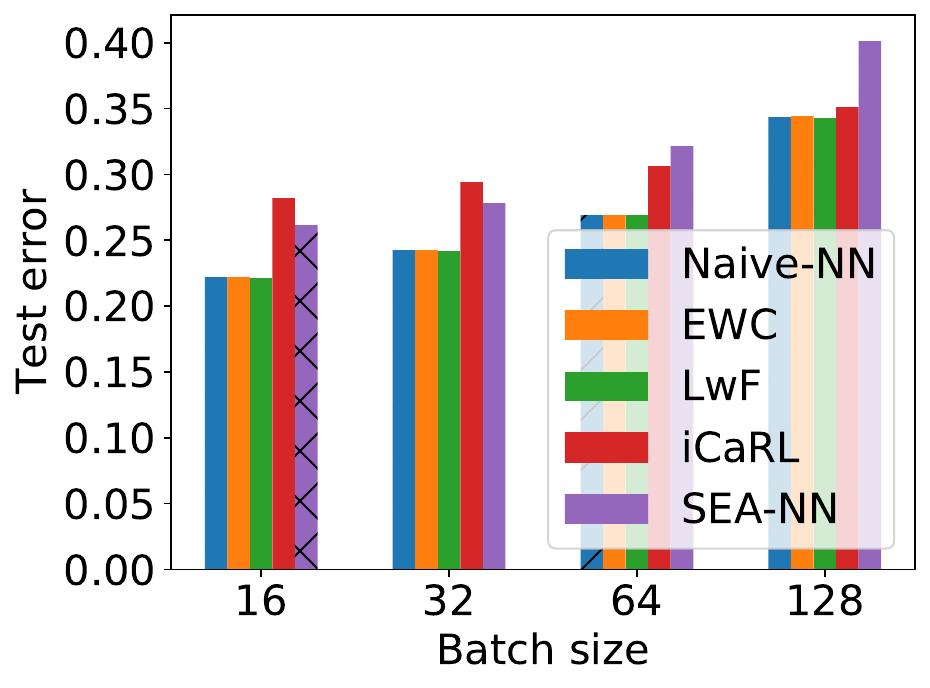}}
    \subfloat[AIR]{\includegraphics[width=0.2\textwidth]{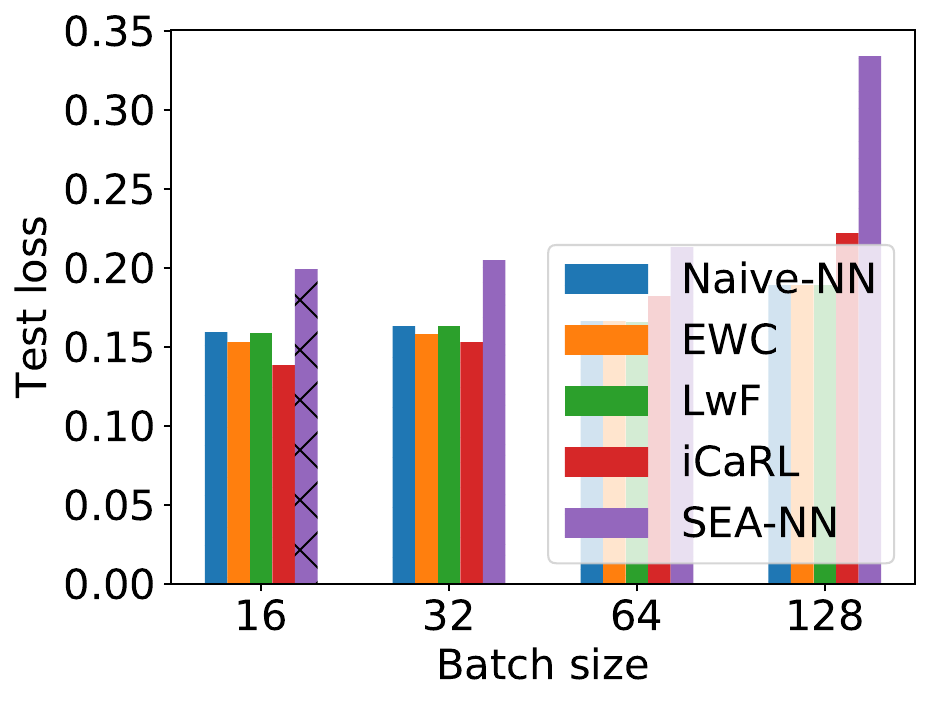}}
    \subfloat[POWER]{\includegraphics[width=0.2\textwidth]{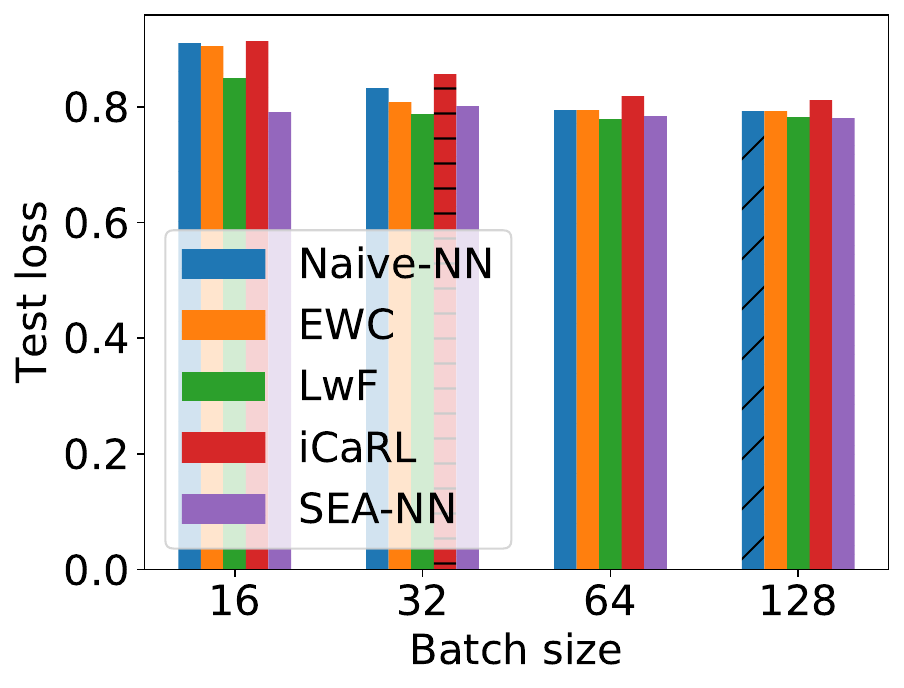}}
    \caption{Test error / test loss with different batch size.}
    \label{fig:batchsize}
\end{figure*}

In this part, we focus on understanding the role of batch size in model accuracy. Keeping the number of epochs fixed at 10, we vary the batch size across $\{16,32,64,128\}$. Figure \ref{fig:batchsize} shows that a smaller batch size enhances model accuracy in four out of the five datasets, except the POWER dataset. Given that smaller batch size inherently increases computational demands, we generally recommend to choose batch size as 64 or 32.

\subsection{Effect of Larger Models}
\label{sec:large}
\noindent
\begin{tcolorbox}[width=\linewidth,colback=white,boxrule=1pt,arc=0pt,outer arc=0pt,left=0pt,right=0pt,top=0pt,bottom=0pt,boxsep=1pt,halign=left]
\rev{\textbf{Finding (8):} Decision trees or lightweight neural networks already achieve good model accuracy on real-world relational data streams. Large models do not necessarily lead to better accuracy. A possible reason is that large models are prone to overfitting and hard to generalize towards the new environment. When it comes to practical relational data streams, we recommend to consider lightweight models. }
\end{tcolorbox}

\begin{figure*}[h]
    \centering
    \subfloat[ROOM]{\includegraphics[width=0.2\textwidth]{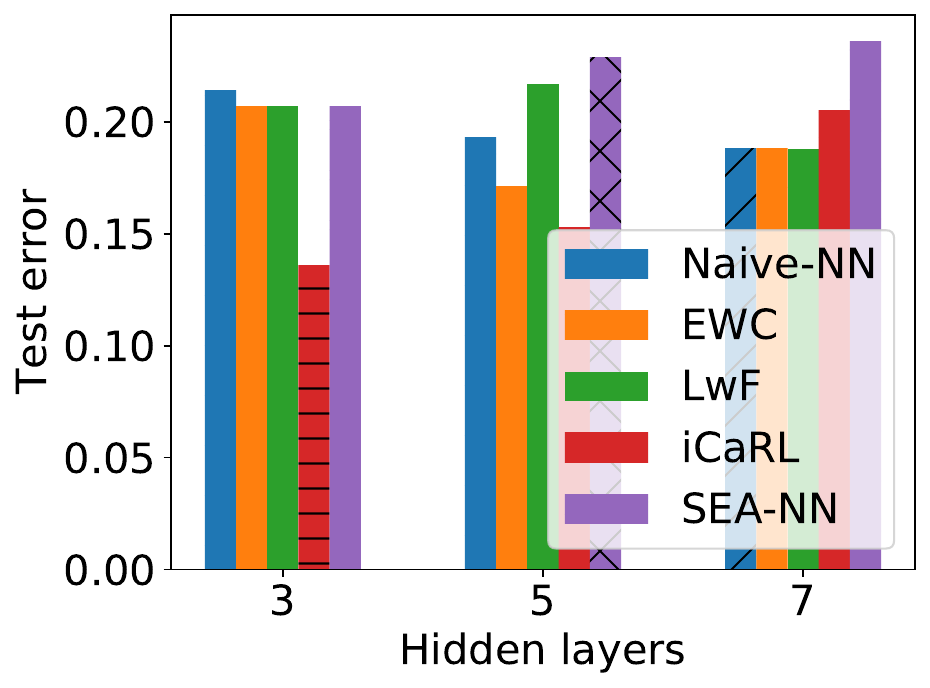}}
    \subfloat[ELECTRICITY]{\includegraphics[width=0.2\textwidth]{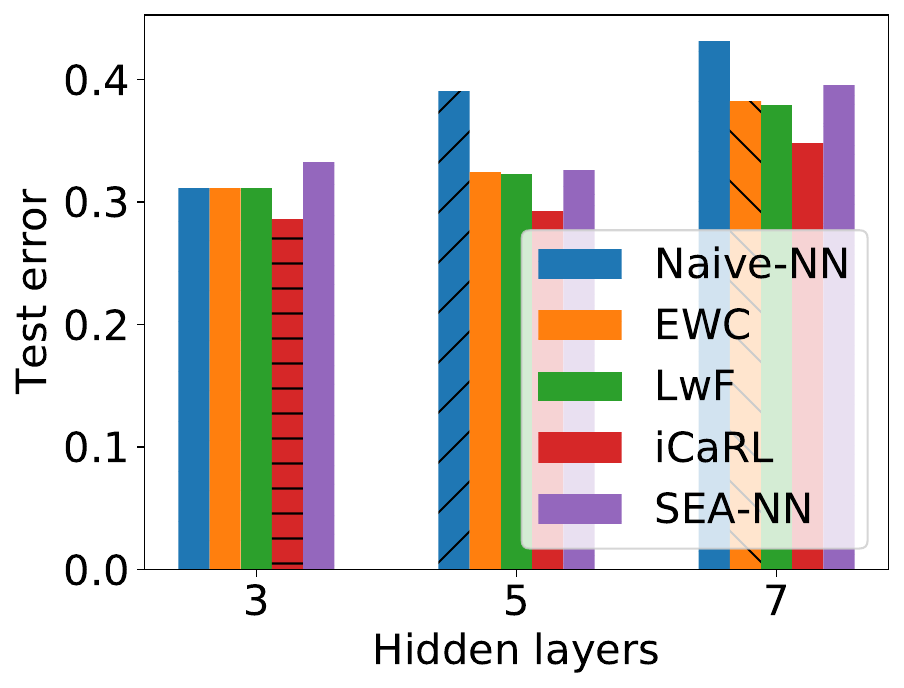}}
    \subfloat[INSECTS]{\includegraphics[width=0.2\textwidth]{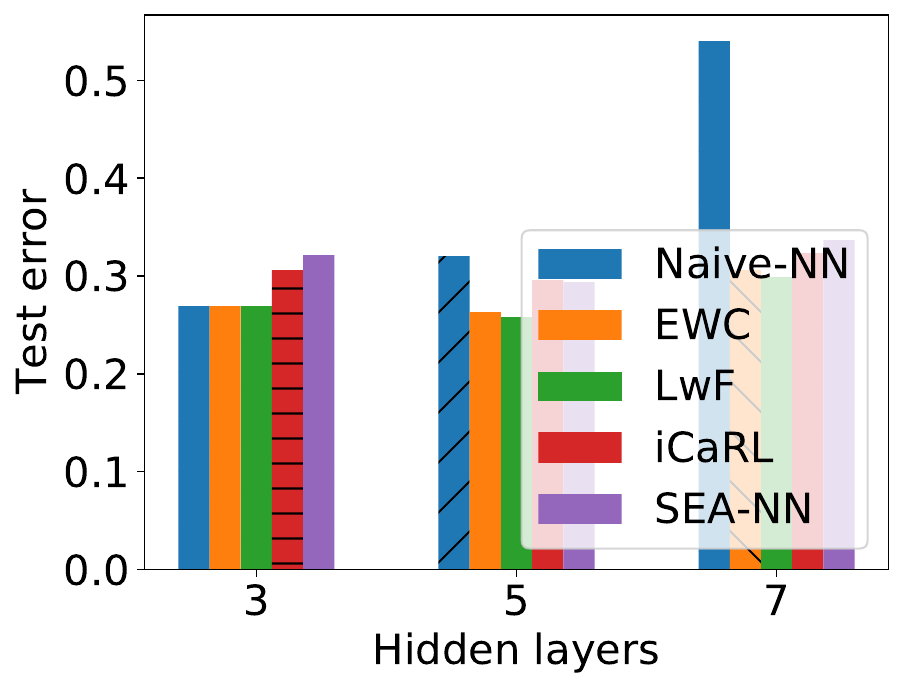}}
    \subfloat[AIR]{\includegraphics[width=0.2\textwidth]{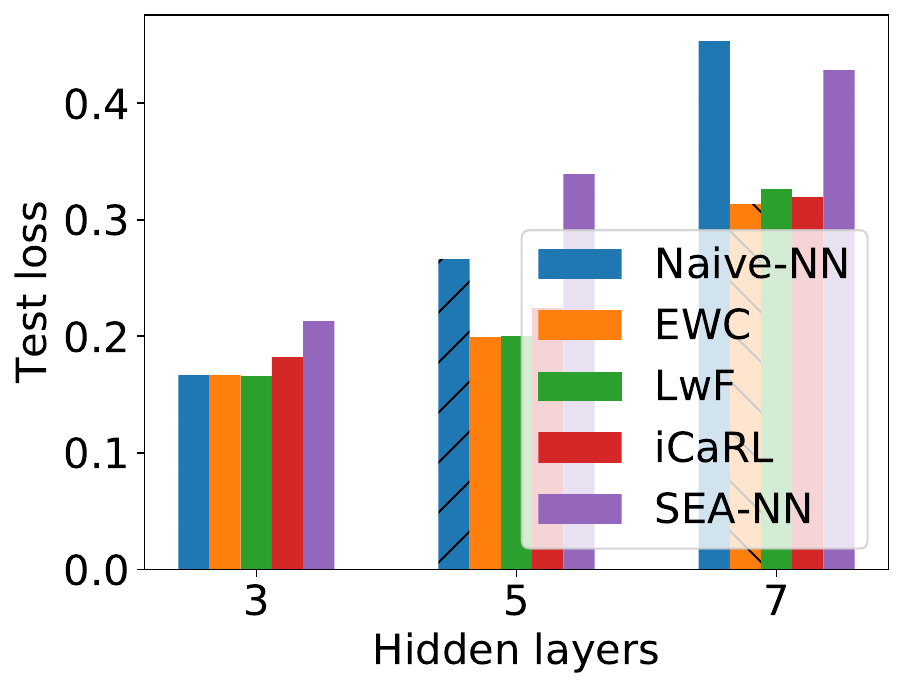}}
    \subfloat[POWER]{\includegraphics[width=0.2\textwidth]{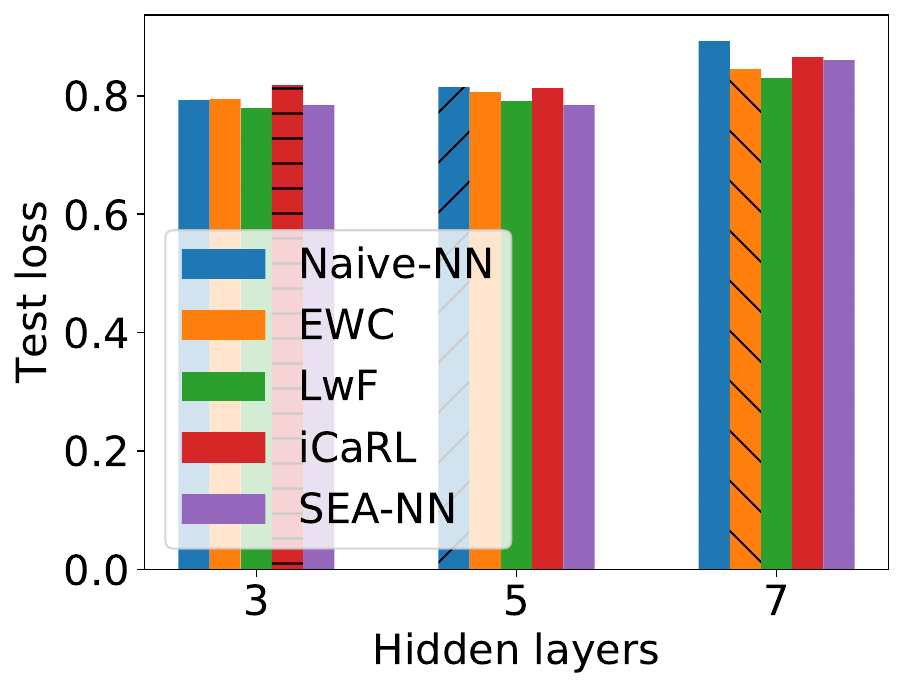}}
    \caption{Test error / test loss on NN with different number of hidden layers.}
    \label{fig:heavy}
\end{figure*}

In this section, we experiment with larger neural network models, in particular training MLPs with 3, 5, and 7 layers. The number of neurons in each layer is (32, 16, 8), (32, 32, 16, 16, 8), and (32, 32, 32, 16, 16, 16, 8), respectively. As shown in Figure \ref{fig:heavy}, larger neural networks perform worse in most cases, since larger neural networks tend to overfit the data of the current window. 

\begin{figure}[h]
    \centering
    \subfloat[ROOM]{\includegraphics[width=0.23\textwidth]{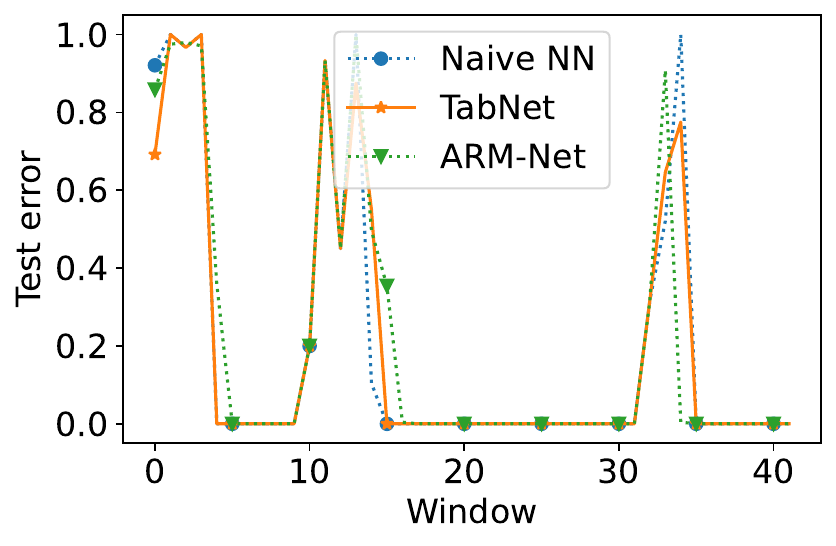}}
    \subfloat[AIR]{\includegraphics[width=0.23\textwidth]{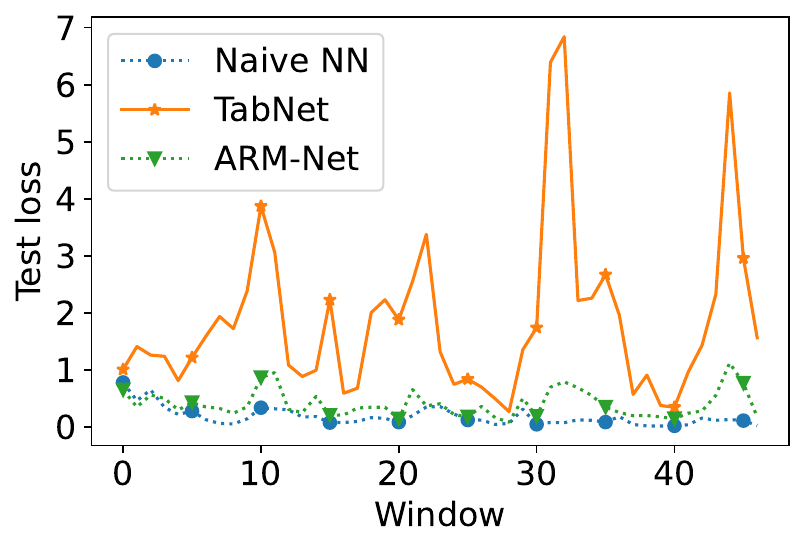}}
    \caption{Test error / test loss on TabNet and ARM-Net compared with naive NN.}
    \label{fig:tabarm}
\end{figure}

\vl{We also show the training curves of advanced models TabNet and ARM-Net in Figure \ref{fig:tabarm}. Compared with three-layer NN, both TabNet and ARM-Net achieve similar model accuracy in the ROOM dataset, while much worse accuracy in the regression task ELECTRICITY.} Therefore, we recommend lightweight NN models or tree models for real-world relational data streams.

\subsection{Effect of Regularization Factor}
\begin{figure}[h]
    \centering
    \subfloat[EWC]{\includegraphics[width=0.25\textwidth]{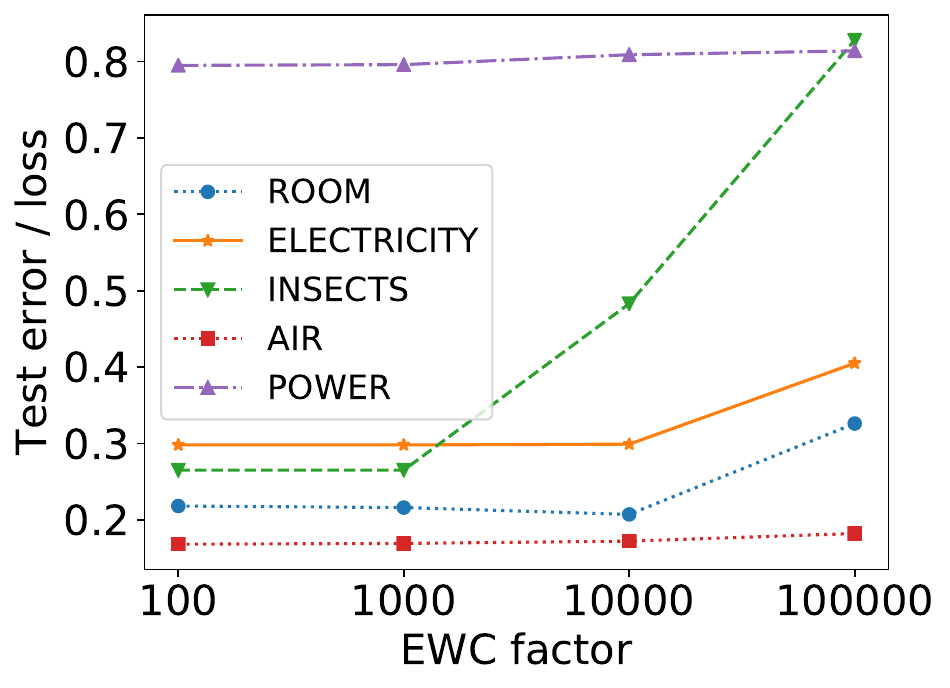}}
    \subfloat[LwF]{\includegraphics[width=0.24\textwidth]{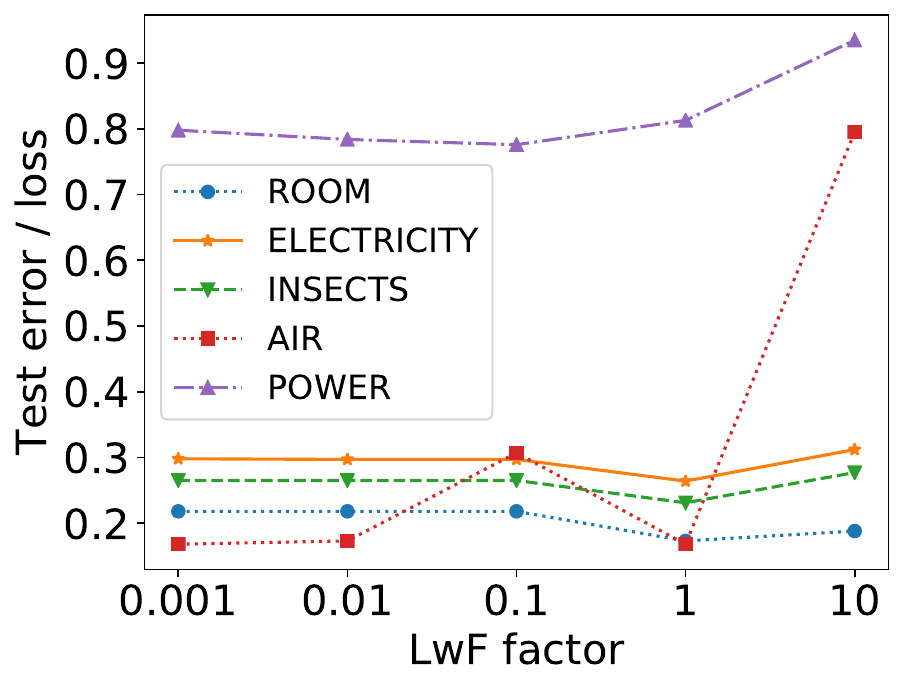}}
    \caption{Test error / test loss on different regularization weights.}
    \label{fig:reg}
\end{figure}

In this section, we explore the effect of regularization factor in EWC and LwF. For EWC, the regularization factor is varied in $\{10^2,10^3,10^4,10^5\}$. For LwF, we test among $\{0.001,0.01,0.1,1,10\}$. Results are shown in Figure \ref{fig:reg}. As illustrated in Figure \ref{fig:reg}, EWC generally exhibits improved model accuracy when the regularization factor is set at 100 or 1000, while LwF performs the best with a regularization factor of 0.01. It is notable that a too large regularization factor can degrade the model accuracy.

\subsection{Effect of Exemplar Buffer Size}

\noindent
\begin{tcolorbox}[width=\linewidth,colback=white,boxrule=1pt,arc=0pt,outer arc=0pt,left=0pt,right=0pt,top=0pt,bottom=0pt,boxsep=1pt,halign=left]
\rev{\textbf{Finding (9):} Large buffer size or ensemble size increase the memory and computation costs, but do not necessarily boost model accuracy. This verifies that memorizing more old data is not always effective in open environment learning. }
\end{tcolorbox}

\begin{figure}[h]
    \centering
    \includegraphics[width=0.6\columnwidth]{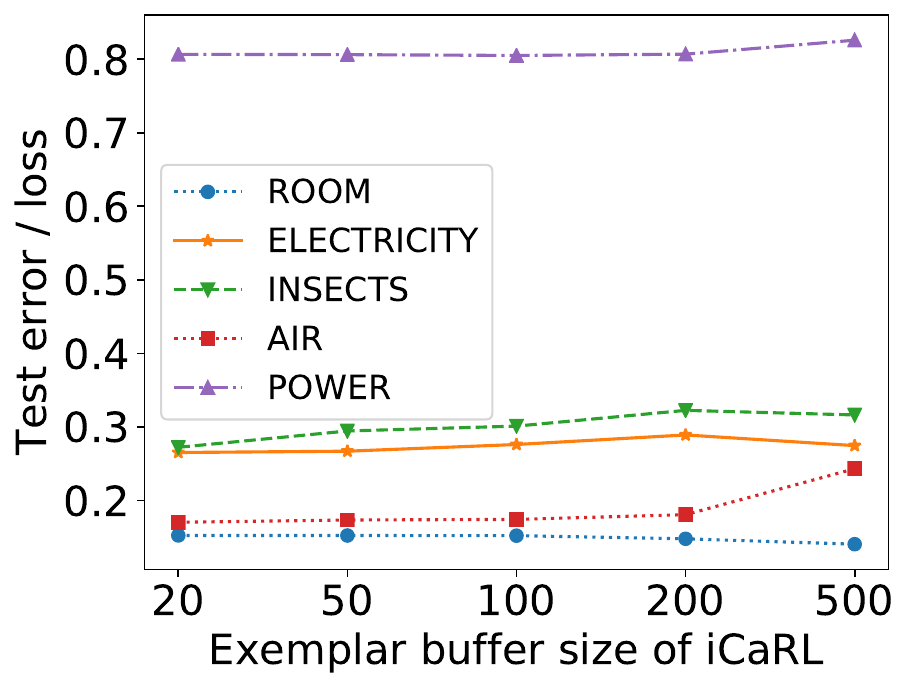}
    \caption{Test error / test loss on different buffer size of iCaRL.}
    \label{fig:buffer}
\end{figure}

In this section, we explore the exemplar buffer size in iCaRL. The buffer sizes tested are $\{20,50,100,200,500\}$. Figure \ref{fig:buffer} illustrates that the exemplar buffer size does not drastically affect the test loss. However, a considerably large buffer size, like 500, could lead to a higher loss, suggesting that more exemplars does not always translate to better model accuracy. In light of this, we suggest a relatively smaller buffer size to improve efficiency.

\subsection{Effect of Ensemble Size}
\begin{figure*}[h]
    \centering
    \subfloat[ROOM]{\includegraphics[width=0.2\textwidth]{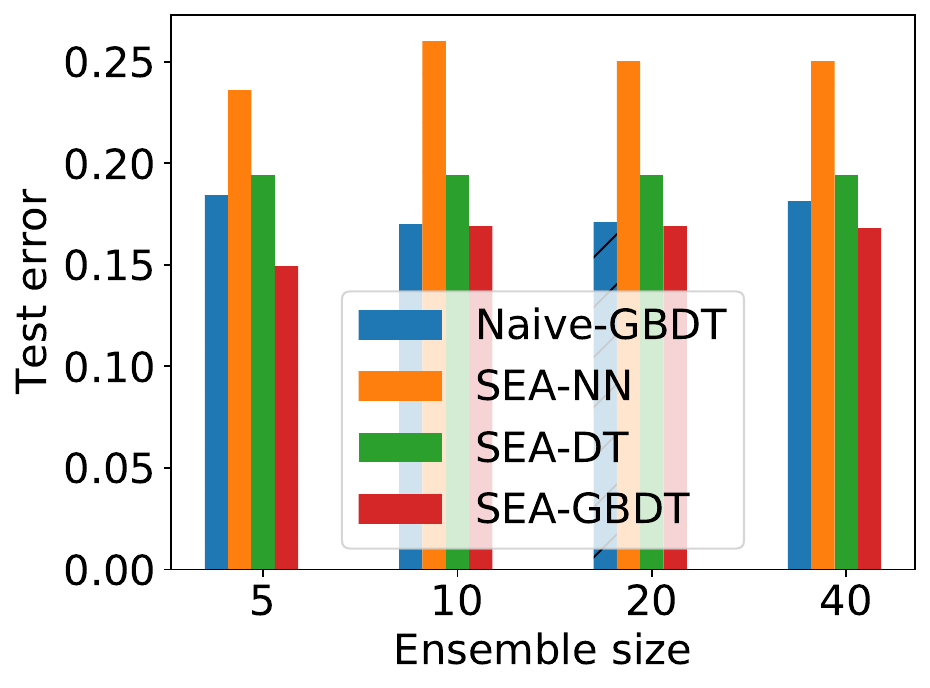}}
    \subfloat[ELECTRICITY]{\includegraphics[width=0.2\textwidth]{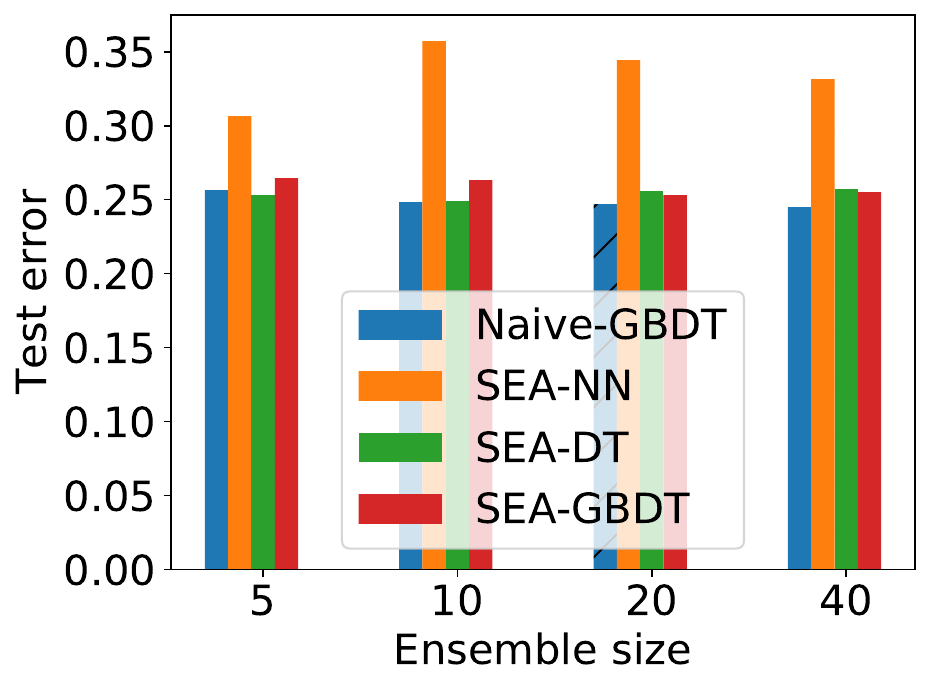}}
    \subfloat[INSECTS]{\includegraphics[width=0.2\textwidth]{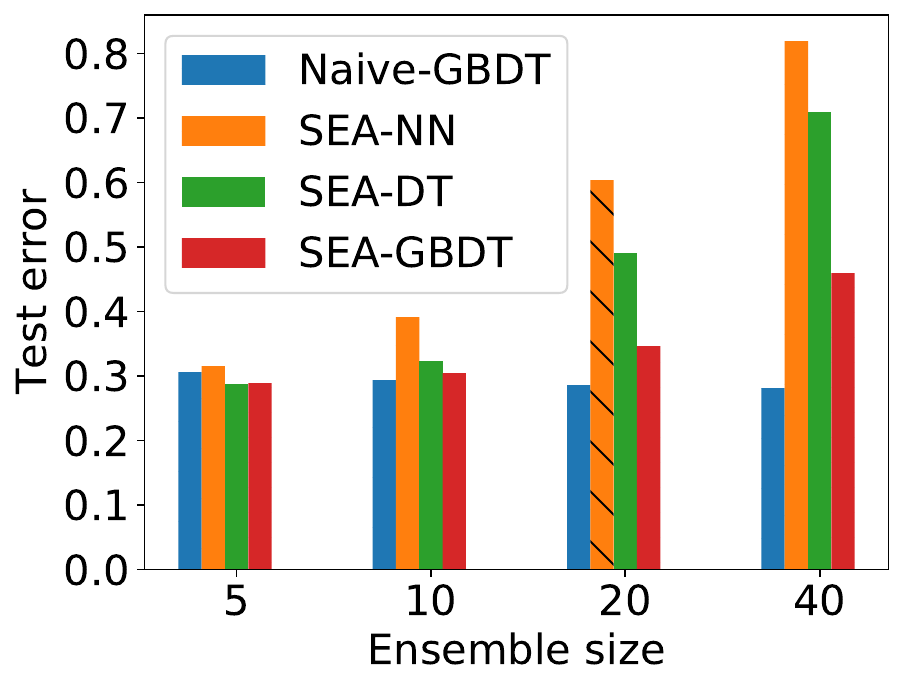}}
    \subfloat[AIR]{\includegraphics[width=0.2\textwidth]{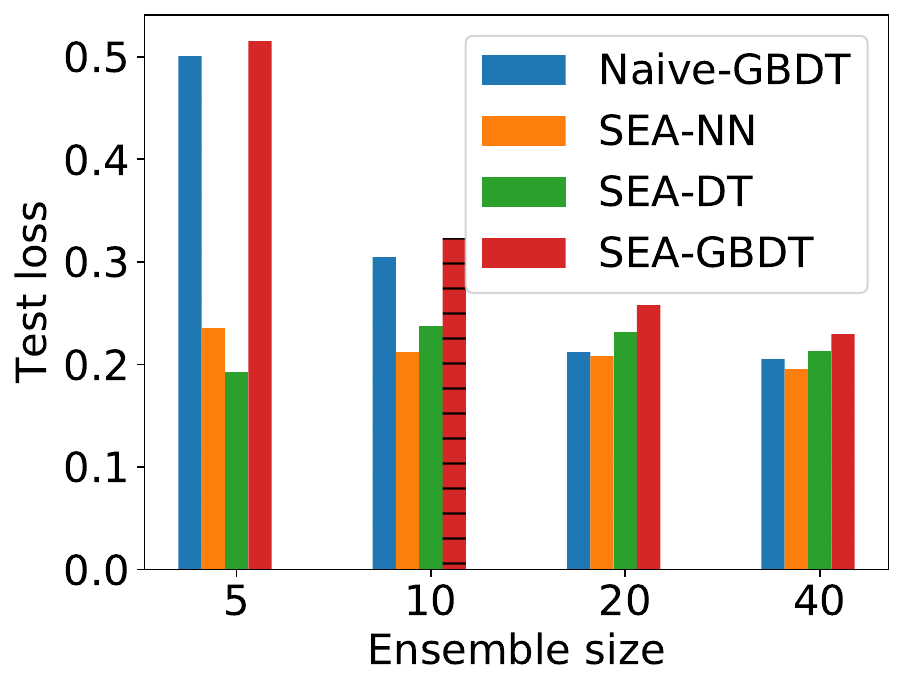}}
    \subfloat[POWER]{\includegraphics[width=0.2\textwidth]{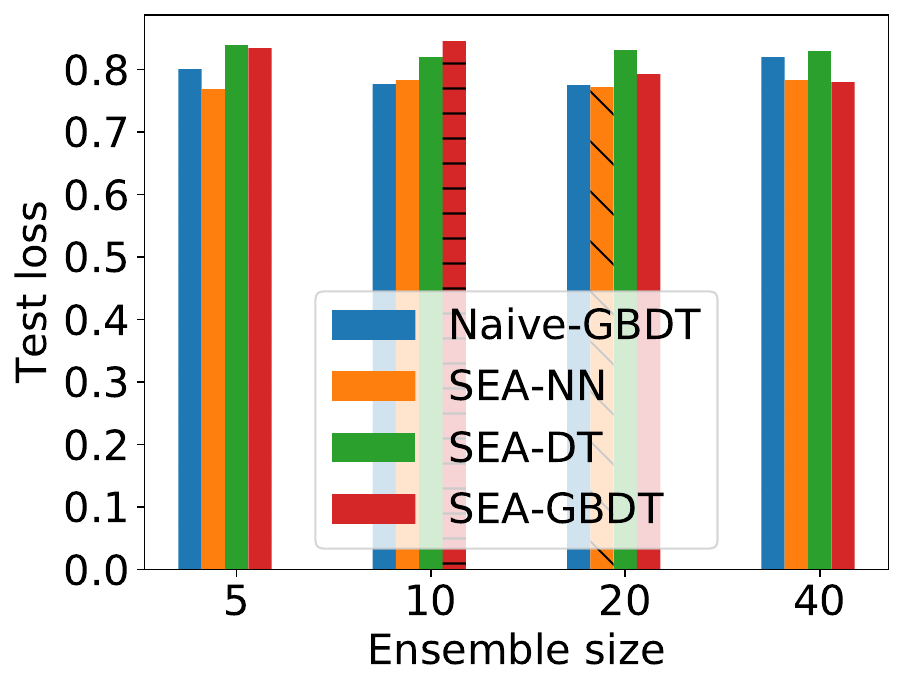}}
    \caption{Test error / test loss on GBDT and SEA with varying ensemble size.}
    \label{fig:ensemble}
\end{figure*}

We evaluate the model accuracy of GBDT and SEA with ensemble size varying among $\{5,10,20,40\}$. ARF is excluded due to its relative inefficiency without significant improvement in model accuracy. As shown in Figure \ref{fig:ensemble}, a naive GBDT generally performs better with a larger ensemble size, whereas SEA exhibits different trends depending on the dataset. For instance, a larger ensemble size leads to higher loss in the INSECTS dataset, yet results in lower loss in the AIR dataset. Therefore, a larger ensemble size does not always improve ensemble model accuracy.

\subsection{Running Time}
We show the running time of these algorithms in Table \ref{tbl:time}. As we can see, decision trees are more efficient than NN-based methods. Therefore, we recommend to apply decision trees when high throughput is required.

\begin{table*}[h]
\centering
\caption{Running time in seconds of explored stream learning algorithms on selected real-world datasets. All experiments run on 4 CPU threads. Lower value indicates better performance. We report the average of three runs.}
\label{tbl:time}
\resizebox{2.1\columnwidth}{!}{
\begin{tabular}{|c|c|c|c|c|c|c||c|c|c|c|c|c|}
\hline
Dataset & Epochs & Naive-NN & EWC & LwF & iCaRL & SEA-NN & Naive-DT & Naive-GBDT & ARF & SEA-DT & SEA-GBDT \\ \hline
\multirow{4}{*}{ROOM} & 1 & 0.16 & 0.38 & 0.32 & 0.33 & 0.38 & \multirow{4}{*}{0.05} & \multirow{4}{*}{0.23} & \multirow{4}{*}{43.22} & \multirow{4}{*}{0.37} & \multirow{4}{*}{0.37} \\ \cline{2-7}
& 5 & 0.72 & 1.64 & 1.31 & 1.38 & 0.84 &&&&& \\ \cline{2-7}
& 10 & 1.24 & 3.00 & 2.18 & 2.42 & 1.43 &&&&& \\ \cline{2-7}
& 20 & 2.75 & 6.20 & 5.54 & 5.55 & 3.21 &&&&&  \\ \hline
\multirow{4}{*}{ELECTRICITY} & 1 & 0.64 & 1.31 & 1.14 & 1.53 & 1.04 & \multirow{4}{*}{0.17} & \multirow{4}{*}{0.63} & \multirow{4}{*}{141.55} & \multirow{4}{*}{0.89} & \multirow{4}{*}{1.61}  \\ \cline{2-7}
& 5 & 2.72 & 6.00 & 5.38 & 5.35 & 3.58 &&&&& \\ \cline{2-7}
& 10 & 5.41 & 13.18 & 9.44 & 9.91 & 6.04 &&&&& \\ \cline{2-7}
& 20 & 10.86 & 25.05 & 22.81 & 22.02 & 12.81 &&&&& \\ \hline
\multirow{4}{*}{INSECTS} & 1 & 1.64 & 4.71 & 2.98 & 2.81 & 4.73 & \multirow{4}{*}{1.44} & \multirow{4}{*}{25.42} & \multirow{4}{*}{1831.80} & \multirow{4}{*}{11.14} & \multirow{4}{*}{46.68}  \\ \cline{2-7}
& 5 & 6.49 & 15.05 & 12.27 & 12.37 & 10.31 &&&&& \\ \cline{2-7}
& 10 & 13.06 & 30.41 & 21.66 & 23.72 & 15.83 &&&&& \\ \cline{2-7}
& 20 & 24.12 & 54.45 & 49.81 & 49.67 & 30.75 &&&&&  \\ \hline
\multirow{4}{*}{AIR} & 1 & 0.81 & 1.41 & 1.29 & 1.55 & 1.07 & \multirow{4}{*}{0.73} & \multirow{4}{*}{1.08} & \multirow{4}{*}{N/A} & \multirow{4}{*}{1.68} & \multirow{4}{*}{2.29}  \\ \cline{2-7}
& 5 & 2.64 & 4.68 & 4.18 & 4.20 & 3.09 &&&&&  \\ \cline{2-7}
& 10 & 4.46 & 9.30 & 6.85 & 7.85 & 4.75 &&&&& \\ \cline{2-7}
& 20 & 8.42 & 15.12 & 13.54 & 13.92 & 9.01 &&&&&  \\ \hline
\multirow{4}{*}{POWER} & 1 & 0.63 & 1.39 & 1.21 & 1.20 & 0.75 & \multirow{4}{*}{0.31} & \multirow{4}{*}{0.39} & \multirow{4}{*}{N/A} & \multirow{4}{*}{0.44} & \multirow{4}{*}{0.57}   \\ \cline{2-7}
& 5 & 3.16 & 5.64 & 5.25 & 5.11 & 3.32 &&&&& \\ \cline{2-7}
& 10 & 5.66 & 10.54 & 8.30 & 9.12 & 5.82 &&&&& \\ \cline{2-7}
& 20 & 9.77 & 20.25 & 16.73 & 17.03 & 9.94 &&&&&  \\ \hline
\hline
\multicolumn{2}{|c|}{Rank} & 1 & 5 & 3 & 4 & 2 & 1 & 2 & 5 & 3 & 4 \\ \hline

\end{tabular}
 }
\end{table*}

\subsection{Investigating Stream Clustering Algorithms}

\vl{In this section, we test four stream clustering algorithms implemented in River library \citep{montiel2021river}: CluStream \citep{aggarwal2003framework}, DBSTREAM \citep{hahsler2016clustering}, DenStream \citep{cao2006density}, and STREAMKMeans \citep{o2002streaming}. We keep the default settings for these algorithms in the River library. We use the Silhouette coefficient \citep{rousseeuw1987silhouettes} to evaluate the clustering results without requiring ground-truth labels. It has the range from -1 to 1. A higher positive value indicates that the samples are well-clustered. A negative value indicates that the samples may be wrongly clustered. A near-zero value indicates that the clusters are overlapping with each other. }

\vl{Table \ref{tbl:cluster_value} shows the Silhouette coefficient values and Table \ref{tbl:cluster_time} shows the running time. Considering both effectiveness and efficiency, STREAMKMeans is much more computational efficient, and it achieves relatively higher Silhouette coefficient in 3 out of the 5 datasets. However, note that STREAMKMeans get mostly near-zero Silhouette coefficient value, its clusters are overlapping with each other, which is still not a good solution. }

\vl{In summary, the clustering task of real-world relational data streams is challenging and under-explored, where the tested four stream clustering algorithms cannot achieve good clustering results. It requires more systematic methodology to study stream clustering, which we leave as future work. }

\begin{table}[h]
\centering
\caption{\vl{The Silhouette coefficient values of stream clustering algorithms. Higher value is better. ``Error'' in DBSTREAM means that all samples are assigned into one cluster, in which the metric cannot be calculated. ``TLE'' means the algorithm cannot finish in 12 hours. } }
\label{tbl:cluster_value}
\resizebox{\columnwidth}{!}{
\begin{tabular}{|c|c|c|c|c|}
\hline
Dataset & CluStream & DBSTREAM & DenStream & STREAMKMeans \\ \hline
ROOM & -0.160 & \textbf{0.543} & 0.196 & -0.045 \\ \hline
ELECTRICITY & -0.150 & Error & -0.306 & \textbf{0.003} \\ \hline
INSECTS & -0.011 & -0.106 & -0.005 & \textbf{0.000} \\ \hline
AIR & -0.030 & TLE & \textbf{0.148} & 0.035 \\ \hline
POWER & -0.014 & TLE & 0.025 & \textbf{0.674} \\ \hline

\end{tabular}
 }
 
\end{table}

\begin{table}[h]
\centering
\caption{\vl{The running time in seconds of stream clustering algorithms. } }
\label{tbl:cluster_time}
\resizebox{\columnwidth}{!}{
\begin{tabular}{|c|c|c|c|c|}
\hline
Dataset & CluStream & DBSTREAM & DenStream & STREAMKMeans \\ \hline
ROOM & 172 & 2,024 & 8 & \textbf{0.45} \\ \hline
ELECTRICITY & 34 & N/A & 243 & \textbf{0.82} \\ \hline
INSECTS & 3,801 & 6,645 & 138 & \textbf{4.5} \\ \hline
AIR & 2,476 & N/A & 18 & \textbf{1.7} \\ \hline
POWER & 702 & N/A & 11 & \textbf{0.9}  \\ \hline

\end{tabular}
 }
 
\end{table}

\section{Dataset Description}
\label{sec:datades}

Table \ref{tbl:datasetcls} and \ref{tbl:datasetreg} present statistical information of the datasets explored in this work. We also list the extracted statistics of three open environment challenges (drift, anomaly, and missing value), which are the three dimensions visualized in Figure \ref{fig:clustershow} and the criteria in Table \ref{tbl:all}. We briefly describe these datasets by categories in the following sections.

\begin{table*}[htpb]
\caption{Basic information of real-world relational data streams (classification task).}
\label{tbl:datasetcls}
\centering

\newcommand{\y}{\ding{51}}
\newcommand{\n}{\ding{55}}
\resizebox{2.1\columnwidth}{!}{
\begin{tabular}{|c|c|c|c|c|c|c|c|c|c|}
\hline
Dataset  & Instances & Features & Classes & Type & Anomaly & Drift & Missing value \\
\hline
Bitcoin Heist Ransomware Address & 2,916,697 & 6 & 27&Commerce & 0.054&0.301&0\\ \hline
Room Occupancy Estimation & 10,129 & 16 & 4&Others &0.106&0.213&0\\ \hline
Electricity Prices & 45,312 & 7 & 2&Commerce&0.052&0.221&0\\ \hline
Airlines & 539,383 & 6 & 2 &Commerce &0.007&0.143&0\\ \hline
Forest Covertype & 581,012 & 54 & 7 &S\&T &0.043&0.249&0 \\ \hline
INSECTS-Abrupt (balanced) & 52,848 & 33 & 6 &S\&T &0.041&0.131&0\\ \hline
INSECTS-Abrupt (imbalanced) & 355,275 & 33 & 6 &S\&T&0.051&0.125&0 \\ \hline
INSECTS-Incremental (balanced) & 57,018 & 33 & 6&S\&T & 0.033&0.183&0\\ \hline
INSECTS-Incremental (imbalanced) & 452,044 & 33 & 6&S\&T & 0.046&0.143&0\\ \hline
INSECTS-Incremental-abrupt-reoccurring (balanced) & 79,986 & 33 & 6&S\&T &0.055&0.196&0 \\ \hline
INSECTS-Incremental-abrupt-reoccurring (imbalanced) & 452,044 & 33 & 6 &S\&T &0.044&0.218&0 \\ \hline
INSECTS-Incremental-gradual (balanced) & 24,150 & 33 & 6 &S\&T &0.035&0.198&0 \\ \hline
INSECTS-Incremental-gradual (imbalanced) & 143,323	& 33 & 6  &S\&T &0.038&0.203&0\\ \hline
INSECTS-Incremental-reoccurring (balanced) & 79,986 & 33 & 6 &S\&T & 0.040&0.192&0\\ \hline
INSECTS-Incremental-reoccurring (imbalanced) & 452,044 & 33 & 6 &S\&T &0.044&0.202&0 \\ \hline
INSECTS-Out-of-control & 905,145 & 33 & 24 &S\&T &0.039&0.067&0\\ \hline
KDDCUP99 & 494,021 & 41 & 23 &S\&T &0.005&0.111&0\\ \hline
NOAA Weather & 18,159 & 8 & 2 &Ecology &0.028&0.199&0\\ \hline
Safe Driver & 595,212 & 57 &2 &Commerce & 0.008&0.062&0 \\ \hline
BLE RSSI dataset for Indoor Localization Data Set & 9,984 & 5 & 3&Others& 0.044&0.217&0\\
\hline
\end{tabular}
}

\end{table*}

\begin{table*}[htpb]
\centering
\caption{Basic information of real-world relational data streams (regression task).}
\label{tbl:datasetreg}
\newcommand{\y}{\ding{51}}
\newcommand{\n}{\ding{55}}
\resizebox{2.1\columnwidth}{!}{
\begin{tabular}{|c|c|c|c|c|c|c|c|c|}
\hline
Dataset  & Instances & Features & Target & Type& Anomaly & Drift & Missing value  \\
\hline
Italian City Air Quality & 9,358 & 12 & PT08.S1 (CO) &Ecology &0.035&0.303&0.163 \\ \hline
Energy Prediction & 19,735 & 25 & Appliances & Power &0.061&0.326&0 \\ \hline
Beijing Multi-Site Air-Quality Aotizhongxin & 35,064 & 11 & PM2.5&Ecology & 0.022&0.145&0.066\\ \hline
Beijing Multi-Site Air-Quality Changping & 35,064 & 11 & PM2.5&Ecology &0.027&0.145&0.029\\ \hline
Beijing Multi-Site Air-Quality Dingling & 35,064 & 11 & PM2.5&Ecology&0.026&0.146&0.040\\ \hline
Beijing Multi-Site Air-Quality Dongsi & 35,064 & 11 & PM2.5 &Ecology&0.043&0.139&0.038\\ \hline
Beijing Multi-Site Air-Quality Guanyuan & 35,064 & 11 & PM2.5 &Ecology&0.031&0.142&0.033\\ \hline
Beijing Multi-Site Air-Quality Gucheng & 35,064 & 11 & PM2.5 &Ecology&0.035&0.136&0.020\\ \hline
Beijing Multi-Site Air-Quality Huairou & 35,064 & 11 & PM2.5 &Ecology&0.029&0.134&0.057\\ \hline
Beijing Multi-Site Air-Quality Nongzhanguan & 35,064 & 11 & PM2.5&Ecology&0.028&0.138&0.017\\ \hline
Beijing Multi-Site Air-Quality Shunyi & 35,064 & 11 & PM2.5&Ecology &0.029&0.139&0.040\\ \hline
Beijing Multi-Site Air-Quality Tiantan & 35,064 & 11 & PM2.5 &Ecology&0.042&0.136&0.032\\ \hline
Beijing Multi-Site Air-Quality Wanliu & 35,064 & 11 & PM2.5 &Ecology&0.022&0.144&0.035\\ \hline
Beijing Multi-Site Air-Quality Wanshouxigong & 35,064 & 11 & PM2.5 &Ecology&0.032&0.143&0.024\\ \hline
Beijing PM2.5 &	43,824 & 7 & PM2.5 &Ecology&0.052&0.184&0.012\\ \hline
Indian Cities Weather Bangalore & 11,894 & 5 & Average temperature&Ecology&0.025&0.081&0.120 \\ \hline
Indian Cities Weather Bhubhneshwar & 11,894 & 5 & Average temperature&Ecology &0.020&0.044&0.157\\ \hline
Indian Cities Weather Chennai & 11,894 & 5 & Average temperature&Ecology &0.023&0.055&0.158\\ \hline
Indian Cities Weather Delhi & 11,894 & 5 & Average temperature &Ecology&0.021&0.115&0.146\\ \hline
Indian Cities Weather Lucknow & 11,894 & 5 & Average temperature &Ecology&0.022&0.111&0.198\\ \hline
Indian Cities Weather Mumbai & 11,894 & 5 & Average temperature &Ecology&0.020&0.044&0.157\\ \hline
Indian Cities Weather Rajasthan & 11,894 & 5 & Average temperature &Ecology&0.025&0.081&0.120\\ \hline
Household Electric Consumption & 2,075,259 & 6 & Global active power&Power&0.035&0.399&0.005\\ \hline
Metro Interstate Traffic Volume & 48,204 & 7 & Traffic volume&Commerce&0.023&0.077&0 \\ \hline
PM2.5 Data of Five Chinese Cities Beijing & 52,584 & 8 & PM2.5&Ecology&0.040&0.218&0.139 \\ \hline
PM2.5 Data of Five Chinese Cities Chengdu & 52,584 & 8 & PM2.5&Ecology&0.067&0.186&0.197 \\ \hline
PM2.5 Data of Five Chinese Cities Shanghai & 52,584 & 8 & PM2.5 &Ecology&0.028&0.186&0.193\\ \hline
PM2.5 Data of Five Chinese Cities Shenyang & 52,584 & 8 & PM2.5 &Ecology&0.056&0.192&0.255\\ \hline
PM2.5 Data of Five Chinese Cities Guangzhou & 52,584 & 8 & PM2.5 &Ecology&0.031&0.297&0.128\\ \hline
Power Consumption of Tetouan City & 52,417 & 7 & Zone 3 power consumption &Power&0.022&0.352&0\\ \hline
Bike Sharing Demand & 10,886 & 7 & Count &Commerce&0.025&0.287&0 \\ \hline
Allstate Claims Severity & 188,318 & 130 & Loss &Commerce&0.013&0.045&0\\ \hline
Portugal Parliamentary Election & 21,843 & 28 & Final mandates&Social&0.042&0.080&0 \\ \hline
News Popularity & 93,239 & 11 & Popularity index &Social&0.027&0.169&0.001\\ \hline
Taxi Trip Duration & 1,458,644 & 11 & Duration &Commerce&0.023&0.217&0\\

\hline
\end{tabular}
}

\end{table*}

\subsection{Ecology}

 \textbf{Italian City Air Quality} \citep{de2008field}: This dataset records the hourly averaged concentrations for air pollutants as well as meteorological data in an Italian city from 2004 to 2005. We regard PT08.S1 (CO) as the target of the regression task. 
 
 \textbf{Beijing Multi-Site Air Quality} \citep{zhang2017cautionary}: This dataset contains the hourly averaged concentrations
for air pollutants and meteorological data from 12 sites in Beijing from 2013 to 2017. We regard PM2.5 as the target of the regression task.
 
 \textbf{Beijing PM2.5} \citep{liang2015assessing}:  This dataset contains the hourly averaged concentrations for air pollutants and meteorological data from 2010 to 2014. We regard PM2.5 as the target of the regression task.
 
\textbf{PM2.5 Data of Five Chinese Cities} \citep{liang2016pm2}: This dataset contains the hourly averaged concentrations for air pollutants
and meteorological data in five Chinese cities from 2010 to 2015. We regard PM2.5 as the regression target.

\textbf{Indian Cities Weather}: This dataset contains temperature and precipitation measurements from seven cities in India. We regard the average temperature as the regression target.

\textbf{NOAA Weather} \citep{ditzler2013incremental}: This dataset contains meteorological measurements in Bellevue,
Nebraska for 50 years. The task is to predict whether it rains or not. 

\subsection{Power Consumption}


\textbf{Energy Prediction}
\citep{candanedo2017data}: The task of this dataset is to predict appliance energy use. We discard the last two unrelated random variables as they are not real-world data.

\textbf{Power Consumption in Tetouan City} \citep{salam2018comparison}: This dataset contains power consumption and meteorological data in 3 zones in Tetouan city. We regard zone 3 power consumption as the regression task.

\textbf{Household Electric Consumption} \citep{Dua:2019}: This dataset contains electric usage measurements collected in an individual household in Sceaux from 2006 to 2010. We set global active power as the regression target.

\subsection{Commerce}


\textbf{Bitcoin Heist Ransomware Address Dataset} \citep{goldsmith2020analyzing}: This dataset contains Bitcoin transaction features and the respective ransomware labels from 2009 to 2018. The goal is to predict the ransomware label. We discard the Bitcoin address feature since it is hard to be converted to statistics.

\textbf{Electricity Prices} \citep{harries1999splice}: This dataset contains electricity price information in Australian New South Wales Electricity Market over around 5 months. The task is to predict whether the electricity price goes up or down.

\textbf{Airlines} \citep{ikonomovska2011learning}: This dataset contains flight departure and arrival information in the USA
from 1987 to 2008. The task is to predict whether a flight will be delayed or not.


\textbf{Bike Sharing Demand} \citep{bike-sharing-demand}: This dataset contains the bike sharing data in a city. The task is to predict how many bikes are rented in an hour. Feature ``casual'' and ``registered'' are discarded, since their sum is the answer. 

\textbf{Porto Seguro's Safe Driver} \citep{porto-seguro-safe-driver-prediction}: This dataset contains information about automobile insurance claims. The task is to predict whether the driver will file an insurance claim in the next year.


\textbf{Allstate Claims Severity} \citep{allstate-claims-severity}: This dataset records insurance claims filed with Allstate, an insurance company. The regression task is to predict the amount of insurance claims.


\textbf{Metro Interstate Traffic Volume}: This dataset records hourly traffic volume in a metropolitan interstate along with weather and holiday information. We regard the traffic volume as the regression target.

\textbf{Taxi Trip Duration}: This dataset contains the information of taxi trips. The regression target is to predict the duration of a taxi trip.

\subsection{Social}

\textbf{Portugal Parliamentary Election} \citep{moniz2019real}: This dataset tracks the evolution of the 2019 Portuguese Parliamentary Election and the related information. The target is to predict the final mandates.

\textbf{News Popularity} \citep{moniz2018multi}: This dataset contains feedback and responses of several hot topics on 3 social media platforms for 8 months. We choose the popularity of the news item on Facebook as the regression target.

\subsection{Science \& Technology}


\textbf{KDDCUP99} \citep{tavallaee2009detailed}: This dataset contains records of network connections and TCP dump data in a local area network. Each record is labeled as either normal or one of the 22 different types of attack. The task is to predict the type of attack.

\textbf{Forest Covertype} \citep{blackard1999comparative}: This dataset contains forest information of 30*30 meter cells in the USA. The task is to predict the predominant cover type.

\textbf{INSECTS} \citep{souza2020challenges}: This dataset is collected from biology experiments of insects where drifts are simulated by changing the temperature. The task is to predict the insect type. It contains 11 sub-datasets.

\subsection{Others}


\textbf{BLE RSSI dataset for Indoor Localization Data Set} \citep{mohammadi2017semi}: This dataset contains Bluetooth Low Energy (BLE) Received Signal Strength Indications (RSSI) collected from smartphones for 20 minutes. Participants are instructed to walk inside their limited area with activated smartphones. The task is to predict which area the participant is in.

\textbf{Room Occupancy Estimation} \citep{singh2018machine}: This dataset contains environmental sensor statistics in a room. The task is to predict the number of people in a room, which varies between 0 and 3. 

\textbf{Rialto Bridge Timelapse} \citep{losing2016knn}: This dataset contains images captured for 10 buildings from a fixed camera in 20 days. 27 key features are extracted for the picture of each building. The changing weather and light can lead to distribution drifts in features. The task is to identify the correct building.

\end{document}